\documentclass[symmetry,article,accept,pdftex,moreauthors]{Definitions/mdpi} 

\usepackage{blindtext}

\usepackage{xcolor}
\usepackage[utf8]{inputenc} 
\usepackage[T1]{fontenc}    
\usepackage{url}            
\usepackage{textcase}
\usepackage{booktabs}       
\usepackage{amsfonts}       
\usepackage{amsmath}
\usepackage{nicefrac}       
\usepackage{microtype}      
\usepackage{xcolor}         
\usepackage{bbold} 
\usepackage{tikz}
\usepackage{pgfplots}
\usepackage{multirow}
\usepackage{float}
\usepackage{lastpage}
\usepackage{graphicx}
\usepackage{subfig}
\usepackage{wrapfig}
\usepackage{upgreek}
\usepackage{makecell}

\newcommand*{\horzbar}{\rule[.5ex]{2.5ex}{0.5pt}}

\newcommand{\matr}[1]{#1}                  
\newcommand{\rvector}[1]{\mathcal{\mathbf{#1}}}   
\newcommand{\vcomp}[2]{#1_{#2}}              
\newcommand{\rvcomp}[2]{\mathit{#1}_{#2}}    

\newcommand{\filter}{\boldsymbol{\psi}}
\newcommand{\gen}{\mathfrak{G}}

\firstpage{1} 
\pubvolume{1}
\issuenum{1}
\articlenumber{0}
\pubyear{2025}
\copyrightyear{2025}
\externaleditor{Firstname Lastname} 
\datereceived{14 January 2025} 
\daterevised{4 March 2025} 
\dateaccepted{8 March 2025} 
\datepublished{12 March 2025} 
\hreflink{https://doi.org/} 

\Title{SymmetryLens: Unsupervised Symmetry Learning via Locality and Density Preservation}

\TitleCitation{SymmetryLens: Unsupervised Symmetry Learning via Locality and Density Preservation}

\Author{Onur Efe 
 \orcidA{} and Arkadas Ozakin *\orcidB{}}

\AuthorNames{Onur Efe, Arkadas Ozakin}

\isAPAStyle{%
       \AuthorCitation{Efe, O., \& Ozakin, A.}
         }{%
        \isChicagoStyle{%
        \AuthorCitation{Efe, Onur and Arkadas Ozakin.}
        }{
        \AuthorCitation{Efe, O.; 
 Ozakin, A.}
        }
}

\address[1]{%
Department 
 of Physics, Bogazici University, Istanbul, 34342
, Turkey; onur.efe@bogazici.edu.tr}

\corres{\hangafter=1 \hangindent=1.05em \hspace{-0.82em}Correspondence: arkadas.ozakin@bogazici.edu.tr}

\abstract{We develop a new unsupervised symmetry learning method that starts with raw
    data and provides the minimal generator of an underlying Lie group of symmetries, together
    with a symmetry-equivariant representation of the data, which turns the hidden symmetry into an explicit one. The method is able to learn the
    pixel translation operator from a dataset with only an approximate 
    translation symmetry and can learn quite different types of symmetries that 
    are not apparent to the naked eye. The method is based on the formulation of an
    information-theoretic loss function that measures both the degree of symmetry
    of a dataset under a candidate symmetry generator and a 
    proposed notion of \emph{locality
} of the samples, which is coupled to symmetry.
    We demonstrate that this coupling between symmetry and locality, together with 
    an optimization technique developed for entropy estimation, results in a stable
    system that provides reproducible results.}


\begin{document}

\section{Introduction}\label{sec:introduction}

The spectacular success of convolutional neural networks (CNNs) has triggered a wide range of approaches to their generalization. CNNs are equivariant under pixel translations, which is also a property of the information content in natural image data. By~using a weight-sharing scheme that respects such an underlying symmetry, it has been possible to introduce a powerful inductive bias in line with the nature of the data, resulting in highly accurate predictive models. How can this approach be generalized?

Translations form a \emph{group}, and, for those cases where a more general underlying symmetry group is known, a~mathematically elegant generalization can be developed in the form of group convolutional networks \citep{cohen2016group}. However, data often have underlying symmetries that are \emph{not known explicitly beforehand}. In~such cases, a~method for \emph{discovering} the underlying symmetry from data would be highly~desirable.

Learning symmetries from data would allow one to develop an efficient weight-sharing scheme as in CNNs, but, even without this practical application, simply being able to discover unknown symmetries is of considerable interest in itself. In~many fields, symmetries in data provide deep insights into the nature of the system that produces the data. For~example, in~physics, continuous symmetries are closely related to conservation laws, and knowledge of the symmetries of a mechanical system allows one to develop both analytical \citep{Babelon_Bernard_Talon_2003} and numerical
\citep{hairer2010geometric} methods for investigating dynamics. More generally, the~discovery of a new symmetry in a dataset would almost surely trigger research into the underlying mechanisms generating that~symmetry.

In this paper, we develop a new method for discovering symmetries from raw 
data. For a dataset that has an 
underlying unknown symmetry (analogous to translation symmetry) provided by the action of a Lie group, the~model provides a generator of 
the symmetry (analogous to pixel translations). In~other words, we develop an 
unsupervised method, which has no 
other tasks such as classification or regression: the data are provided to the system without any labels of any 
sort, and~out comes the generator of the symmetry group action. The~setup also
creates, as~a byproduct, a~symmetry-based representation of the data,
whose importance is forcefully emphasized by \citet{higgins2022symmetry} and \citet{anselmi2022symmetry}. This symmetry-based representation 
can also be used as an adapter between raw data and regular~CNNs.

In the context of symmetry learning, one sometimes restricts attention to 
symmetry transformations that act on samples only through an action on their
component indices. For~instance, in~the case of images, a~translated image is 
obtained by shifting the pixel indices in an appropriate way, without~otherwise 
transforming the data. While such transformations form an important class of 
symmetry actions, our setting is more general in that we consider a symmetry group of 
transformations acting \emph{directly on samples}, and~so-called 
``regular representations'' (of which pixel translations are an example) are 
only a special case of our~approach.

\begin{figure}[H]
    \includegraphics[width=\textwidth]{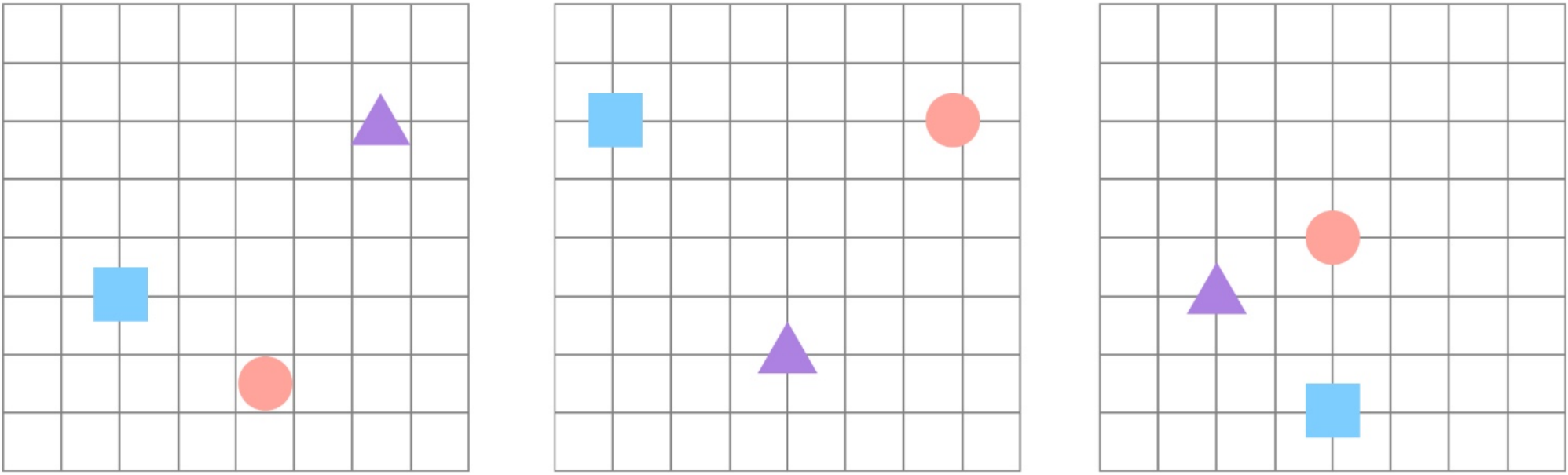}
    \caption{Symmetry 
 actions on local~objects.}
    \label{fig:group-acting-locally}
\end{figure}

Another avenue of generalization relevant to our methodology is 
the notion of a \emph{local} symmetry, as~opposed to the more 
commonly encountered notion of global or~``rigid''  symmetries.
The translation symmetry of the underlying distribution of an image
dataset does not come from the fact that rigidly translated versions of images are 
commonly encountered; they are not. One sometimes synthetically creates
such samples for purposes of data augmentation, but, in reality, multiple 
pictures of the same exact setting are rare. What does happen is that 
real images consist of \emph{local building blocks} whose locations 
have a distribution that is approximately invariant
under translations (see Figure~\ref{fig:group-acting-locally}). 
Each image consists of the superposition of a number of such building blocks,
and the approximate translation symmetry of image data 
is, in~this sense, tied to an underlying \emph{local} 
version of the translation symmetry. \emph{We ignore the dependence
between the locations of different object types.}
The importance of being able to deal with such a
flexible/local notion of symmetry has been emphasized by
\citet{bronstein2021geometricdeeplearninggrids} and \citet{anselmi2019symmetry}. 

With this motivation, we seek to develop a symmetry learning methodology 
that finds transformations that respect the underlying distribution
of a dataset of samples and~respect a notion of locality that is intertwined 
with the~symmetry. 

\begin{figure}[H]
    \includegraphics[width=\textwidth]{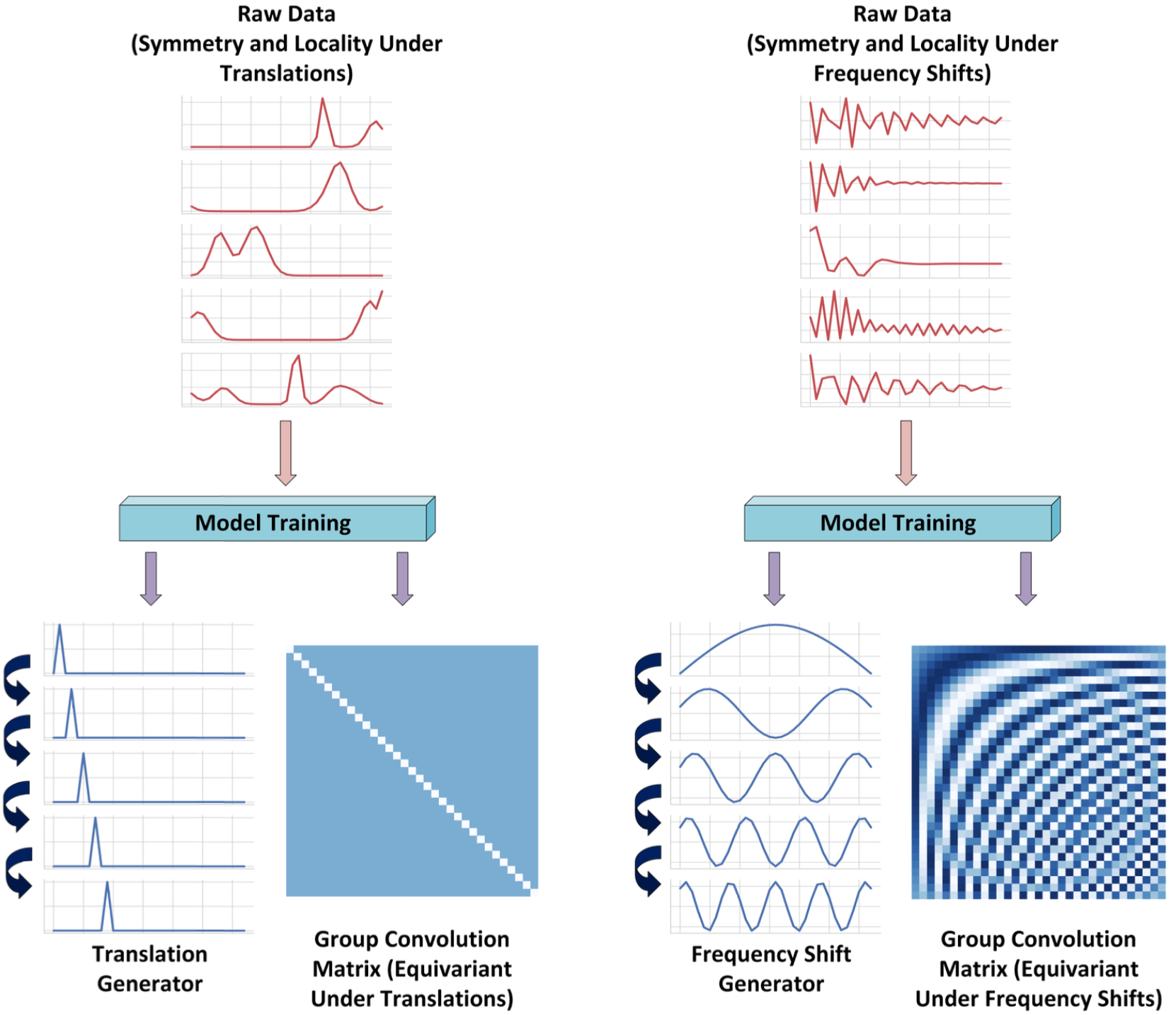}
    \caption{A visual representation of the model input and output. At~the top, we see samples from the datasets used for model training; these datasets are approximately invariant and local under a given group action. The~output, after~training, is the learned minimal generator of the group action shown pictorially here and~a symmetry-based representation that is shown
    as a matrix here. For~the case on the left, the~dataset is local and invariant under ``1-dimensional pixel 
    translations''. For~the case on the right, 
    the dataset is local/symmetric under frequency shifts (sending each Fourier basis vector to the next), and~    the relevant symmetry-based representation is a discrete Fourier transform. 
    The model can recover the given symmetry generators in each one of these example cases
    and others.}
    \label{fig:intro}
\end{figure}

Before providing an outline of our methodology, let us provide a quick overview of the
capabilities of our system.
Consider a dataset whose samples are superpositions of local basis signals
with locations randomly selected from a uniform distribution over a finite range 
(see Figure~\ref{fig:intro}). Each localized signal, with possibly a different width and height,
is a local building block in the sense described above. 
The distribution of such samples is approximately invariant under 
translations (modulo edge effects)
. The~question is, given such a dataset,
without any other guidance (e.g., a~supervised learning problem 
on this dataset with ground truth labels that are invariant under the translation symmetry), 
can one find that the operation of (one-dimensional) pixel translations is a symmetry of this dataset?
Our model is, indeed, able to start with a dataset of this form and discover 
the matrix that represents pixel translations, sending each component of a sample to the next 
component. 

\vspace{-6pt}

Although this is not a straightforward task, rediscovering the familiar case of (\mbox{one-dimensional}) 
pixel translations from such a dataset is perhaps the first thing to demand of a method that claims 
to learn generalized symmetries from raw data in an unsupervised manner. However, our approach 
allows the system to discover symmetries that are much less obvious. Suppose we take a dataset that 
is again approximately invariant under pixel translations and apply a fixed 
permutation to the pixels to obtain a new dataset. The~underlying symmetry of this new dataset will be much less 
obvious to the naked eye: the generator of the symmetry will now be an operator like Pixel 1$\,\to\,$Pixel 
17$\,\to\,$\mbox{Pixel 12}, etc. We demonstrate that our method can discover this symmetry just as well as it 
discovers pixel translation. Moreover, the~model learns a symmetry-based 
representation as a byproduct, and~this representation
unscrambles the pixel order to make the hidden symmetry manifest. 
While we find this highly encouraging, this, too, is a setting where 
the symmetry operation acts on the pixel~indices.

As a final example, consider a symmetry action in the form of 
\emph{frequency shifts}, in~the sense that the dataset under consideration
comprises time-series data that are approximately symmetric under the three-step action of 
(1) Fourier transformation (more precisely, we use discrete sine transformation I), (2) shifting all the frequency
components by the same amount (i.e., a~translation in frequency space), and~
(3) inverse Fourier transformation (see Figure~\ref{fig:intro}). This action clearly is not 
obtained by simply shuffling around the component indices of the input vectors.
Our method learns the matrix that represents the minimal generator of this action as well.
In this case, the~symmetry-based representation constructed by the model as a byproduct
comes out to be almost exactly a discrete Fourier transform matrix.
In other words, by~simply ``looking at'' a dataset that has an approximately
local symmetry under frequency shifts, the~method learns that the appropriate 
symmetry-based representation for this dataset is a Fourier~transform.

In the following sections, we provide the technical details of the method itself, but~the fundamental 
intuition is rather simple, and~we next provide a quick~description.

Our approach is based on formulating two important properties to generate a 
\emph{symmetry group} for a dataset (for the definition of a group in the context of
symmetry learning, see, e.g.,~\citet{helgason1979differential}), namely \emph{invariance} and \emph{locality}.
An appropriate loss function that measures the degree to which these properties are satisfied
enables us to find the underlying symmetries by gradient-based optimizers. 
While versions of \emph{invariance} are often used in the symmetry learning 
literature, we believe that the inclusion of a \emph{locality} property is what provides 
our method the power it~has.

\textbf{Invariance. 
} The first natural characteristic of symmetry is the preservation of the 
underlying distribution. In~other words, the~original data and their transformed version under a 
symmetry map should have \emph{approximately} the same distribution. In~the case of images, this 
means that the joint distribution of all the pixel activations should be invariant 
under the global translation of pixels (ignoring boundary effects). 
For a general symmetry group action (beyond translations), 
such an invariance may not be obvious to the naked eye, 
but, once one knows the underlying symmetry, one should be 
able to confirm the behavior. (In the
\emph{continuous} case, one views data samples as sample functions or
realizations of an underlying
continuous \emph{stochastic process}, and~a stochastic 
process with the required invariance property is 
called a \emph{strongly stationary} process under the action of the symmetry group. Apart from the theoretical setting in Section \ref{section:data_model}, in~this paper, we 
restrict our attention to the discretized version of such a continuous underlying symmetry.)

As emphasized by \citet{desai2022symmetry}, the~space of all the density-preserving maps \citep{ozakin2011density} for a multidimensional dataset is large and includes maps that one would not normally want to call symmetries, so simply seeking density- (or 
distribution-) preserving maps will not necessarily allow one to discover symmetries. This property, 
nevertheless, is a reasonable characteristic to expect from a symmetry transformation. Implementing a loss function 
that measures the degree of density preservation 
will help us to achieve transformations with this~property.

\textbf{Locality.} To identify the other fundamental property we will demand of a symmetry group action, we turn back to CNNs for inspiration. A~CNN model not only respects the underlying translation symmetry but does so in a \emph{local} way, in~the sense that each filter has a limited spatial extent. This inductive bias, as~well, corresponds to an underlying property of image data. As~mentioned above, images contain representations of \emph{objects}, which themselves have locality along the symmetry directions, and~local filters do a good job of extracting 
such local information. While locality and symmetry are distinct properties, they are often coupled, and~this coupling often hints at something fundamental about the processes generating the data. In~physics, space and time translation symmetries and spacetime locality are intimately coupled, and~this locality-symmetry
is a fundamental property of quantum field theories describing the standard model of particle 
physics \citep{weinberg1997quantumfieldtheorydid}. 

We choose to enforce a generalized notion of locality as the second fundamental property that the symmetry group should satisfy. The symmetry directions should be coupled to the locality directions in data. To~make this proposal concrete, consider the action of
a single local CNN filter. If~we apply the same filter to an image and its slightly translated versions, we obtain various scalar representations
of the image. These scalars will be strongly 
\emph{dependent} random variables that contain similar information
when the amount of translation is small, but, with larger 
translations, the~dependence and similarity~decrease.

We will require that a general unknown group of symmetries should satisfy a similar locality property. Once again, let us start
with the assumption that we have a ``local filter'', i.e.,~
a scalar-valued map, this time \emph{local along the unknown symmetry direction}. Applying this filter to a sample, we obtain a scalar. 
First transforming the sample by a symmetry transformation and~\emph{then} applying the filter, we obtain another scalar. 
Assume that the information in the samples is indeed local along the symmetry direction. Then, for~
``small'' transformations, the~two
scalars should be strongly dependent random variables
that encode similar information. For~``larger'' transformations,
the similarity should decrease. A~loss function that quantifies 
this behavior would thus help us to find 
symmetry groups that have the required coupling with the~locality.

In addition to the problem of making the concepts in the scare quotes 
above precise (which we address in our method sections),
this approach also faces an immediate chicken--egg problem: 
to conduct a search for a set of symmetry transformations
in this way, we would first need a scalar representation, a~\emph{filter}, which itself is local along 
the symmetry direction. But, to find such a filter, we first 
need to know the set of symmetry transformations (under which the filter will be local).

\begin{figure}[!ht]
    \includegraphics[scale=0.175]{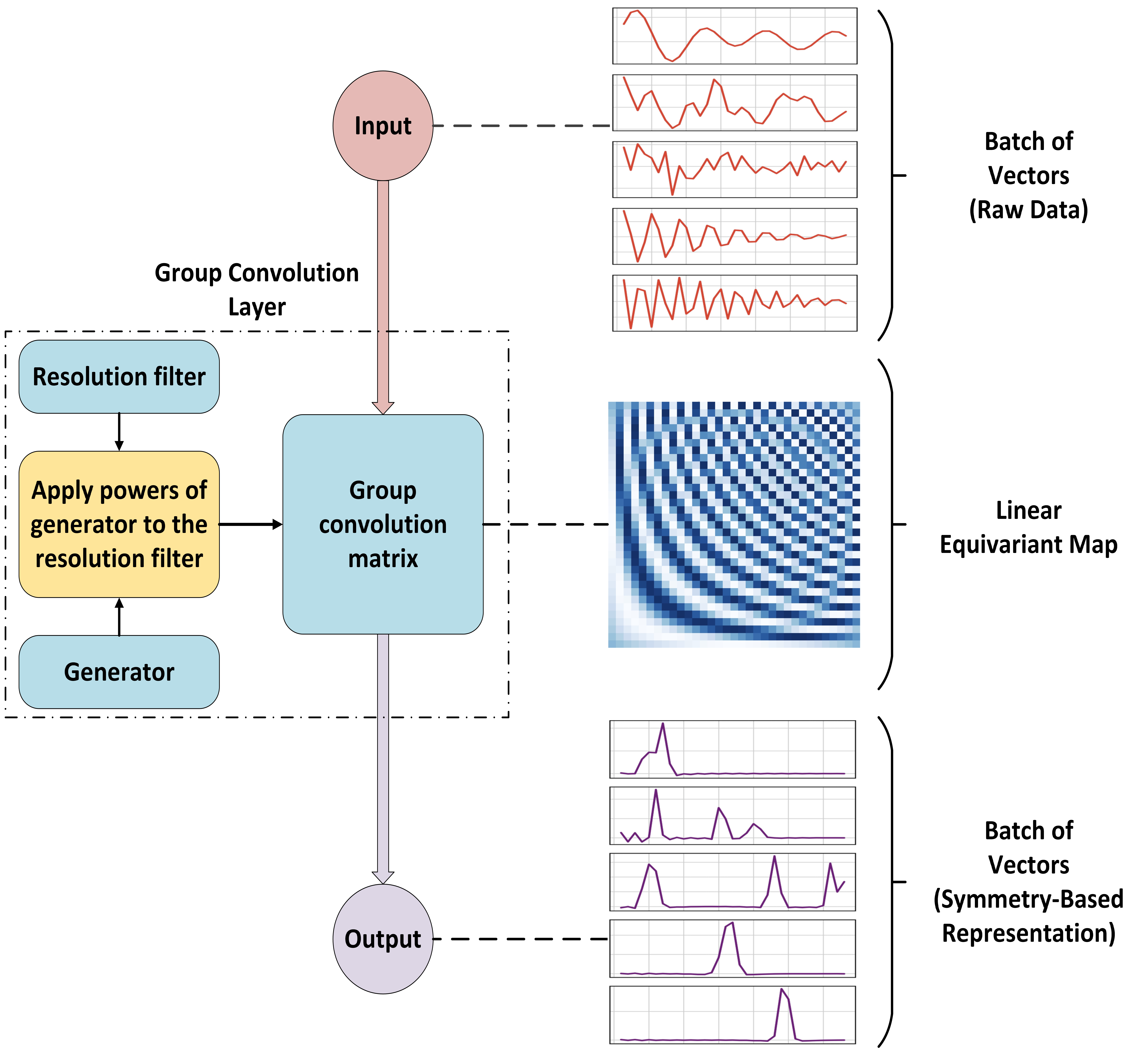}
    \caption{Overview 
 of the method. Each sample goes through the group convolution
    map formed by combining the resolving filter with the powers of the candidate symmetry
    transformation. The~locality and symmetry losses are applied to the resulting representation. In~the case shown here, the~samples are local/symmetric under
    frequency shifts, and~the group convolution matrix learned is effectively
    the matrix representing the discrete sine transform.}
    \label{fig:architecture}
\end{figure}

We solve this problem by a sort of bootstrapping: we seek the 
local scalar representation and the symmetry group simultaneously 
and self-consistently using an appropriate 
loss function that applies to both. As~described 
below, this coupled search actually works, finding the appropriate 
minimal generators of symmetry, together with 
a filter that behaves as a ``delta function'' in the direction of symmetry transformations. 
As a byproduct, one also obtains a group-equivariant symmetry-based 
representation of data, which makes the hidden symmetries~explicit. See Figure \ref{fig:architecture} for a visual representation of the method.

From the outset, we emphasize that we do not assume the data space 
to be a so-called homogeneous space under the action of the Lie 
group of symmetries. More explicitly, we do not assume that the 
symmetry group is large enough to map any given point in the 
sample space to any other point. This would be a highly restrictive 
assumption, which is not, for~example, satisfied by image~data.  

We find it satisfying that the approach described here 
actually works, but~the details, of~course, matter.
The dynamics of the optimization of a loss function are complicated, 
and various seemingly similar implementations of the same intuition 
can result in widely varying results. The~method we detail in the 
following sections is a specific implementation of the ideas of 
locality and invariance described above, and~it results in a 
highly accurate and \emph{stable} system that provides 
reproducible~results.

In summary, the~main contributions of the paper~are the following:
\begin{itemize}
    \item The formulation of a new approach to symmetry learning 
    via maps that respect both symmetry and (a proposed notion 
    of) locality, coupled in a natural way. The~outputs of the method
    are both \emph{minimal generators} of the symmetry and a symmetry-based data representation that, in~a sense, makes the hidden symmetry manifest. This representation can be used as input for a regular CNN,  allowing the model to work as an adapter between raw data and the whole machinery of CNNs.
    \item The formulation of an information-theoretic loss function 
    encapsulating the formulated symmetry and locality properties.
    \item \textls[-15]{The development of optimization techniques (``time-dependent rank'', see \mbox{Section~\ref{par:controlling_joint_entropy_rank}})} that 
    result in highly robust and reproducible results.
    \item The demonstration of the symmetry recovery and symmetry-based representation capabilities on quite different 
    sorts of examples, including 1D pixel translation symmetries, a~shuffled version of pixel translations, and~frequency shifts
    using dataset dimensionalities as high as 33 (i.e., the~relevant symmetry generators with the shape $33\times 33$). 
\end{itemize}

We note that, in the search for symmetry generators, there
are no hard-coded simplifications such as sparsity in the 
symmetry generators or symmetry-based representations. 
The model indeed needs to learn the relevant matrices without 
utilizing an underlying simplifying assumption or 
factorization.

The rest of the paper is structured as follows. We will briefly summarize the current literature in Section~\ref{section:related_work}, while Section~\ref{section:data_model} includes the theoretical setup motivating our approach. We describe the details of the method in Section~\ref{sec:materials_and_methods}, and~we investigate the performance and robustness of the method in Section~\ref{section:results}. Section~\ref{section:discussion} includes a summary of our contributions, future research directions, and~limitations.

\section{Related~Work}\label{section:related_work}
Symmetry learning in neural networks is a rapidly evolving area
of research, driven both by the value of symmetries for
natural sciences and by the usefulness of symmetry-based inductive biases in
model architectures. The~current literature involves nuanced 
definitions for the symmetry learning problem as well as different
symmetry learning schemes. The~techniques used in symmetry learning
can be roughly classified into  supervised, self-supervised, and~
unsupervised learning~approaches.

\textbf{Supervised learning-based approaches}
combine a supervised problem with a search for symmetry maps
that work in harmony with the supervised model in an appropriate sense. 
\citet{benton2020learning} use parametrized transformations of input
data (such as rotations or affine transformations) for data augmentation and average the predictions over augmented samples to obtain a final 
prediction that is invariant under the chosen transformations.
Using a loss function that encourages the exploration of a range of 
transformations, they are able to find, e.g.,~rotations of images as appropriate 
transformations. \citet{romero2021learning} start with a candidate symmetry group and focus on learning a subgroup of symmetry.
Another supervised approach to symmetry learning is via meta-learning. 
\mbox{\citet{zhou2020meta}} form many supervised tasks tailored to the symmetry learning problem
and use a weight-sharing architecture that is shared between tasks. A~cascaded optimization 
tries to improve the supervised performance by updating both the matrix that 
determines the weight-sharing and the weights that are used in each task. 
This way, the~model can discover convolutional weight-sharing from 1D
translation-invariant data, which is impressive. However, the~supervised tasks 
are a bit unnatural in that the synthetic data generation is conducted by
using sampled filters, and~the supervised task is to discover the 
ground truth filters. As~symmetries are essential ingredients in physical theories, a~wide range of approaches have been
tried for symmetry learning in physics-adjacent settings. \citet{craven2022machine} train a neural network (NN) to approximate a function, and~then 
candidate symmetries are tested by computing the values of the function on
inputs transformed by the candidate symmetry and considering the ``small transform''
asymptotic behavior of the error of the NN. \citet{forestano2023accelerated} and \citet{forestano2023deep} 
start with specific quadratic forms used in the
definition of various Lie groups (as the set of maps that preserve
the given form) and~then look for linear transformations near identity that 
leave the form invariant to recover the Lie algebra. 
\citet{krippendorf2020detecting} set up a classification problem for predicting 
the value of a potential function with some underlying symmetries and then
use the representations in the embedding layer to search for transformations
that relate points that are close to each other in this~layer.



\textbf{Self-supervised approaches} adopt auto-regressive setups for 
the symmetry learning problem. For~datasets involving sequential frames, \citet{sohl2010unsupervised} extract linear transformations relating subsequent
time steps in a self-supervised manner, whereas, in the setup proposed by \citet{dehmamy2021automatic}, one has access to original
data and a transformed version and aims to learn a Lie algebra generator that would relate the two by an approximate group action. 
In the setting of dynamical systems, \citet{greydanus2019hamiltonian} focus on 
learning a Hamiltonian, whereas \citet{alet2021noether} learn conserved quantities,
which are both related to underlying symmetries. The~self-supervised approach 
is most useful and interesting in the setting of dynamics of systems, but~this also limits its applicability. However, within~this setting, the~lack of a 
need for labeled data is an advantage.


\textbf{Unsupervised approaches} to symmetry discovery work with a raw dataset
and aim to output maps that represent symmetries of the dataset. 
A prominent line of research in this setting is based on Generative Adversarial Networks (GANs)
\cite{goodfellow2014generativeadversarialnetworks}.
\citet{desai2022symmetry} propose using a generator applying candidate symmetry transformations from a parametrized 
set, such as 
affine transformations, and~a discriminator
aiming to distinguish real samples from the transformed versions.
The training converges to parameter values that
represent symmetries of the dataset, but~this approach neither provides the generator of the relevant transformation
nor works with a high-dimensional setting without a low-dimensional parametrization for the set of candidate transformations.
\citet{yang2023generative} propose enhancements such as 
regularization terms preventing identity collapse and encouraging
the discovery of multiple Lie algebra generators and~sampling
the exponential coefficients of the generators from
distributions to enable the discovery of subgroups. The~main
setup is similar in that only low-dimensional parametrized 
generators are discovered. \mbox{\citet{yang2023latent}} 
extend this idea to learn nonlinear group actions by 
learning a representation where the group action becomes
linear simultaneously with the group generator. Except~for the autoencoder used for the representation, the~setup
is similar to the setup proposed by \citet{yang2023generative}.
\citet{tombs2022method} propose an approach that is similar in spirit to the GAN-based symmetry learning, where candidate symmetries
are not generated by the model but are provided externally, and~a model is trained to discriminate
between real and transformed samples. The~idea is that, if the model assigns similar probabilities for 
the sample being real or transformed, then one concludes that the candidate transformation
is indeed a symmetry. 

Instead of learning symmetry generators explicitly, one can search
for a symmetry-equivariant representation directly. \citet{anselmi2019symmetry} 
propose an unsupervised approach for learning such equivariant representations 
for the case of permutation groups. They demonstrate that the
proposed approach can learn equivariant representations for various
subgroups of the symmetric group of  six objects, acting on a six-dimensional dataset
via component permutations. The~generated synthetic datasets are completely
explicitly symmetric (created by the action of the full group on an initial 
set of samples).

Among the approaches described above, the~requirements for labeled datasets or 
structured data representations limit the applicability of supervised and self-supervised
symmetry learning paradigms. Unsupervised methods offer the widest applicability in principle, 
but they are demonstrated to work for only up to  
four-dimensional irreducible representations (and our experiments reported below 
imply that they perform poorly for higher dimensions). Real-world datasets often 
involve much higher dimensionalities, which state-of-the-art unsupervised methods cannot deal~with.

In this study, we formulate the symmetry learning problem in a representation 
learning framework where the symmetries are learned via a group convolution map.
This makes the symmetry manifest, turning it into a simple translation 
symmetry, which we believe is unique in the symmetry learning literature. 
Our setup is able to accurately extract symmetries from datasets with  
dimensionalities as high as 33 without any hard-wired factorizations, which 
is  by far the highest dimensionality we have seen in the unsupervised approach
to symmetry learning. Our approach based on coupling locality and invariance also 
allows the method to consistently learn the minimal generators, thus enabling one 
to construct the full symmetry~group.

\section{Theoretical Setting and Data~Model}\label{section:data_model}

In this section, we describe a simple setting for data-generating
mechanisms that have the symmetry and locality properties described in our introductory
discussion. This will both motivate our symmetry learning approach and guide the  
data generation for our experiments. Readers primarily interested in the 
description of the method and the results can skip this section in a first reading
and peruse Appendix \ref{app:synthetic_data} for information on our~datasets.

Our aim is to use the spatial information locality and the translation symmetry of images
for inspiration to describe the more general setting of symmetry and locality 
under a different \emph{group} of transformations. We first consider the ideal
case of continuous data (corresponding to images with ``infinite resolution'') and then
turn to a discretized version (corresponding to pixelated images). The~relevant 
translation symmetry group for image data is the group of 2-dimensional translations, 
but we focus on the simpler case of \mbox{1-dimensional} translations appropriate for
``1-dimensional images'' (or, more familiarly, time-series data).

\subsection{The Continuous~Setting}\label{section:cts-data-model}
\unskip

\subsubsection{Introduction}
A group is a set with an associative binary operation that has an identity element
and an inverse for each element. The~set of invertible transformations
for a wide class of mathematical objects is described
under the setting of group theory.
The group of \mbox{1-dimensional} translations is 
isomorphic to the additive group $\mathbb{R}$, each number 
representing the translation amount and the identity element 
corresponding to a translation by~0. 

As mentioned in our Introduction in Section~\ref{sec:introduction}, 
images consist of building blocks that are
spatially local, and~the approximate translation symmetry of an image dataset is borne
out of the random distribution of the locations of the building blocks (see
Figure~\ref{fig:group-acting-locally}). In~order to formalize the case of 1-dimensional
translations, we first model the data-generating mechanisms for ``1-dimensional images'' as a simple class of stochastic processes and then move on to more general~symmetries.

\subsubsection{Processes with Symmetry and Locality Under 1-Dimensional~Translations}
We will think of each 1-dimensional image as a sample function 
(or a realization) of an underlying stochastic process. Intuitively, a~real-valued 1-dimensional stochastic process is a ``machine'' that 
provides us a random function $f: \mathbb{R}\to\mathbb{R}$  each time we press a 
button, using an underlying distribution. A~translation-invariant (or \emph{stationary})
process is one where a sample function $f(t)$ and all its translated versions $f(t-\tau)$, 
$\tau\in\mathbb{R}$ are ``equally likely''. Of~course, it does not 
make sense to talk about 
the probability of a single sample function, and~rigorously defining 
probability measures on function spaces is rather tricky.
However, the~intuitive picture mentioned provides a useful 
viewpoint that we will utilize below. (In the literature on stochastic processes, a~stochastic process
is more properly defined in terms of a family of maps from a 
probability space, and~stationarity is 
commonly defined in terms of the joint distributions
of function values $f(t_j)$ on finite sets $\{t_j\}_{j=1}^{N}$. 
For a formal definition of stationary processes,
see \citet{doob1990stochastic}.)

To create a mathematical model for a translation-invariant process with local building blocks,
consider a set $\{\phi_i\}_{i\in I}$ of functions 
that are localized (compactly supported). We 
would like to think of 
each $\phi_i$ as representing a type of object. If~we shift each 
$\phi_i$ by an amount $\tau_i$ and add up the resulting functions, we 
obtain a signal $f(t)$ consisting of local building blocks at various
locations, $f(t) = \sum_i \phi_i(t - \tau_i)$. If~we could select each $\tau$ randomly using a uniform distribution, the~process of creating such $f(t)$ would result in a stationary stochastic~process. 

While it is not possible to have a uniform probability distribution over the whole real line, this is a technical difficulty that can be overcome by, e.g.,~considering a 
stochastic process on a ``large'' circle instead of $\mathbb{R}$
(or  using a stationary point process such as the Poisson process to 
obtain a collection of centers $\tau_{ik}$ for 
each $\phi_i$ with a translation-invariant distribution on 
$\mathbb{R}$ rather than a single center). 

More generally, each $\phi_i$ could have additional parameters $\lambda_{ik}$ such as width or amplitude, which one could independently sample from their own distributions. In~short, starting with a set of basis signals (``objects types'') $\phi_i$ and sampling the centers and other parameters
in an appropriate way obtains a stochastic process that is both 
symmetric (uniform distribution of centers) 
and local under translations. Sample functions of such a process are given by
\begin{equation}\label{translation-inv-local-signal}
f(t) = \sum_i \left[\sum_k \phi_i(t - \tau_{ik}; \lambda_{ik})\right].
\end{equation}

\begin{figure}[H]  
    \includegraphics[scale=0.175]{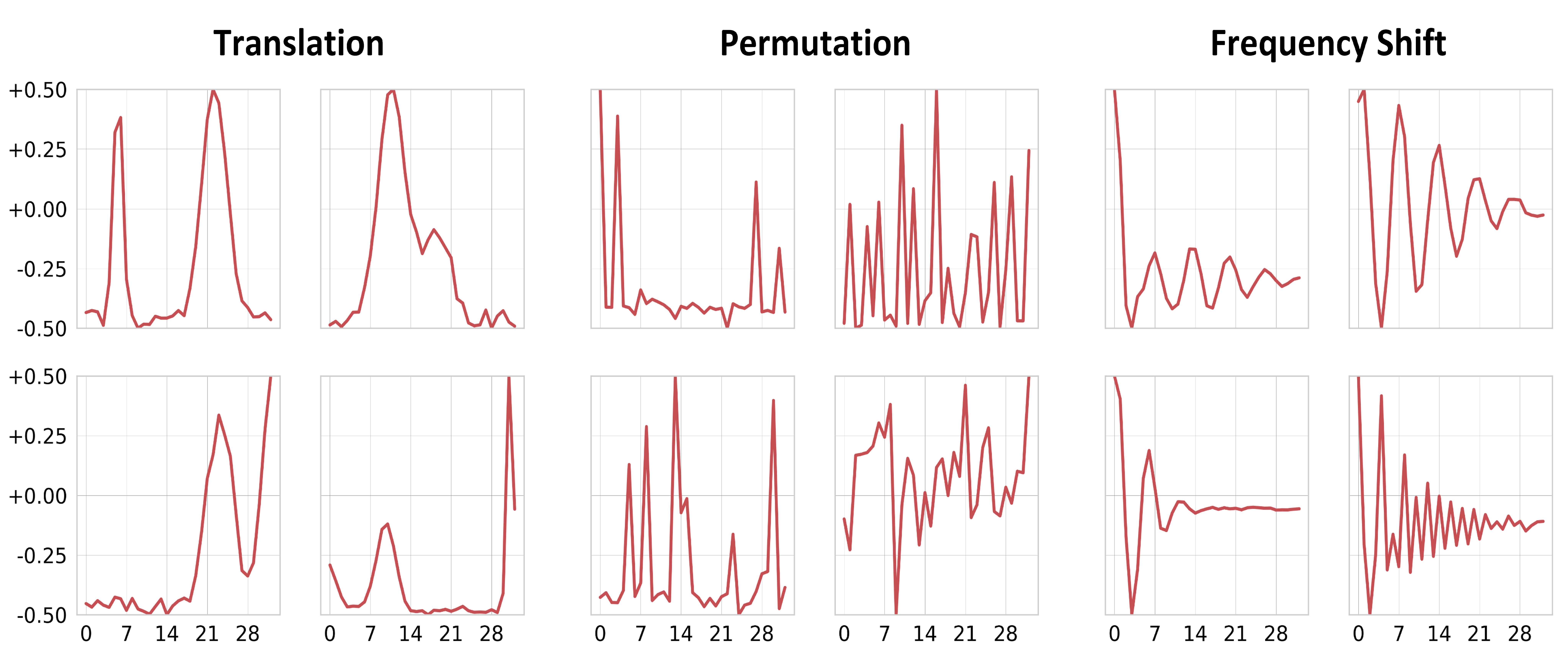}
    \caption{
      Samples 
 from the synthetic datasets used for model training. We show 4 samples from datasets generated from Gaussian basis signals. In~each figure, the~\emph{x}$-$axis represents the component index (the “pixel index”) of the sample vectors, and~the \emph{y}$-$axis is the amplitude of the corresponding component. The~group on the
     left has samples from the raw datasets, which are local and invariant under simple translations of
     components, the~second column uses datasets invariant under permuted translations, and~the third
     column datasets invariant under frequency shifts, i.e.,~shifts of components in Fourier (discrete sine transform I) space.}
    \label{fig:dataset-samples}
\end{figure}

To complete the analogy with (finite-size) images, one can crop 
each such sample $f(t)$ to a finite interval. (In real 
images, opaque objects actually hide each other rather than combining 
in an additive way, but~we ignore such considerations in this simple 
setting.) See the 
left-hand side of Figure~\ref{fig:dataset-samples} for samples of
the kind described here. 

\vspace{-6pt}

\subsubsection{Processes with Symmetry and Locality Under a General 1-Dimensional Group~Action}
We would like to generalize the simple setting above to more general 
group actions. We will describe an abstract setting first and will 
then provide concrete~examples.

Consider stochastic processes defined on a space
$\mathcal{X}$, with~the space of sample functions $x$ denoted by
$\mathcal{F}(\mathcal{X}) = \{x: \mathcal{X}\to\mathbb{R}\}$. Suppose 
we have an Abelian
group $G$ (group operation written as $+$)
with a given group action $\rho$ on 
$\mathcal{F} (\mathcal{X})$. In~other words,
for a sample function $x\in \mathcal{F}(\mathcal{X})$ and group element $\tau\in G$, 
the result of the action $\rho(\tau)\cdot x$ is another 
sample function, and~$\rho$ satisfies $\rho(0)=0$, 
$\rho(\tau_1)\cdot\rho(\tau_2) = \rho(\tau_1 + \tau_2)$,
$\rho(-\tau) = [\rho(\tau)]^{-1}$. 
We will specialize to the case $\mathcal{X} = \mathbb{R}$
and $G$ isomorphic to $\mathbb{R}$, but~much of what we write will apply to more general Abelian Lie groups. We will keep $\rho$ unspecified;
importantly, we do not assume that
$\rho$ is  given by the ``regular representation''. 
In particular, 
we do not assume $\rho$ acts by simple translations as in
$(\rho(\tau)\cdot x)(t) = x(t-\tau)$.

Given such an action $\rho$, we would like to
describe a process that generates sample functions
with invariance and locality properties as above,
but, this time, the~locality and invariance will be
under $\rho$ instead of translations. 
Taking $\rho$ to be a simple translation for the moment, 
the following representation of the signal $f(t)$ of 
\eqref{translation-inv-local-signal} motivates our generalization:
\begin{align}\label{eqn:general-basis-signal-from-delta}
   f(t) &= \int \delta(t-\tau) f(\tau) d\tau\\
        &= \int  (\rho(\tau)\cdot \delta^{(\rho)})(t) f(\tau) d\tau\label{eq:transforming-by-rho} \,.
\end{align}
Here 
, in~\eqref{eqn:general-basis-signal-from-delta},
$\delta(t)$ denotes the usual Dirac delta function,
and, in \eqref{eq:transforming-by-rho}, 
$\delta^{(\rho)}(t)$ is just another name for it, the~notation suggesting that this delta function is ``a delta function along the symmetry action $\rho$ of simple translations''.
The action $\rho(\tau)\cdot \delta^{(\rho)}$
by the translation operator $\rho(\tau)$ provides the shifted delta,
$\delta(t-\tau)$. 

If we now view the right-hand side of 
\eqref{eq:transforming-by-rho} as applying to a general $\rho$
instead of only translations, we obtain the generalization we want.
For a given group action $\rho$ for the group $G$ isomorphic to $\mathbb{R}$, assuming one has an appropriate 
notion of a ``delta function $\delta^{(\rho)}$ along $\rho$'',
one can create a process that is invariant (stationary)
and local under 
the group action $\rho$ from a process that is invariant and 
local  in the usual sense on $\mathbb{R}$. One obtains
samples $f(t)$ from a process such as $\eqref{translation-inv-local-signal}$ on $G$ and~transforms each sample to a sample $x(t)$ on 
$\mathcal{X}=\mathbb{R}$ via
\begin{equation}\label{eqn:signal-via-delta}
x(t) = \int  (\rho(\tau)\cdot \delta^{(\rho)})(t) f(\tau) d\tau \, .
\end{equation}
Here, $\rho$ no longer represents simple translations, and~$\delta^{(\rho)}$
is no longer the usual delta function but~is an appropriate delta 
``function'' (or measure, or~distribution) defined on 
$\mathcal{X}$. (Here, the~integration measure
$d\tau$ on the right-hand side should properly be viewed as the Haar 
measure on the relevant Lie group, but, in this 1-dimensional setting, we do not lose much by treating it as the Lebesgue measure on 
$\mathbb{R}$; for higher-dimensional, possibly non-Abelian Lie groups, 
one will have to be more careful.) While we specialize to $\mathcal{X}=\mathbb{R}$ here (and mention a generalization
to $\mathbb{R}^D$ below), the~formalism suggests greater~generality.

To provide another concrete example of this abstract setting (in addition to 
translations), consider the group action of $G\cong \mathbb{R}$ on sample 
functions  given by frequency shifts instead of translations. In~other 
words, let $\rho(s)$ act on a sample function $x(t)$ 
via 
\begin{equation}
   (\rho(s) \cdot x)(t) = e^{-i s t}x(t)
\end{equation}
which corresponds to a shift in Fourier space (for simplicity, we take the sample functions in this case to be complex-valued). For~this action, an~
appropriate delta function along $\rho$ is a Fourier basis function
$\delta^{(\rho)}_{k_0}(t) = e^{ik_0 t}$, where $k_0\in\mathbb{R}$. Just as 
the regular Dirac delta function is sharply localized under translations in 
the sense that $\delta(t-t_0)$
and $\delta(t-t_0-\tau)$ have zero overlap, the~functions $\delta^{(\rho)}_{k_0}
(x) = e^{ik_0t}$
and $(\rho(s)\cdot\delta^{(\rho)}_{k_0})(t) = e^{i(k_0-s)t}$ have zero overlap (e.g., these functions are orthogonal by Fourier analysis).
Specializing to the case $k_0=0$, 
we can construct signals that are local along the 
action of $\rho$ by using \eqref{eqn:general-basis-signal-from-delta} from a signal $f(t)$ as in \eqref{translation-inv-local-signal} 
defined on the group $G=\mathbb{R}$:
\begin{equation}
  x(t)  := \int  (\rho(s)\cdot \delta_{\rho})(t) f(s) ds = \int e^{-ist}f(s) ds \, .
\end{equation}

Thus, we obtain the pleasing result that, according to this setup, a~signal that is local along the ``frequency shift'' action $\rho$ of the 1-dimensional group $G\cong\mathbb{R}$ is given
by the Fourier transform of a signal $\Phi_0(s)$ that is local under translations (of $s$).
In other words, to~obtain a stochastic process that is invariant and local under 
frequency shifts, we can start with a spatially local and 
spatially stationary process on the group $\mathbb{R}$
and take the Fourier transform of each~sample. 

We will leave an attempt at a rigorous description of the fully general version of this setup for future work and~note that, in the general case, the~delta function 
$\delta^{(\rho)}$ on $X$ is closely related
to what is called an ``approximate identity'' in the abstract harmonic 
analysis literature (see \citet{folland2016course} for an introduction). In~
particular, such a delta will be required to satisfy
\begin{equation}
  (\rho(s)\cdot\delta^{(\rho)})*(\rho(t)\cdot\delta^{(\rho)}) = \delta(s-t)
\end{equation}
where $*$ denotes the appropriate convolution on $\mathcal{X}$,
and the $\delta$ on the RHS is the Dirac delta 
function on the group $G\cong\mathbb{R}$. This means, in~particular, that
separately transformed versions of $\delta^{(\rho)}(x)$ have zero overlap
unless the transform parameters are the~same. 


\subsection{The Discrete Setting: Synthetic Data~Generation}\label{section:discrete-data-model}
\unskip

\subsubsection{Discrete Translation~Symmetry}
To create discrete signals that are local and symmetric under 1-dimensional discrete translations, we follow the procedure of creating a sample out of local basis signals
as in \eqref{translation-inv-local-signal} and~replace $t$ with a discrete index 
$n$. The~local basis signals $\phi_i(t)$ and sample functions $x(t)$ become vectors with 
components $\phi_{in}$ and $x_n$, respectively. 

In principle, the~component index $n$ ranges over all integers, but, for data generation purposes, it is restricted to a finite range. 
The basis signals $\phi_{in}$ we use are Gaussians (parametrized by width and amplitude)
and Legendre functions (parametrized by width, amplitude, and~order), with~the
index of the center being sampled uniformly over a finite range. The~details 
of the process 
are given in Appendix \ref{app:synthetic_data}, but, as a quick summary, we use
finite-dimensional vectors for each sample vector and use uniform distributions 
for centers, widths, and~amplitudes of local basis signals, including superposition of a few such signals per sample. Finally, we add Gaussian noise on top of each sample for more realistic experimentation. Examples of the resulting samples can be seen in Figure~\ref{fig:dataset-samples} under the title of~``Translation''.

\subsubsection{General Representations: Behavior Under Orthogonal/Unitary~Transformations}\label{app:input_orthonormal_transformation}

Starting with a dataset that is symmetric and local under discrete 
translations, we can create datasets that are symmetric/local under
different group representations by using a discrete version of the 
transformation \eqref{eqn:signal-via-delta}. Here, we start by 
proving a transformation property of discrete versions of
the symmetry generators and the ``Dirac delta along group action'' 
described above and~then provide the procedures used in generating 
synthetic~data.

The discrete version $\mathbb{Z}$ of the translation group 
$G\cong\mathbb{R}$ has a minimal generator that is the pixel (or 
component) translation operator, which we denote by 
$1\in\mathbb{Z}$. Similarly, for~any representation of the 
1-dimensional group $G\cong\mathbb{R}$, there will be a matrix for the
minimal generator $\rho(1)$, which we call $\gen$.
With this notation, 
the discrete version of \eqref{eqn:signal-via-delta} is
\begin{align}
    x_n = \sum_{p} \sum_j \gen^p_{nj} \updelta^{(\rho)}_j f_p \, . 
\end{align}
For samples $f_m$ ($m\in \mathbb{Z}$)
with a distribution local and invariant under component shifts, 
this provides a sample distribution for $x_{n}$ that is local and invariant under the action of the generator $\gen$.

Now, let us consider the action of an invertible matrix 
$\matr{Q}$ on both sides. We obtain
\begin{align}
    \sum_n Q_{mn} x_n  &= \sum_{m, j, p} Q_{mn} \gen^p_{nj} \updelta^{(\rho)}_j f_p = \sum_{m, k, p, k, l} Q_{mn} \gen^p_{nj} Q^{-1}_{jk} Q_{kl} \updelta^{(\rho)}_l f_p \\
    x'_n &= \sum_{l} \gen'^p_{nl} \updelta^{(\rho')}_l f_p
\end{align}
where primes denote the transformed version of the objects under $\matr{Q}$.
Thus, the~transformed signal $\mathbf{x}'$ has the symmetry/locality generator
$\gen'$ given in terms of the old one via a similarity transformation 
$\gen' = Q \cdot\gen \cdot Q^{-1}$, and~the new ``delta along the 
symmetry'' $\boldsymbol{\updelta^{(\rho')}}$ is given in terms of the old one via \(\boldsymbol{\updelta^{(\rho')}} = Q \cdot \boldsymbol{\updelta^{(\rho)}}\).
In other words, starting with symmetry/locality under a given group representation $\rho$ with generator $\gen$, we obtain symmetry/locality under $\rho'$ with generator $\gen'$ by using a transformation. If~the transformation matrix $\matr{Q}$ is orthogonal (resp. unitary), then the new generator $\gen'$ will also be orthogonal (resp. unitary) assuming the old one $\gen$ is~so.

To create synthetic data in the case of 1-dimensional symmetry groups described above, 
we consider three symmetry group actions in our experiments: 
\textbf{translations}, \textbf{permuted translations}, and~\textbf{frequency shifts}. Data samples for each symmetry can be 
found in Figure~\ref{fig:dataset-samples}. In~each case, we start with a 
$33$-dimensional dataset that is symmetric/local under component 
translations $\gen = T$ with
$(T\mathbf{x})_i = x_{i-1}$ (ignoring 
edge effects; one could use cyclic translations to get rid of 
them) and apply an appropriate $Q$.
For translations, we take $Q=I$.
For frequency shifts, we take $Q = D^{-1}$, where $D$ 
is the discrete sine transform matrix of type I. This results
in the generator $\gen = D^{-1}\cdot T\cdot D$, which
shifts the frequency components of the samples. 
For permuted translations, take $Q=P$, where $P$ is a
permutation matrix. Applying $P$ to each raw sample,
we obtain samples that are symmetric and local under
the powers of the permutation operator $\gen = P\cdot T\cdot P^{-1}$.
As can be seen in  Figure~\ref{fig:dataset-samples}, 
the symmetry/locality for the permutation and frequency shift
cases is not apparent to the naked eye at~all.

\section{Materials and~Methods}\label{sec:materials_and_methods}

In this section, we formulate the learning problem and the loss function 
and discuss the training loop together with some useful optimization 
techniques. See Appendix \ref{app:complexity-analysis} for information on
the computational complexity of our~method.

\subsection{The Setup: Objects to~Learn}
Our model will start with a dataset \(\mathcal{D} = \{\mathbf{x}_i\}^n_{i=1}\), where \(\mathbf{x} \in \mathbb{R}^d\) and $n$ is the number of samples, 
and will learn an underlying symmetry group representation $\rho(s)$ and a filter $\boldsymbol{\filter}$ that is local along the symmetry action in the sense suggested in our Introduction, 
which will be conducted precisely by the loss function below. An~appropriate combination of these building blocks will provide
a symmetry-based representation given by a matrix $L$,
which we call the \textbf{group convolution matrix}.

\subsubsection{Symmetry~Generator} We will assume that the symmetry action $\rho(s)$ is a real unitary representation; in~other words, for~each $s$, 
$\rho(s)$ will be a $d\times d$ orthogonal matrix, $\rho(s)\in 
O(d)$. We will focus on symmetry actions that can be 
continuously connected to the identity, which restricts the actions
of group elements to those with determinant 1, i.e.,~
elements of $SO(d)$. While the underlying symmetry action $\rho(s)$ 
will be the representation of a 1-dimensional Lie group parametrized by the 
continuous parameter $s$, the~model will learn a minimal 
discrete \textbf{generator} $\gen$ of this action appropriate for 
the dataset at hand. With~an appropriate choice of the scale of the 
$s$ parameter that parametrizes the group, we can write $\gen = 
\rho(s=1)$, which provides (for any integer $s$) $\rho(s) = \rho(1)^s 
= \gen^s$. This will allow us to obtain the action of any group element by taking an appropriate power of the~generator.

Any element of $SO(d)$ can be obtained by using the exponential map 
on an element of the Lie algebra of $SO(d)$, which is the algebra 
of $d\times d$ antisymmetric 
matrices~\citep{helgason1979differential}. In~particular, our 
generator $\gen$ should be given as the matrix exponential of an 
antisymmetric matrix. Instead of explicitly constraining $\gen$ to 
be an element of $SO(d)$, we write this matrix in terms of an arbitrary $d\times d$ matrix $A$ via \(\gen = \operatorname{exp}\left(\frac{A - A^T}{2}\right)\),
where the matrix exponential can be defined in terms of the Taylor 
series. The~optimizer will thus seek an appropriate $A$ instead of a direct search for $\gen$. (To optimize memory usage, one could use a more 
efficient parametrization of antisymmetric matrices.)

\subsubsection{The Resolving~Filter}
We will assume that the resolving filter $\boldsymbol{\filter}$
acts by an inner product, i.e.,~$\boldsymbol{\filter}$ itself will be given as a $d$-dimensional (column) vector. This will be closely related to the discrete analog of the delta function $\delta^{(\rho)}$ described in Section~\ref{section:data_model}.
We do not use any constraints on this $d$-dimensional vector; 
however, we use a normalized version of it in our computations of
the relevant dot~products.

\subsubsection{The Group Convolution~Matrix}
Following the approach sketched in the Introduction,
we will formulate the learning problem in terms of a symmetry-based 
representation \(\mathbf{y} \in \mathbb{R}^d\) of the data $\mathbf{x} \in \mathbb{R}^d$ 
given in terms of the generator $\gen$ and the filter $\boldsymbol{\filter}$. For~each $d$-dimensional 
sample vector $\mathbf{x}$, the~components $y_p$ of the symmetry-based 
representation $\mathbf{y}$ will consist of the application of the filter $\boldsymbol{\filter}$ to transformed versions $\gen^p \mathbf{x}$
of the data, $y_p = \sum_{ij} \filter_i (\gen^T)^p_{ij} x_j$, where
\linebreak  $p=0, \pm 1, \pm 2, \ldots$. Note that this is completely analogous to a CNN, in~which case the generator $\gen$ is a pixel translation operator and~$\boldsymbol{\filter}$ is a local CNN~filter. 

These scalar representations $y_p$ of a sample will be the fundamental 
quantities on which we define the loss function measuring
departures from stationarity and locality. In~the $\mathbf{y}$ representation, the~action of the  generator $\gen$ is represented by a simple shift in $p$, 
and thus, if~$\gen$ is indeed a symmetry generator, the~joint 
distribution of the scalars $y_p$ should be 
invariant under shifts of $p$: $p\mapsto p + n$, which is 
a simple translation symmetry. Similarly, if~data are local along the action of $\gen$ and~$\filter$ is indeed local along the symmetry direction,
$y_p$ and $y_{p'}$ should be similar/strongly dependent 
as random variables if $p$ and $p'$ are close and~should be 
approximately independent otherwise (locality). Thus, if~the distribution of
$\mathbf{x}$ is invariant and local under actions of $\gen^s$, the~
distribution of $\mathbf{y}$ should be \emph{invariant and local under simple 
shifts in the components} $y_p$.

While there is no a priori restriction on the
range of powers $p$ to use in the representation, in~this paper,
we choose the dimensionality of the representation 
$\mathbf{y}$ to be the same as the dimensionality of 
$\mathbf{x}$. (Our methodology does not rely on this choice, and~one 
could easily consider other ranges for $p$.)

To obtain $\mathbf{y}$ from $\mathbf{x}$ directly via $\mathbf{y} = L\mathbf{x}$, we form 
the matrix $L$, which we call the group convolution matrix:
\begin{equation}\label{eqn:L-group-convolution}
L = 
\left[
  \begin{array}{ccccc}
    \horzbar \text{ } \boldsymbol{\psi}^T (\gen^{P_{min}})^T \text{ }   \horzbar \\
    \horzbar \boldsymbol{\psi}^T (\gen^{P_{min}+1})^T   \horzbar \\
    \cdot \\
    \cdot \\
    \cdot \\
    \horzbar \text{ } \boldsymbol{\psi}^T (\gen^{P_{max}})^T \text{ } \horzbar \\
  \end{array}
\right]
\end{equation}
where $P_{min}$ and $P_{max}$ denote the minimum and maximum powers
of the generator to be used. In~our experiments, we work with odd 
$d$ and~pick $P_{max} = \frac{d-1}{2} = -P_{min}$.

\subsection{The Loss~Function}
Our loss function will be computed from the transformed version 
$\mathbf{y} = L \mathbf{x}$ of \emph{each batch} of vectors \(\mathbf{x}\) and~
will measure the degree to which the distribution of this transformed 
version is symmetric and local under component translations 
$y_p \mapsto y_{p+s}$. The~loss consists of three building blocks we call
\emph{stationarity}, \emph{locality}, and~\emph{information preservation}.
These are given in terms of the correlation between the components $y_p$
of the $\mathbf{y}$ representation, as~well as various entropy and 
probability density terms for these components. We first  
describe the pieces of the loss function assuming one has
access to estimates of these necessary quantities and~will then explain 
the techniques used for estimating these quantities for each~batch. 

Note: For notational simplicity below, we use a non-negative
indexing for the components $y_p$ of $\mathbf{y}$, i.e.,~$p=0, 1, \ldots, d-1$, and~use the notation $\langle\boldsymbol{\cdots}\rangle_{\text{[conditions]}}$ for
the average of the expression $\boldsymbol{\cdots}$ in the angle brackets over index combinations satisfying the~conditions.

\subsubsection{Stationarity/Uniformity}\label{sec:stationarity-uniformity}
For true symmetry, we expect the joint probability distribution of the components of $\mathbf{Y}$ to be invariant under simple shifts in components
(modulo boundary effects). Since estimating the joint probability density is impractical in high dimensions, we instead use the distributions of individual components and conditional distributions for pairs of~components. 

We denote the marginal probability density of the random variable \(Y_j\) as \(p_j\) and~the conditional probability density of $y_j$ given $y_i$ by \(p_{j|i}\). Our uniformity loss, in~terms of these quantities, is given by
\begin{equation}
\mathcal{L}_{uniformity} = \frac{1}{2}\left[\left\langle \hat{D}_{KL}(p_{m}, p_{n}) \right\rangle_{\substack{m=n \mp 1}} + \left\langle\hat{D}_{KL}(p_{i|j}, p_{k|l}) \right\rangle_{\substack{k=i + \Delta\\l=j + \Delta\\ \Delta = \mp 1}}\right]
\end{equation}\label{eqn:uniformity-1d}
where \(\hat{D}_{KL}\) denotes the estimate of the KL divergence (see Appendix \ref{app:kl-divergence-estimation} for the estimation procedure). Averages are taken over all possible \(m, n\), and \(i, j, k, l\) indices satisfying the specified constraints. The~marginal probability term provides a first-order proxy for uniformity, and~the conditional probability term takes relations between pairs of random variables into account. Overall, 
this loss function provides a second-order computationally tractable measure of the uniformity 
of the full distribution under shifts.

\subsubsection{Locality}\label{sec:locality-loss}
Our locality loss consists of two pieces that we call alignment 
and resolution. Alignment loss aims to make neighboring pixels 
have similar information, and~resolution aims to make faraway pixels have distinct~information.

\textbf{Alignment}\label{sec:alignment-loss}
We would like successive components of symmetry-based representation to not only have similar distributions (as in the uniformity loss) but to have similar 
values in each sample. We compute the \emph{sample Pearson correlation coefficient} \(\rho_{i, i+1} \in \mathbb{R}\) of the components $y_i$ and $y_{i+1}$ of symmetry-based representation $\mathbf{y}$ for each batch and~define the alignment loss \(\mathcal{L}_{alignment}\) as the average of this quantity over all successive pairs
\begin{equation}
    \mathcal{L}_{alignment} = 
    -\left\langle \rho_{i, i+1} \right\rangle_{i \in [0, d-1)} \, .
\end{equation}\label{eqn:alignment-loss}

\textbf{Resolution}
We would like the components $y_i$ and $y_j$ of $\mathbf{y}$ to contain 
distinct information when $i$ and $j$ are not close. 
In the information theory literature, the~quantity
\begin{align}
    \operatorname{C}(\mathbf{Y}) = \sum_{i=0}^{d-1} h(Y_i) - h (\mathbf{Y})
\end{align}
called the \emph{total correlation} measures the
degree to which there is shared information between the components
of random vector $\mathbf{Y}$; ($h(Z)$ denotes the differential entropy of the random variable $Z$). For~a local symmetry, only nearby components 
should share information, and~all other pairs of components should be 
approximately independent, so we expect $C(\mathbf{Y})$ to be small
and use an approximation to $\operatorname{C}(\mathbf{Y})$ over each batch as 
our resolution~loss. 

As we describe in Section~\ref{par:controlling_joint_entropy_rank},
we compute entropy estimates by using an epoch-dependent rank $k$.
In effect, this increases the number of variables that comprise the entropy estimate from $k=1$ to $k=d$ over the course of the estimation and 
thus changes the overall scale of the estimated quantity accordingly.
To use a normalized scale, we define the resolution loss for 
each batch  to be
\begin{align}
    \mathcal{L}_{resolution} = \frac{1}{d}\sum_{i=0}^{d-1} h(Y_i) - \bar{h}_k(\mathbf{Y})
\end{align}
where $\bar{h}_k(\mathbf{Y})$ is an estimate of the ``per rank'' contribution 
to entropy (see Appendix \ref{app:low-rank-entropy-estimation} for exact formula).

\subsubsection{Information~Preservation}

While the correct directions in parameter space reduce the locality and 
uniformity loss terms described above, there is also a catastrophic solution 
for $L$ that can minimize these terms: a constant map to a single point. 
Indeed, we would like to learn a transformation that preserves
the information content of the data, and~we add a final term  
to maximize the mutual information between the random vectors corresponding to input $\mathbf{X}$ and the output $\mathbf{Y}$. For~a deterministic map, this mutual information can be maximized by simply maximizing the entropy of the output (see the~InfoMax principle proposed by \citet{bell1995information}). 

As mentioned above, our entropy estimate  will use an epoch-dependent rank $k$, 
and we use a normalized per-rank version $\bar{h}_k(\mathbf{Y})$ 
(see Appendix \ref{app:low-rank-entropy-estimation}) of 
the output entropy as our information preservation loss:
\begin{align}
    L_{preservation} = -\bar{h}_k(\mathbf{Y}) \, .
\end{align}

\subsubsection{Total~Loss}
Merging the overall loss terms, we have
\begin{align}\label{total-loss}
\mathcal{L} = \mathcal{L}_{alignment} + \alpha \mathcal{L}_{resolution} + \beta\mathcal{L}_{uniformity} + \gamma \mathcal{L}_{preservation}
\end{align}
while \(\alpha, \beta, \gamma \in \mathbb{R}\). Limited experimentation
with only one dataset was able to provide a choice for these 
hyperparameters that ended up working well for all our experiments, making
it unnecessary to perform separate tunings for datasets with different dimensionalities
and symmetry properties.  See Appendix \ref{app:hyperparameters} for a discussion of hyperparameter 
choice and sensitivity and~ablation experiments confirming that all 
pieces of this loss function indeed contribute to model performance.

\subsubsection{On the Coupled Effect of the Alignment and Resolution~Terms}

In this section, we argue that the combined effect of the alignment
and resolution terms in the loss function push the symmetry-based
(output) representation towards a (two-sided) Markov~process. 

Let $\mathbf{Y}$ denote the $d$-dimensional output representation learned by the model.
Using the chain rule for entropy, the~joint entropy of $\mathbf{Y}$ can be 
written as $h(\rvector{Y}) = \sum_{i=1}^d h(Y_i|Y_{<i})$,
where \(\rvcomp{Y}{<i}\) denotes the components
\(\{\rvcomp{Y}{1}, \rvcomp{Y}{2}, \dots \rvcomp{Y}{i-1}\}\) 
when \(i>1\), and, for $i=1$, it simply means no conditioning is 
applied (in this section, we use non-negative integers for indexing components). 
We define the following $m$th-order approximation to the chain rule formula,
which only uses local dependencies of window size $m$:
\begin{align}
    h^{(m)}(\rvector{Y}) = \sum_i h(Y_i|Y_{[i-1, i-m]})
\end{align}
where \(Y_{[i-1, i-m]}\) denotes the random variables 
$\{Y_j\}_{i-m\le j \le m-1}$, and~\(m=0\) represents the case with no~conditioning. 

Since conditioning can only reduce entropy \citep{cover1999elements},
we have
\begin{align}\label{entropy-chain}
    h(\rvector{Y}) \leq h^{(d-1)}(\rvector{Y}) \leq \dots \leq h^{(1)}(\rvector{Y}) \leq h^{(0)}(\rvector{Y})\,.
\end{align}
The ``resolution'' (total correlation) part of the
loss \(\operatorname{C}(\mathbf{Y}) = h^{(0)}(\rvector{Y}) - h(\rvector{Y})\)
is minimized when all the conditional entropies 
in this chain are equal, which happens
when the distribution of the variables
is just the product distribution, i.e.,~
when the $Y_i$ are all independent. However, 
the ``alignment'' loss (see Section~\ref{sec:alignment-loss})
aims to maximize the correlation between successive 
components. Since independent variables have zero
correlation, the~alignment loss pushes successive 
components to be dependent. The~mutual information $I(Y_{i}, Y_{i+1}) = H(Y_{i+1}) - H(Y_{i+1}|Y_{i})$
between dependent variables is always positive, so
the tendency of the alignment loss
is to make $H(Y_{i+1})$ strictly greater than
$H(Y_{i+1}|Y_{i})$. 

In short, the~joint effect of the alignment and the
resolution terms is to push the $h^{(1)}(\rvector{Y})$
term in the chain \eqref{entropy-chain} to be strictly 
less than $h^{(0)}(\rvector{Y})$ and
all the other terms $h^{(j)}(\rvector{Y})$ and $j\ge 1$ to be
equal to each other, i.e.,
\begin{equation}\label{entropy-chain-optimal}
h(\rvector{Y}) = h^{(d-1)}(\rvector{Y}) = 
\dots = h^{(1)}(\rvector{Y}) < h^{(0)}(\rvector{Y})
\end{equation}
which means that the learned representation $Y_j$ is a
Markov process. Repeating the argument in the other direction,
we obtain a two-sided Markov process.
Of course, the~degree to which \eqref{entropy-chain-optimal} 
can, in~fact, be satisfied in a given problem is determined by the 
underlying data~distribution. 

\subsection{Training the~Model}
\begin{figure}[!ht]
    \includegraphics[scale=0.16]{figures/group_convolutation_layer_training.png}
    \caption{The 
 optimization loop for the group convolution matrix (symmetry--locality search).}
    \label{fig:training-loop-gcl}
\end{figure}

\begin{figure}[!ht]
    \includegraphics[scale=0.16]{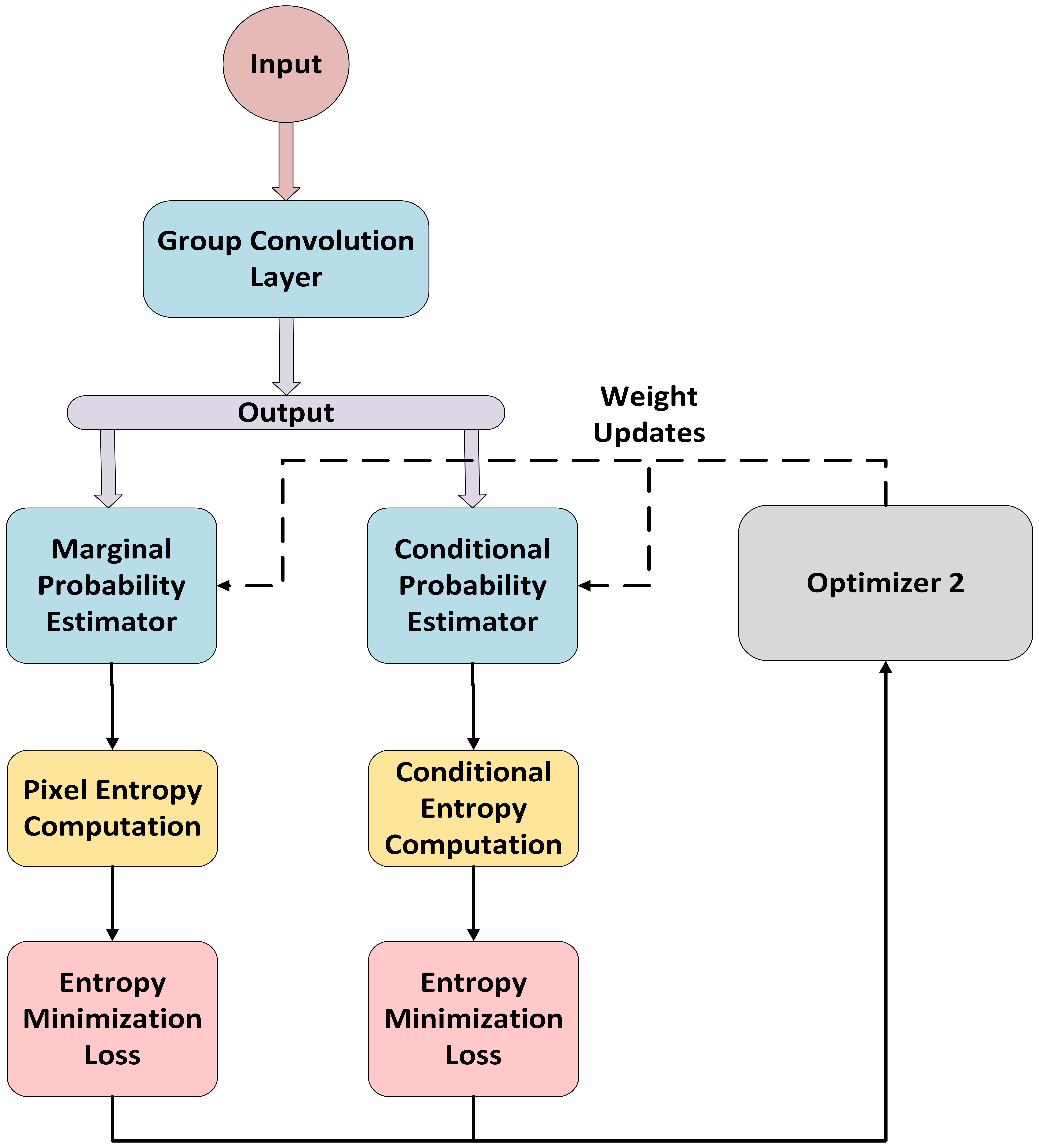}
    \caption{The 
 optimization loop for the probability density~estimators.}
    \label{fig:training-loop-estimators}
\end{figure}

Our system is composed of two separate subsystems and training loops. One subsystem seeks the generator $\gen$ of symmetries and the resolving 
filter $\filter$, which together form the group convolution matrix $L$,
while the other consists of probability estimators. For~each batch, densities 
and entropies of the data transformed by the group convolution matrix 
are estimated and used in the computation of the loss for symmetry 
learning. The~training for the two subsystems is conducted jointly, 
with simultaneous gradient updates. See Figures \ref{fig:training-loop-gcl} and \ref{fig:training-loop-estimators}
for a visual representation of the two subsystems, and 
see Appendix \ref{app:training-details} for the details of the training
process.

\subsubsection{Training the Group Convolution~Layer}\label{subsection:training}

We used the ADAM optimizer with learning rate decay for the group convolution layer, while the other parameters were kept at their default values. Since the higher powers computed for the relevant matrices are very sensitive to numerical errors, we preferred smaller learning rates. Numerical values and details can be found in Appendix \ref{app:training-details}. 

Due to the high dimensionality of the search space of the relevant matrices, accurate identification of true symmetries is challenging. The~special initialization and optimization procedures described below 
help to avoid becoming stuck at the local~minimum. 

\textbf{Controlling the rank of the joint entropy.}\label{par:controlling_joint_entropy_rank}
To achieve efficient and stable optimization by first capturing the data's gross features and then refining them over time, we use a time-dependent rank parameter \(k\) for the entropy estimator \eqref{entropy-rank-weights}. 
To adjust $k$ during training, we use a normalized notion of training time 
$t_n$, measuring the ``amount of gradient flow'' via
\begin{equation}
  t_n = \frac{\sum_{s=1}^{n} \text{lr}(s)}{\sum_{s=1}^{T} \text{lr}(s)}
\end{equation}
where $\text{lr}(s)$ 
 is the learning rate used at the training step (batch) $s$, 
and $n$ and $T$ are the current training step and~the total 
number of training steps, respectively. 
We control the rank $k$ of the low-rank entropy estimator by setting
$k = \text{ceil}(d \times t_n)$
so that, by the end of the training, the~rank is at $d$.

\textbf{Noise injection to the resolution filter.}\label{par:noise_injection}
We initialize the resolving filter with zeros and add Gaussian noise during the early stages of training before computing the loss for each batch, 
\begin{equation}
    \boldsymbol{\psi} \leftarrow \boldsymbol{\psi} + \mathcal{N}(\mu=0, \sigma = 0.1)  \exp(-t/\tau) \, .
\end{equation}
The amplitude of the noise is set to decay exponentially 
with a short time constant (of $\tau = 10$ epochs). 
As mentioned previously, to~compute the transformed data $\mathbf{y}$, 
we use a normalized version of $\filter$ at each step:
$\boldsymbol{\psi} \leftarrow {\boldsymbol{\psi}}/{\|\boldsymbol{\psi}\|_2}$

\textbf{Initialization of the symmetry generator.}\label{par:generator_initialization}
We initialize the matrix $A$ used to obtain the symmetry generator $\gen$
via the exponential map \(\exp{(\frac{A - A^T}{2})}\)
using a normal distribution with \(\sigma=10^{-3}\) and \(\mu=0\)
for its entries. This leads to an approximate identity matrix
for the initial generator $\gen$.
Using smaller standard deviations did not affect the performance; 
however, significantly larger $\sigma$ values occasionally lead to 
learning a higher power of the underlying symmetry generator instead of 
learning the minimal~generator.

\textbf{Padding.}\label{par:padding}
We use padding for the symmetry generator
and the filter in the sense that the symmetry matrix and the filter have
dimensionality that is higher than the dimensionality of the data, but~we
centrally crop the matrix and the filter before applying them to the
data. This is conducted to deal with finite size (edge) effects, and, after experimenting
with padding sizes of 6 to 33, we saw that the results are not sensitive
to padding size. Working with a cyclic/periodic symmetry would make the 
padding unnecessary, but~this would mean working with a restrictive
assumption on the underlying~symmetry.

\subsubsection{Training Probability Density~Estimators}

We train probability density estimators based on entropy minimization loss term, which is proposed by \citet{pichler2022differential}. See Appendix \ref{app:hyperparameters} for the learning rate
used.

\section{Results}\label{section:results}
\unskip

\subsection{Results on Synthetic and Real~Data}

For each dataset listed in Table~\ref{tab:datasets}, 
our model was trained to learn the group action generator $\gen$ and the resolving filter $\filter$, 
which together comprise the group convolution matrix, $L$  \label{eqn:L-group-
convolution}, which provides a symmetry-based representation of data. 
The group generator $\gen$ has a known correct answer in each case (up to a power of $\pm 1$; e.g., both right and left translations are minimal generators of the translation group), which, of~course, was not given to the model in any~way.

\begin{figure}[H]
    \subfloat[\centering \label{fig:translation_generator}]{
        \includegraphics[width=0.48\textwidth]{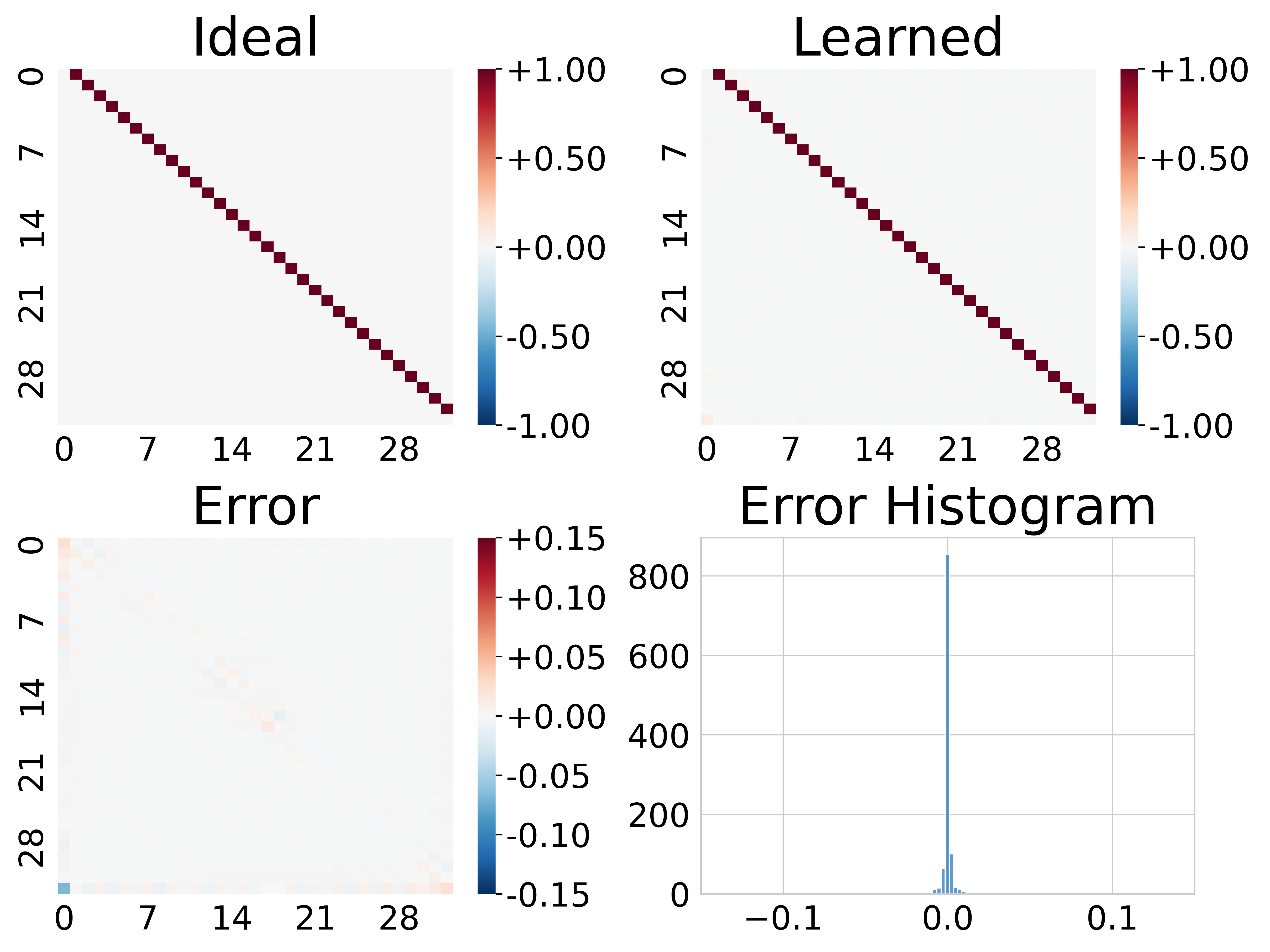}
    }
    \hfill
    \subfloat[\centering \label{fig:translation_group_convolution_matrix}]{
        \includegraphics[width=0.48\textwidth]{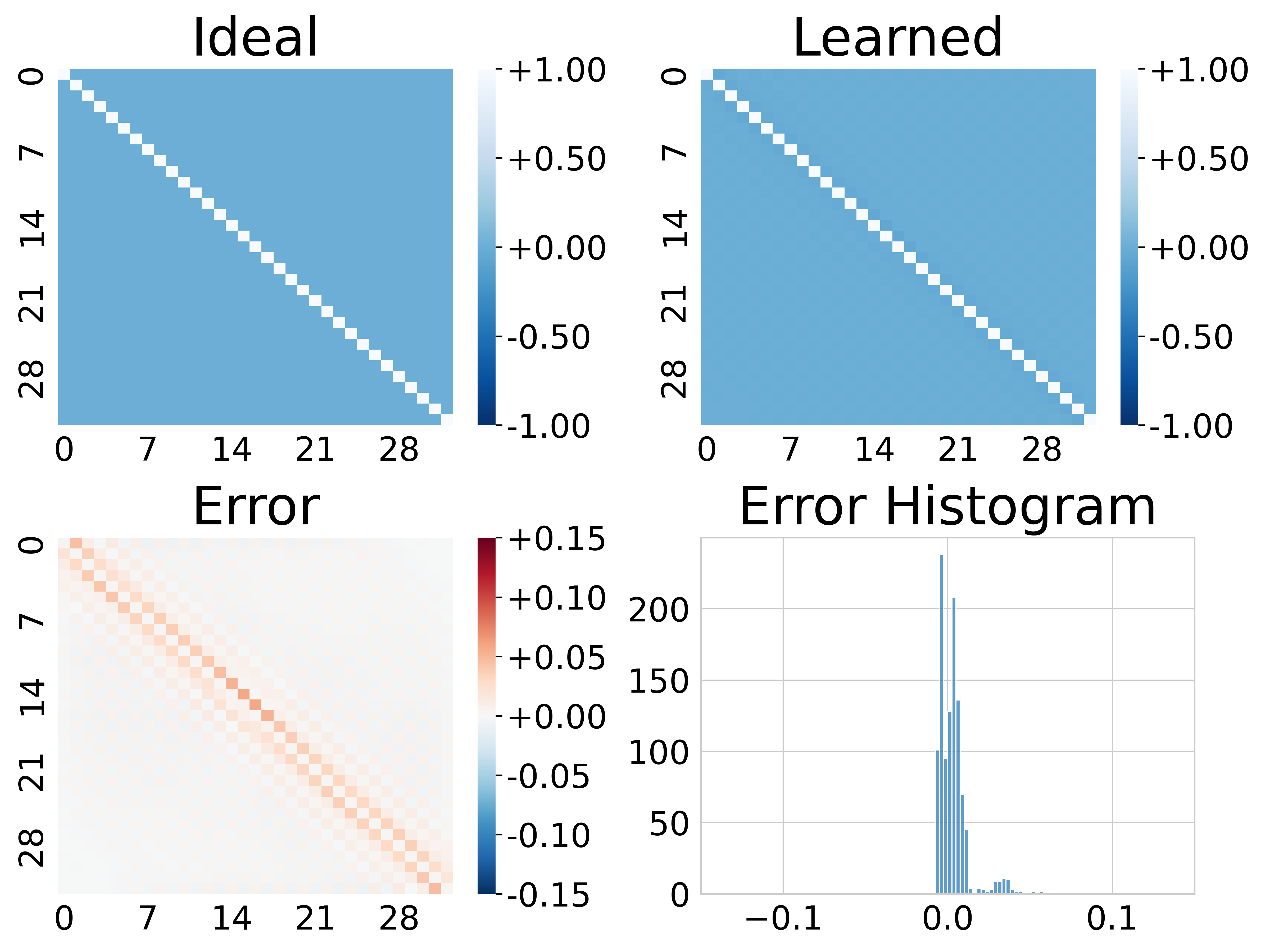}
    }
    \caption{The 
 symmetry generator for the translation-invariant dataset is the 1$-$step
      translation operator, which is simply a shift matrix (with entries just below or 
      above the diagonal equaling 1). On~the left, we see that this matrix is learned with 
      high accuracy. The~group convolution matrix that provides the symmetry$-$based representation 
      for translation symmetry is the identity operator. On~the right, this operator, formed by 
      combining the powers of the group generator with the learned resolving filter, is also 
      learned to a high degree of accuracy
. (\textbf{a}) Ideal and learned symmetry generators (\textbf{top}) and error distributions (\textbf{bottom}). (\textbf{b}) Ideal and learned group convolution matrices (\textbf{top}) and error distributions (\textbf{bottom}).}
    \label{fig7}
\end{figure}

\begin{figure}[H]
    \subfloat[\centering \label{fig:fspace_translation_generator}]{
        \includegraphics[width=0.48\textwidth]{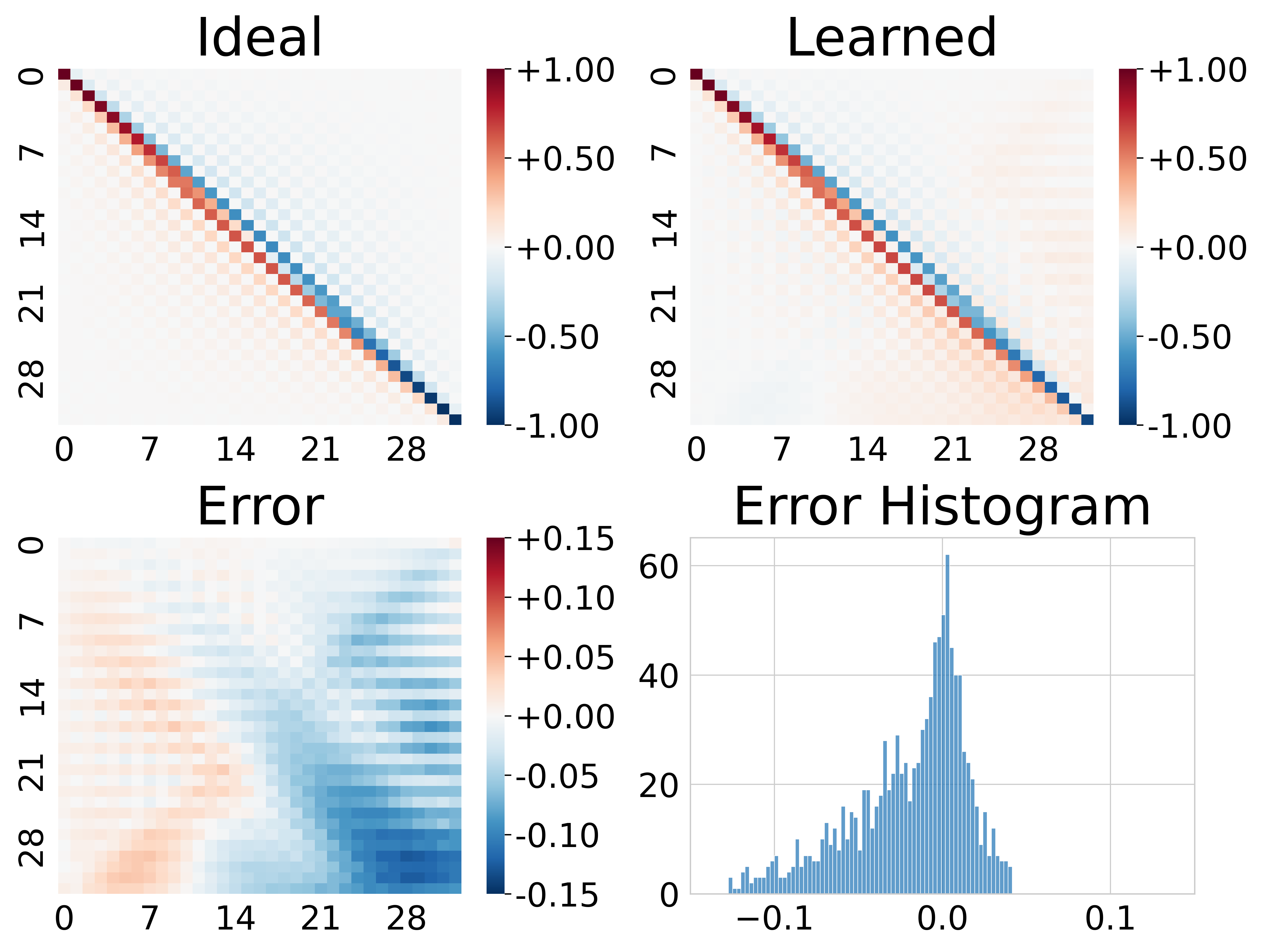}
    }
    \hfill
    \subfloat[\centering \label{fig:fspace_translation_group_convolution_matrix}]{
        \includegraphics[width=0.48\textwidth]{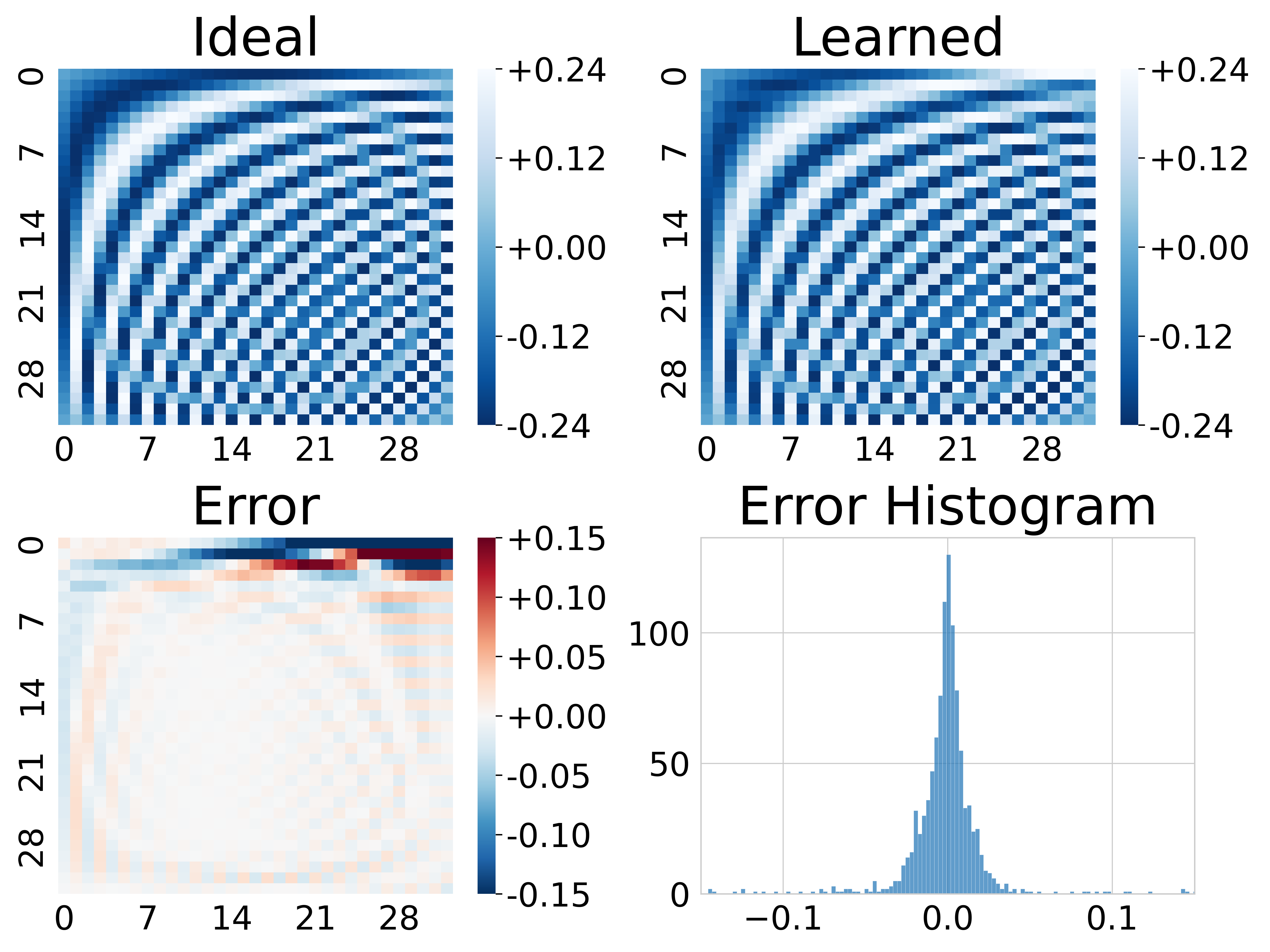}
    }
    \caption{The 
 symmetry generator for the frequency$-$shift$-$invariant dataset is the map
    that sends each discrete sine basis vector to the one with the next highest (frequency). On~the left, 
    we see that this matrix is learned to a high degree of accuracy. The~group convolution
    matrix that provides the symmetry$-$based representation for this case is nothing but the discrete sine transform.
    On the right, we see that this matrix is also learned to a high degree of accuracy. The~model was able to 
    learn that the discrete sine transform was the relevant map for a symmetry$-$based representation by~simply
    looking at samples, without~any hints or hard coding (explicit or hidden). (\textbf{a}) Ideal and learned symmetry generators (\textbf{top}) and error distributions (\textbf{bottom}). (\textbf{b}) Ideal and learned group convolution matrices (\textbf{top}) and error distributions (\textbf{bottom}).}
    \label{fig8}
\end{figure}

\begin{figure}[H]
    \subfloat[\centering \label{fig:permuted_translation_generator}]{
        \includegraphics[width=0.48\textwidth]{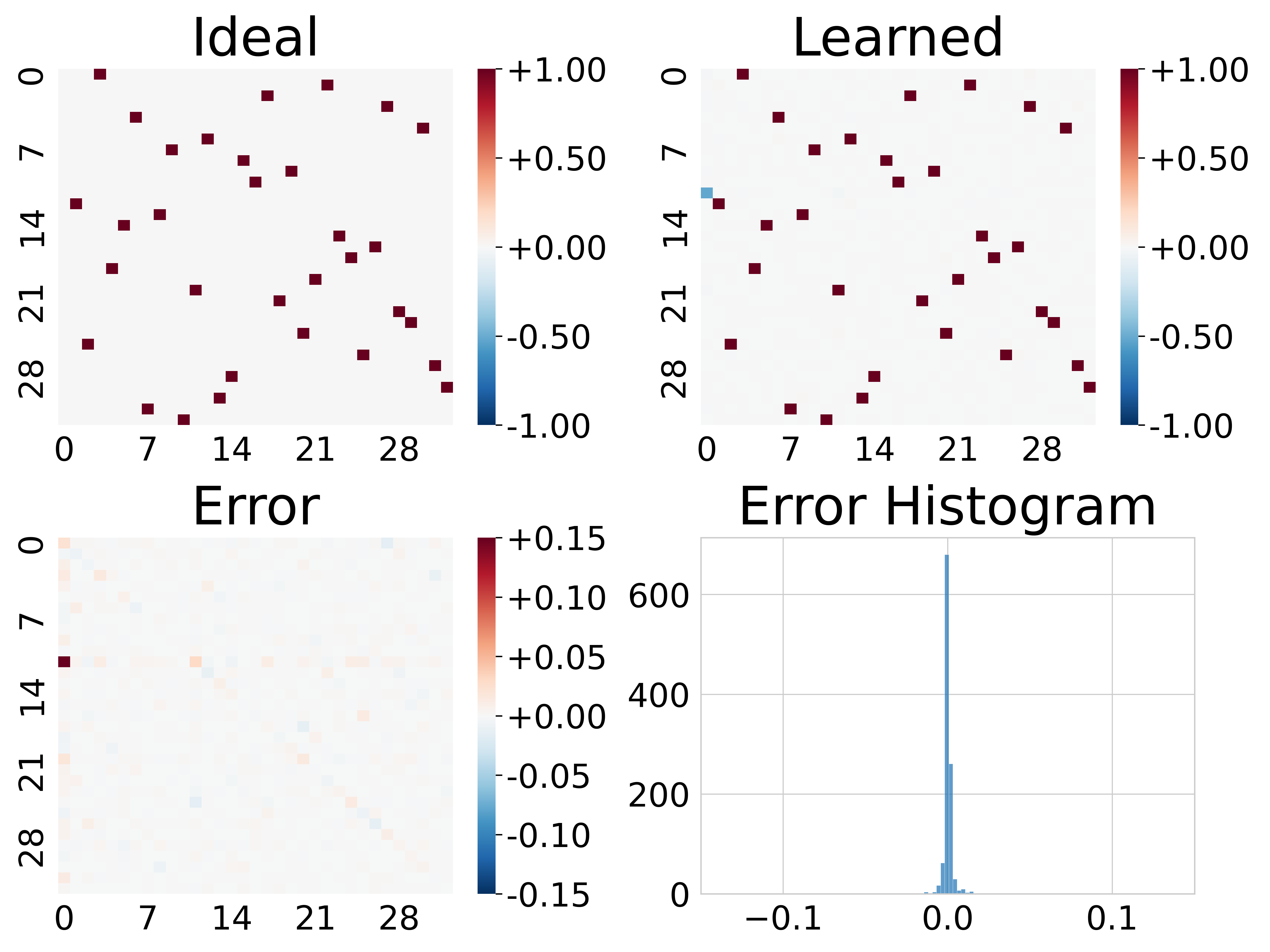}
    }
    \hfill
    \subfloat[\centering \label{fig:permuted_translation_group_convolution_matrix}]{
        \includegraphics[width=0.48\textwidth]{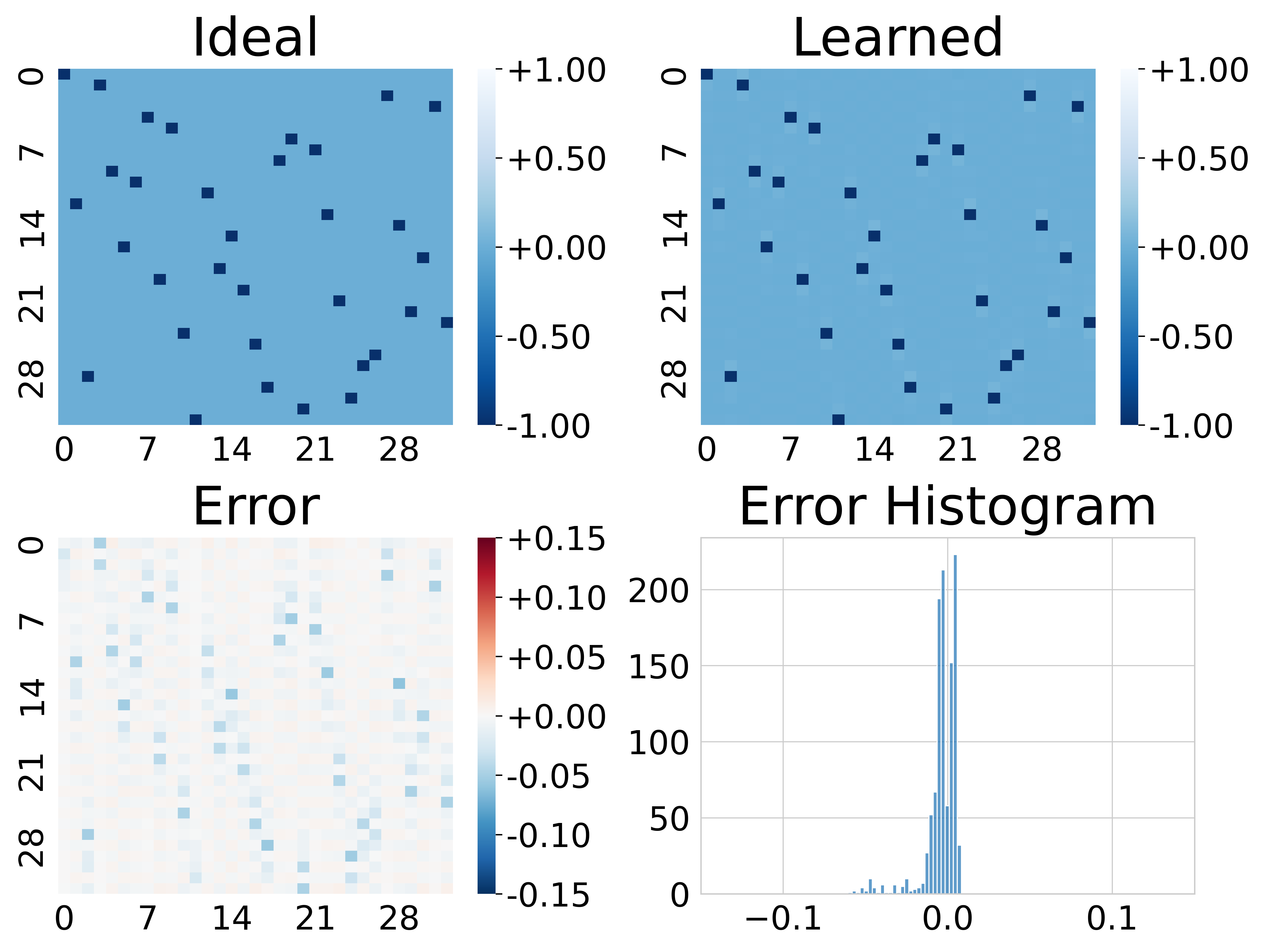}
    }
    \caption{The 
 symmetry generator for the pixel permutation dataset is a specific permutation 
    of the components, and~the dataset is invariant and local under the action of the powers of this permutation.
    On the left, we see that the relevant permutation matrix is learned to a high degree of accuracy. The~group convolution
    matrix that provides the symmetry$-$based representation for this case is the underlying permutation
    that relates the simple translation generator to the permutation generator.
    On the right, we see that this matrix is also learned to a high degree of accuracy. The~model was able to 
    extract the specific permutation of the ``pixels'' that was needed to obtain a manifest locality/symmetry
    from raw data. (\textbf{a}) Ideal and learned symmetry generators (\textbf{top}) and error distributions (\textbf{bottom}). \linebreak  (\textbf{b}) Ideal and learned group convolution matrices (\textbf{top}) and error distributions (\textbf{bottom}).}
    \label{fig9}
\end{figure}

\begin{figure}[H]
    \subfloat[\centering \label{fig:mnist-translation_generator}]{
        \includegraphics[width=0.48\textwidth]{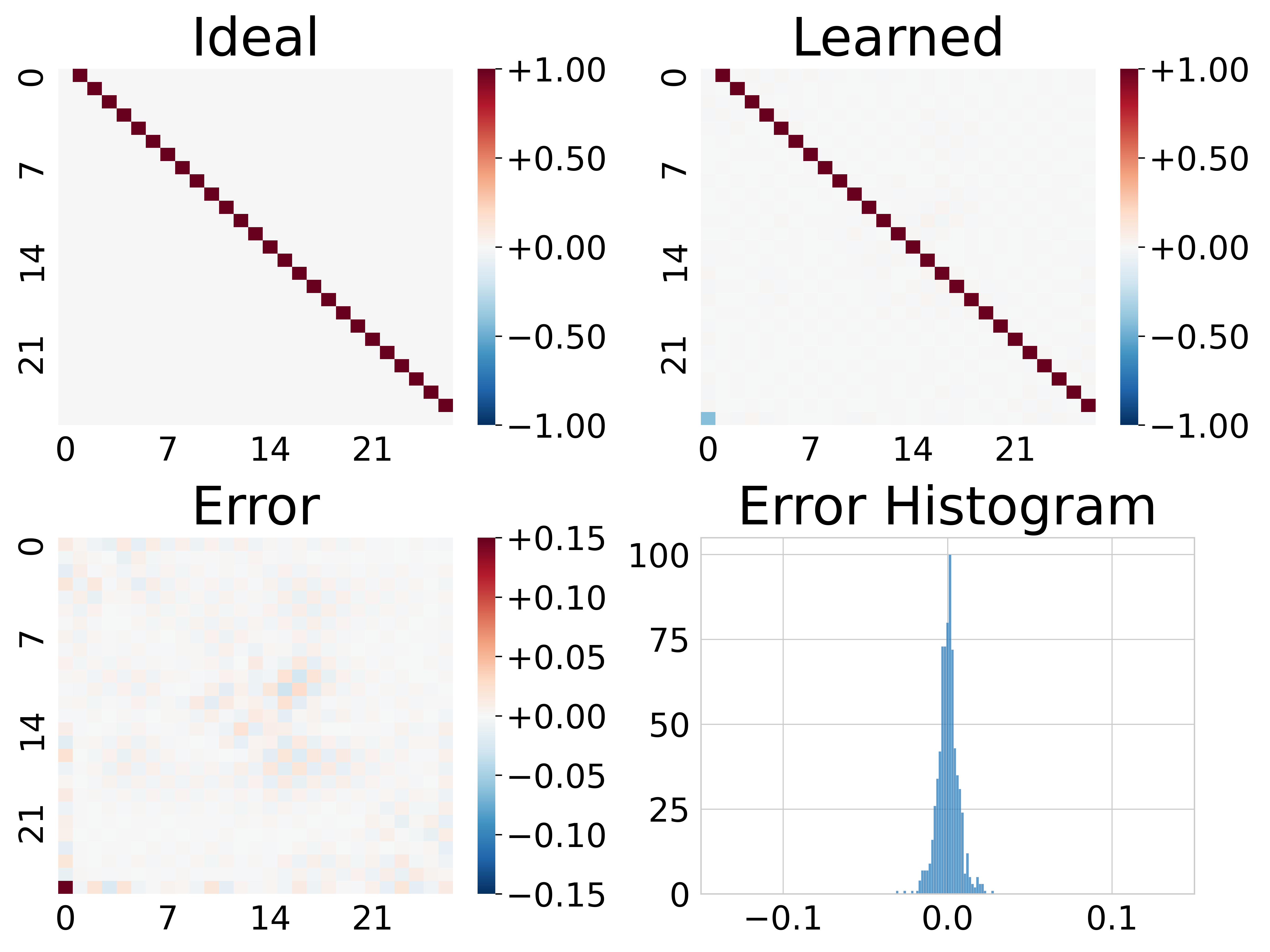}
    }
    \hfill
    \subfloat[\centering \label{fig:mnist-translation_group_convolution_matrix}]{
        \includegraphics[width=0.48\textwidth]{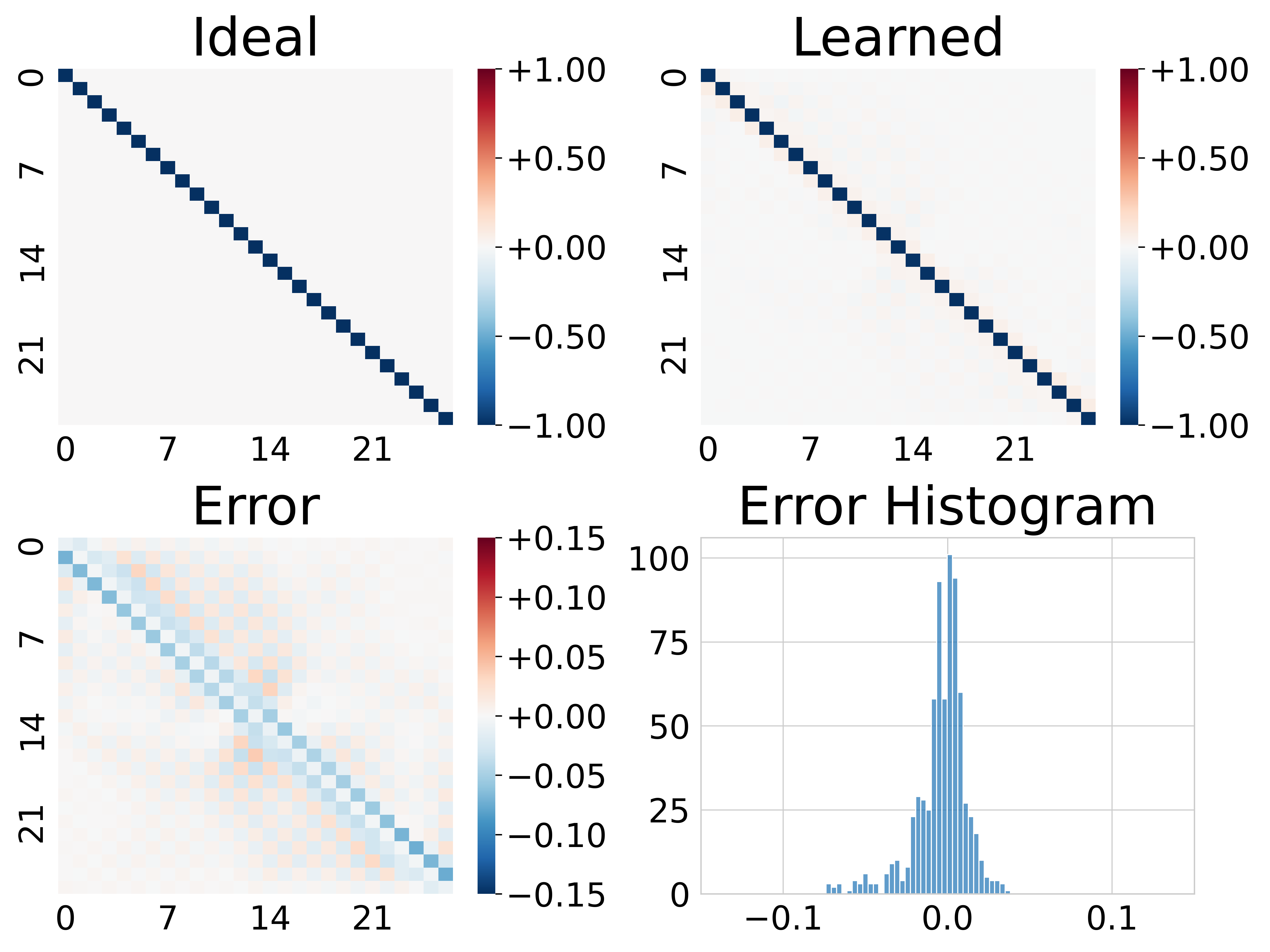}
    }
    \caption{Training 
 results for the 27$-$dimensional real dataset with 
    an approximate translation symmetry, obtained by slicing the MNIST dataset. On~the left, we see that the model learns the 1$-$step translation generator just as well as in the synthetic experiments despite the fact that  the dataset properties are very different from
    that case. On~the right, we see that the group convolution matrix is similarly accurate, approximately equal to the identity map (up to a sign), as~in Figure~\ref{fig7}b. (\textbf{a}) Ideal and learned symmetry generators (\textbf{top}) and error distributions (\textbf{bottom}). (\textbf{b}) Ideal and learned group convolution matrices (\textbf{top}) and error distributions (\textbf{bottom}).}
    \label{fig10}
\end{figure}

\begin{figure}[H]
    \subfloat[\centering \label{fig:model_1d_io_perm}]{%
        \includegraphics[width=0.49\textwidth]{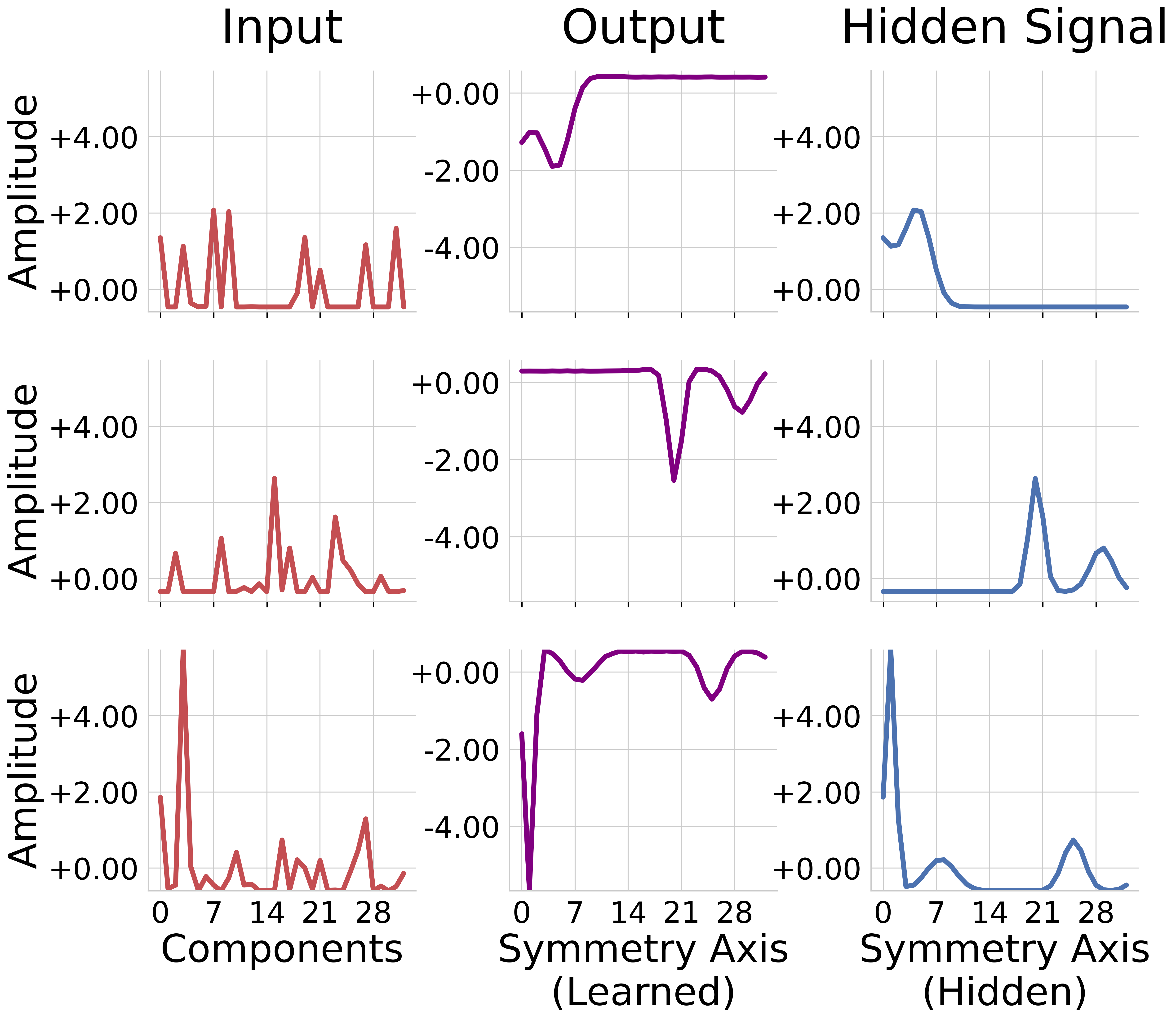}
    }
    \hfill
    \subfloat[\centering \label{fig:model_1d_io_shift}]{%
        \includegraphics[width=0.49\textwidth]{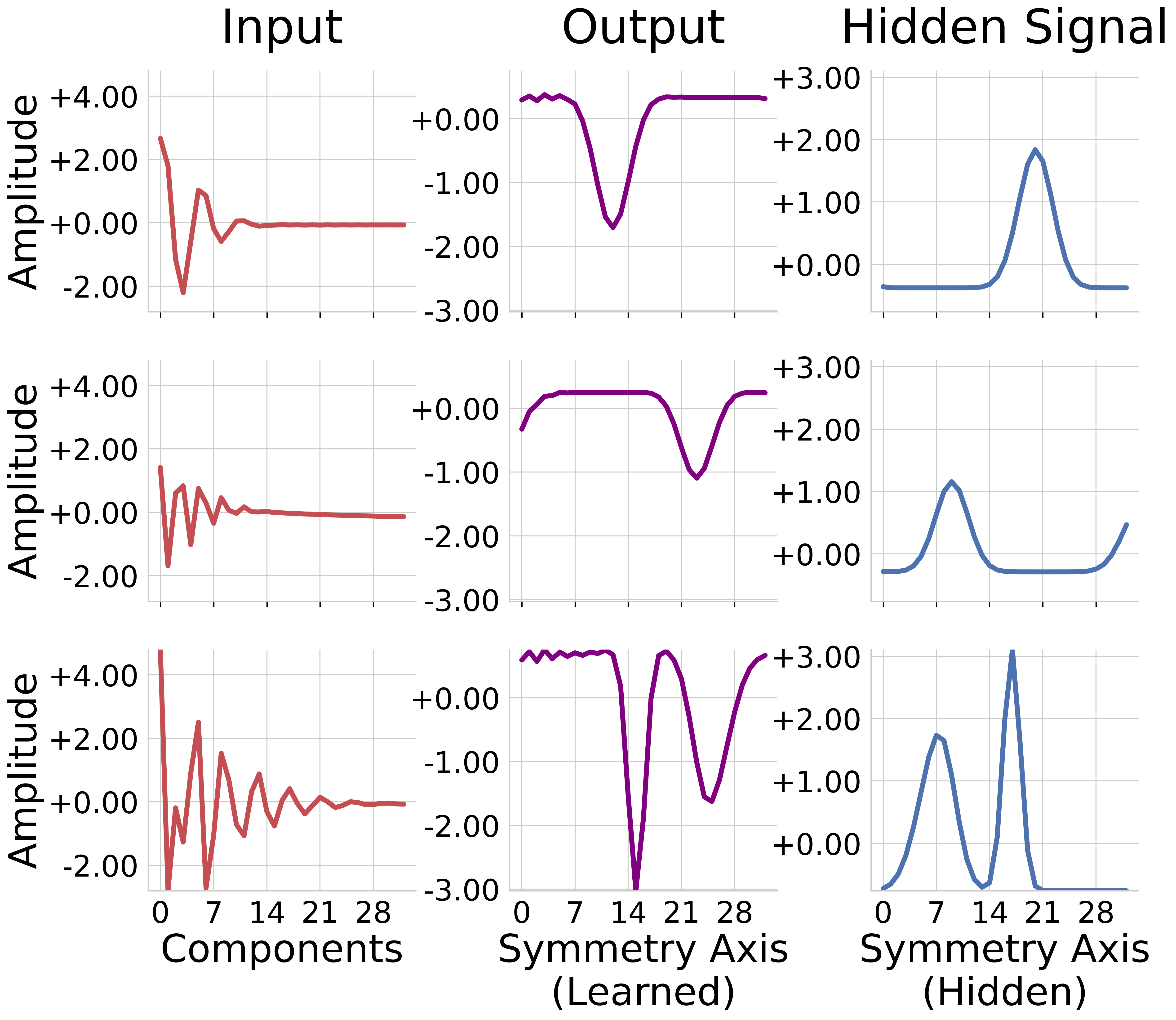}
    }
    \caption{After 
 learning the symmetry generators, the~resulting group
        convolution map can be used to obtain the symmetry$-$based representation
        of each sample. Here, we see that the symmetry$-$based representations
        recover the underlying local signals to a high degree of accuracy for both
        permutation$-$ and frequency$-$shift$-$equivariant data. In~both cases, we see the
        raw data vectors on the left, the~result of the group convolution map in
        the center, and~the hidden (unpermuted) signal on the right. \linebreak  (\textbf{a}) Permutation-equivariant data. (\textbf{b}) Frequency$-$shift$-$equivariant data.}
    \label{fig:model_1d_io_samples}
\end{figure}

\textls[-15]{A visual description of the experimental results is provided
in
Figures
~\ref{fig7}--\ref{fig10}, with~
\mbox{Figures~\ref{fig7}--\ref{fig9}}} containing the results on 33-dimensional synthetic datasets with various symmetries 
and Figure~\ref{fig10} containing the results on a 27-dimensional real dataset (see Appendix \ref{app:real_data}) for details).
In each case, we compare the learned generator to the appropriately 
signed generator of the group. The~main points are as~follows:

\begin{itemize}
    \item In each experiment, the~learned symmetry generator $\gen$ is indeed very close to the underlying correct generator used in preparing the dataset. See Figures~\ref{fig7}a--\ref{fig10}a.
    \item The group convolution operator $L$ formed by combining the learned symmetry generator $\gen$ with the learned resolving filter $\filter$ 
    as in \eqref{eqn:L-group-convolution} is 
    approximately equal to the
     underlying transformation used in generating the dataset (see Figures~\ref{fig7}b--\ref{fig10}b). As~a result,
     the matrix $L$ is highly successful in reconstructing the underlying hidden local signals, 
     resulting in a symmetry-based representation (see Figure~\ref{fig:model_1d_io_samples}).
\end{itemize}

\vspace{-15pt}

\vspace{-15pt}

Thus, by~having access only to raw samples such as those in Figure~\ref{fig:dataset-samples},
the model has been able to learn symmetry generators of quite different
sorts: pixel translations, pixel shuffles, and~
frequency shifts were all learned
with high accuracy using exactly the same model setup and hyperparameters.
The group convolution operator $L$ 
relates the raw data to a representation where the locality and 
 symmetry are manifest. In~the case of frequency shifts, the~
relevant transformation that completes this is the discrete sine transform (DST). 
We find it highly satisfying that, simply by looking at data as in Figure~\ref{fig:dataset-samples}, the~model was able to learn that the DST matrix provides the relevant representation; see Figure~\ref{fig8}b.

\vspace{-12pt}

\vspace{-15pt}

We emphasize that the same hyperparameters have been used in each
case: the learning rates, low-rank entropy estimation scheduling, batch 
size, etc.,  were the same for all the experiments. In~other words, \emph{no fine-tuning was necessary} for different data types. 
The results are also stable in the sense that training the system from 
scratch results in similar outputs each time. In~our various experiments with 
the chosen settings, only once did we encounter a convergence problem where the 
training became stuck in an unsatisfactory local~minimum.

\vspace{-15pt}

We repeated each one of the seven experiments four times and reported the 
cosine similarity in Table~\ref{tab:cosine-similarity-generator} (cosine similarity here is 
obtained by treating the generator matrix as a vector, or, equivalently, using the
Hilbert--Schmidt (or Frobenius) inner product between the two matrices
of the symmetry generator). The~error histograms over the entries of the symmetry
generator are provided in Figures~\ref{fig7}a--\ref{fig9}a.

\begin{table}[H]
    \caption{Cosine 
 similarity between the learned and ideal minimal~generators.}
    \label{tab:cosine-similarity-generator}
    \begin{tabular*}{\hsize}{c@{\extracolsep{\fill}}ccc}
        \toprule
        \textbf{Symmetry}   & \textbf{Gaussian Dataset} & \textbf{Legendre Dataset} & \textbf{MNIST Slices} \\
        \midrule
        Translation         & \(0.999 \pm 0.001\)       & \(0.998 \pm 0.003\) & \(0.999 \pm 0.001 \)      \\ 
        \midrule
        \makecell{Permuted\\translation} & \(0.996 \pm 0.002\)       & \(0.998 \pm 0.001\) &      -      \\
        \midrule
        \makecell{Frequency space\\translation}    & \(0.980 \pm 0.003\)       & \(0.991 \pm 0.003\)   &  -  \\
        \bottomrule
        
    \end{tabular*}
\end{table}

Some notable properties of the method worth emphasizing are as~follows:

\textbf{Learning the minimal group generator.} A dataset with a pixel translation symmetry also has 
a symmetry under two-pixel translations, three-pixel translations, etc.
In fact, each one of these actions could be considered as a generator
of a subgroup of the underlying group of symmetries.
Some works in the literature have successfully learned
symmetry generators from datasets but ended up learning one of many subgroup generators instead of the minimal generator. Our inclusion of locality together with stationarity in the form of alignment and resolution losses has allowed the method to learn the minimal generator in each~case.

\textbf{Learning a symmetry-based representation.} In addition to the symmetry generator, our model learns
a resolving filter that is local along the symmetry direction.
In all our experiments, this resulted in the appropriate
``delta function along the symmetry direction'' 
described in Section~\ref{section:data_model}.
In the case of simple translation symmetry, the~filter has a single
nonzero component; in~the case of the frequency shift, it is 
a pure sinusoidal. The~
convolution operator $L$ obtained by combining the generator and
the filter provides a symmetry-based representation
of the data, turning the hidden symmetry into a simple translation 
symmetry. Such a representation is exactly the 
type of situation where regular CNNs are effective. 
Thus, for~a given supervised 
learning problem, one can first train our model to learn the underlying 
symmetry and~then transform the data into the symmetry-based representation and  feed the resulting form into a regular multi-
layer CNN architecture, making the model an adapter between raw data
and CNNs. In~Appendix \ref{app:equivariance_proof}, we prove the relevant
equivariance property of the learned~representation.

\begin{figure}[H]
    \subfloat[\centering \label{fig:permuted_translation_lifting_map_training}]{
        \includegraphics[width=0.48\textwidth]{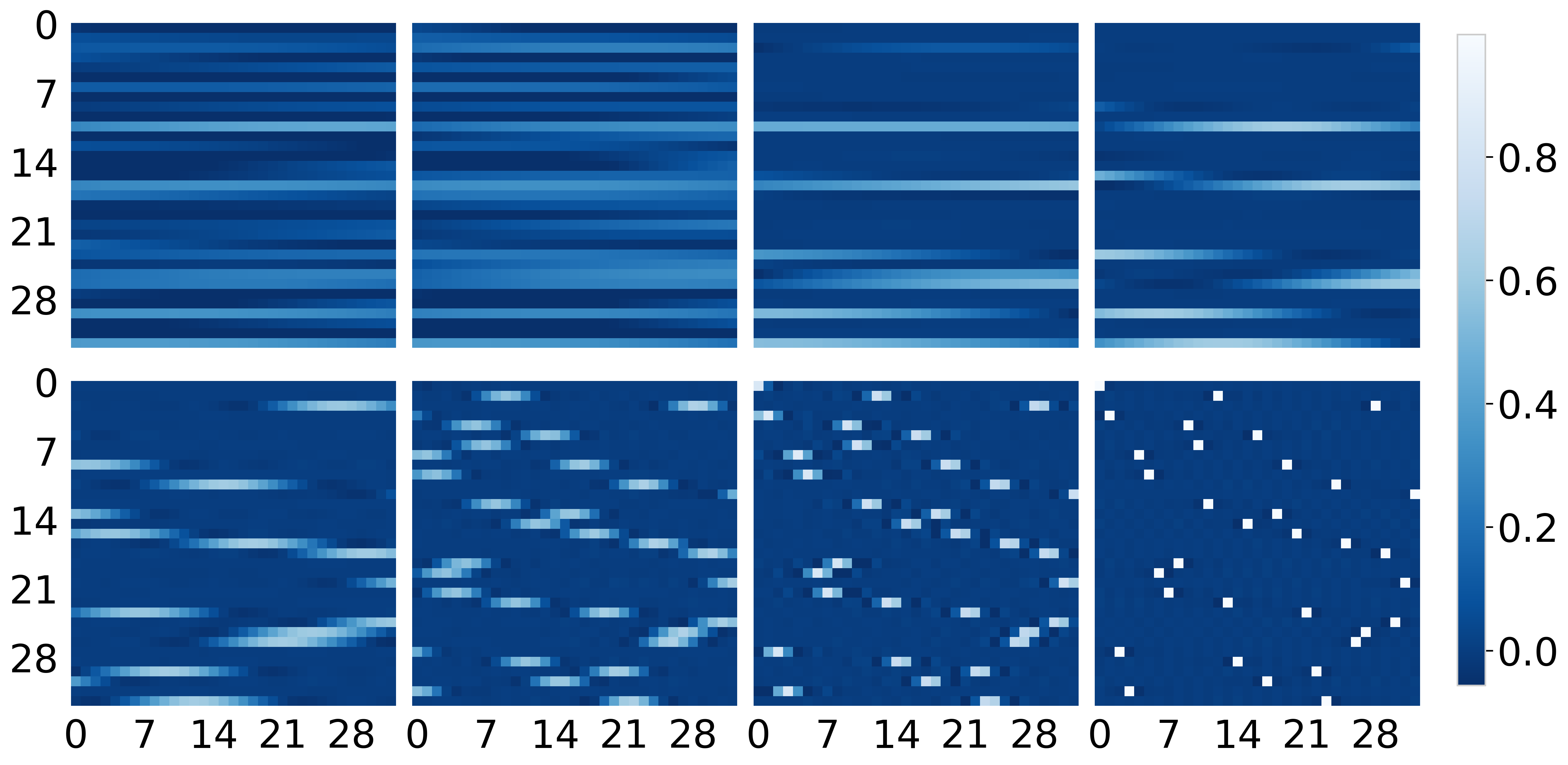}
    }
    \hfill
    \subfloat[\centering \label{fig:dst_translation_lifting_map_training}]{
        \includegraphics[width=0.48\textwidth]{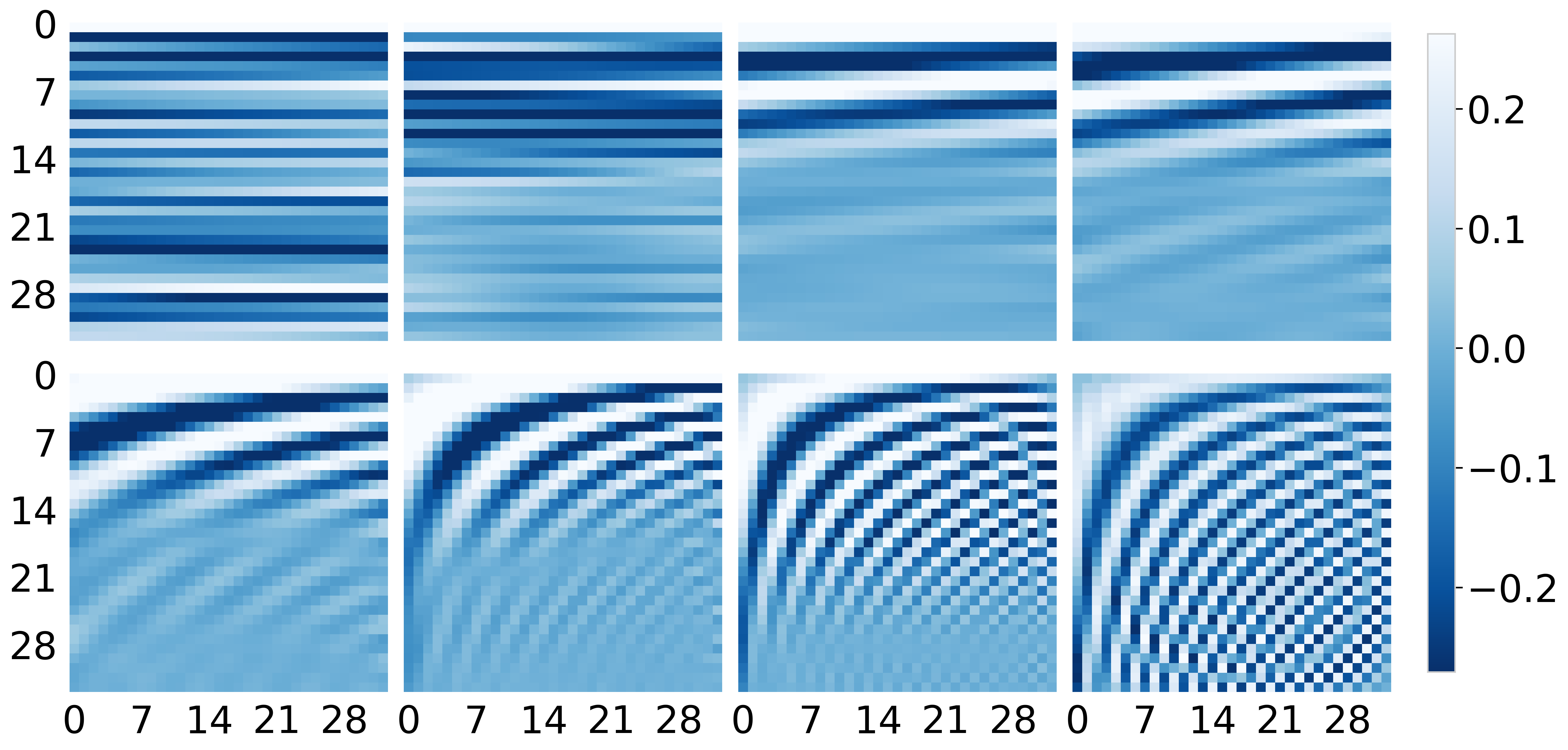}
    }
    \caption{Training snapshots of the group convolution layer. The~
        group convolution matrix evolves to a mapping that makes the
        hidden symmetry manifest. (\textbf{a}) Training snapshots for a dataset with the permuted translational symmetry. The~group convolution matrix evolves to a form that unscrambles the permutation. (\textbf{b}) Training snapshots for a dataset with the frequency$-$shift symmetry. The~group convolution matrix evolves into a form that negates the DST$-$I transformation, i.e.,~converges to the transpose of the DST$-$I matrix.}
    \label{fig:group-convolution-snapshots-training}
\end{figure}

\textbf{Stability.} Our datasets are $33$-dimensional,
and thus the symmetry generator $\gen$ and the symmetry-based representation matrix $L$ have shapes of $33\times 33$. These dimensions are higher than in many of the works 
that have a comparable aim to ours (i.e., unsupervised learning 
of symmetries from raw data \citep{desai2022symmetry, yang2023generative, yang2023latent}). We emphasize that we do not have any built-in sparsity or factorization enforced on 
the matrices. The model has to indeed perform the searches over these large spaces. Our loss function involving locality, stationarity, 
and information preservation has resulted in a highly robust system that reliably avoids local minima and
finds the correct symmetries in each of the examples we implemented.
We emphasize that none of our datasets are explicitly or~cleanly
symmetric under the relevant group actions (see Figure~\ref{fig:dataset-samples} for samples). The~almost perfect recovery of the symmetry
generator from such samples is rather striking.
Examples of the robust optimization dynamics can be seen in Figure~\ref{fig:group-convolution-snapshots-training}.

\vspace{-6pt}

\subsection{Comparison with Other Unsupervised Symmetry Learning~Approaches}


The GAN-based methods SymmetryGAN \citep{desai2022symmetry} and 
LieGAN \citep{yang2023generative}, like our method, are set up 
to learn symmetry transformations starting with raw data in an 
unsupervised manner. To~compare the performance of these methods
with ours, we ran experiments with a seven-dimensional version of 
our translation-invariant Gaussian dataset described above.
In order to comply with the setting of  
\citep{desai2022symmetry} and \citep{yang2023generative}, this time, we set up the dataset so that the translation is circulant 
(periodic). Our method's setup can also be extended to cover circulant 
symmetries, but~we attempted the run without this modification and~in 
this sense provided an advantage to the GAN-based~methods.

\begin{figure}[H]
    \subfloat[\centering \label{fig:liegan-learned-generator}]{
        \includegraphics[width=0.48\textwidth]{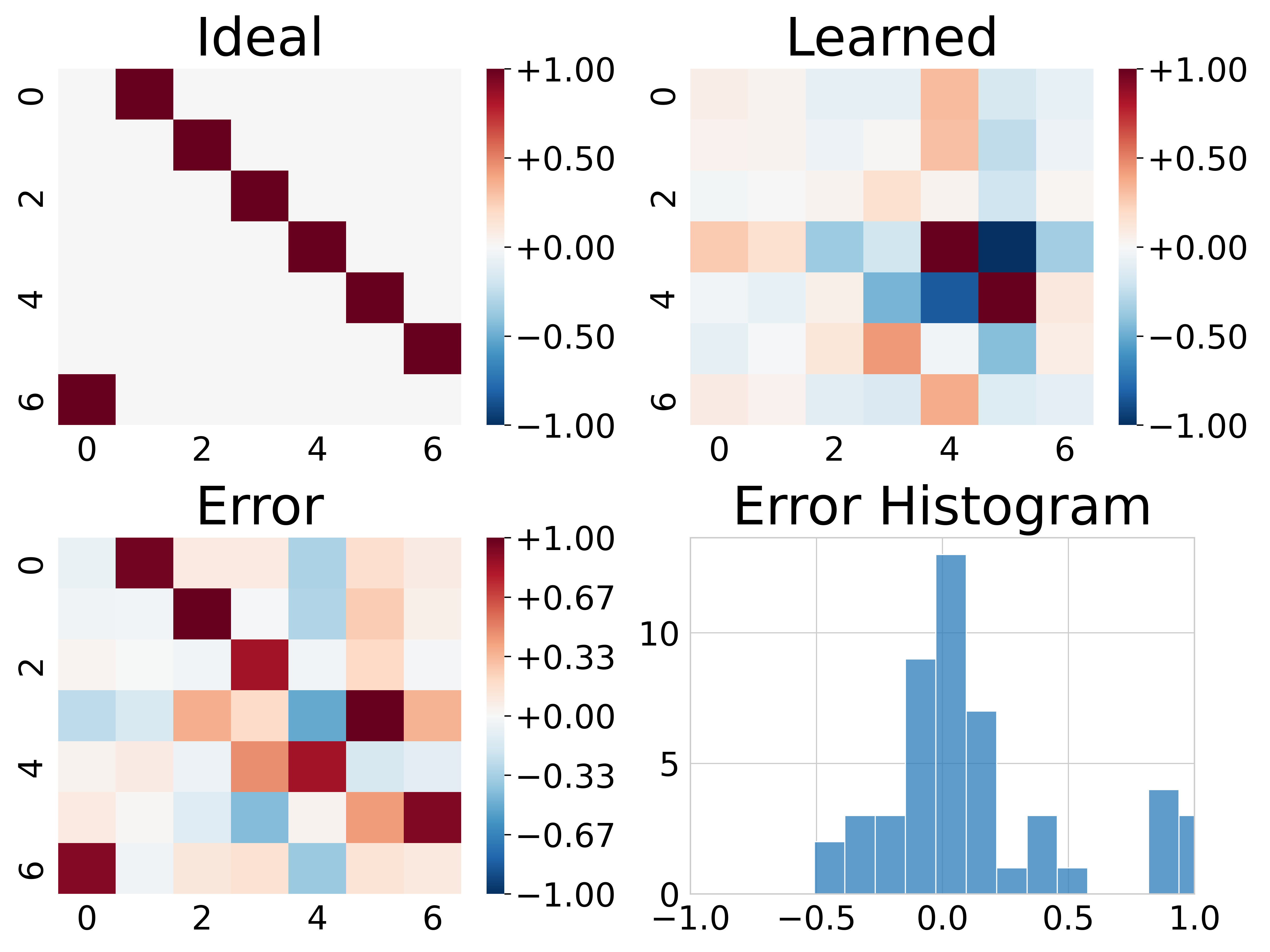}
    }
    \hfill
    \subfloat[\centering \label{fig-symmetrygan-learned-generator}]{
        \includegraphics[width=0.48\textwidth]{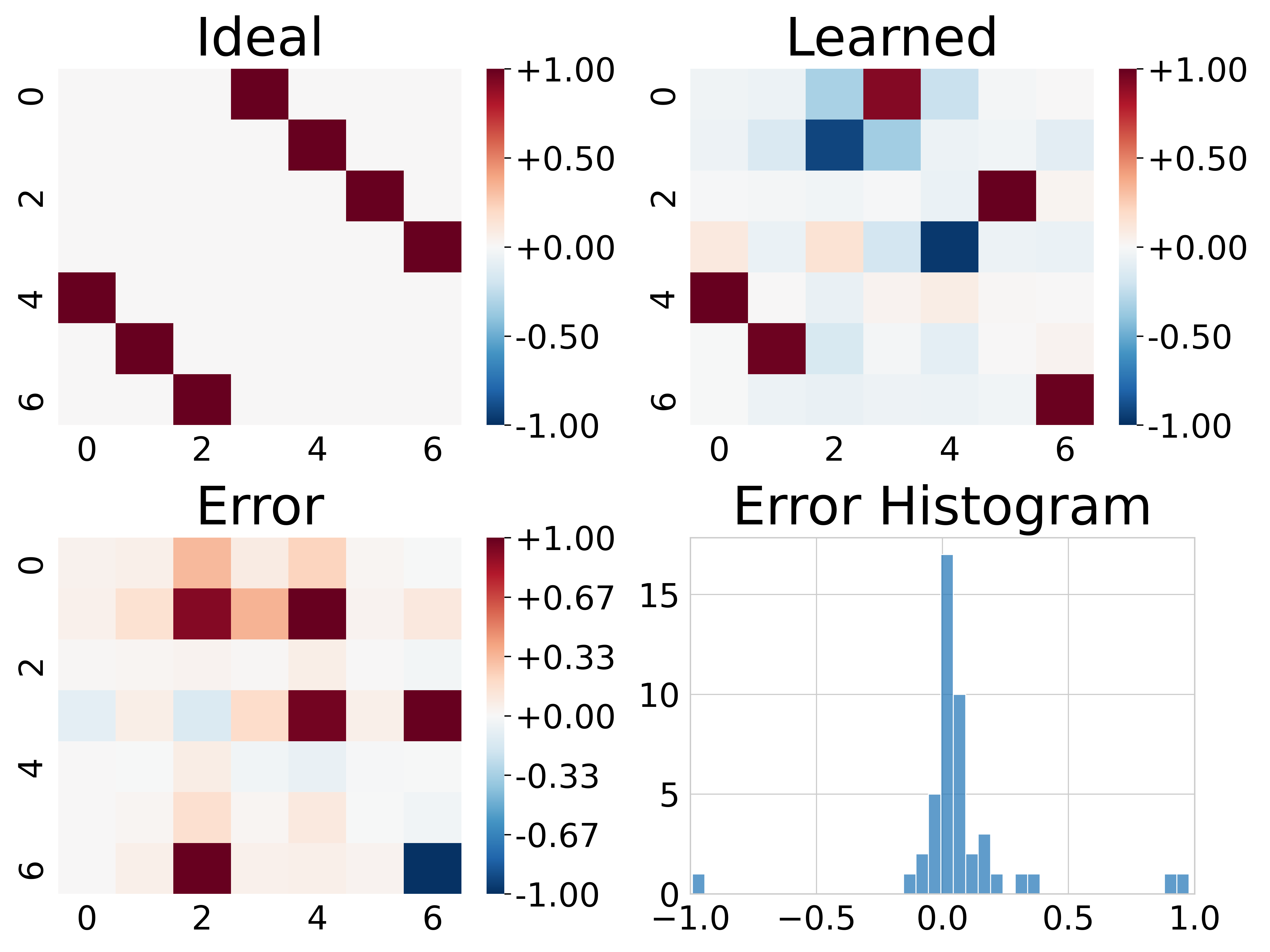}
    }
    \caption{The outcome of the GAN-based methods on the 7$-$dimensional dataset with circulant translation symmetry. We see that GAN-based methods learn the symmetry generators
    rather crudely, with~the error rates being slightly higher for LieGAN compared to SymmetryGAN$-$QR. (\textbf{a}) Ideal and learned symmetry generators (\textbf{top}) and error distributions (\textbf{bottom}) for the LieGAN method. (\textbf{b}) Ideal and learned symmetry generators (\textbf{top}) and error distributions (\textbf{bottom}) for the SymmetryGAN$-$QR~method.}
    \label{fig:circulant-7ts-gan-results}
\end{figure}
\begin{figure}[H]
    \subfloat[\centering \label{fig:learned-generator-7ts}]{
        \includegraphics[width=0.48\textwidth]{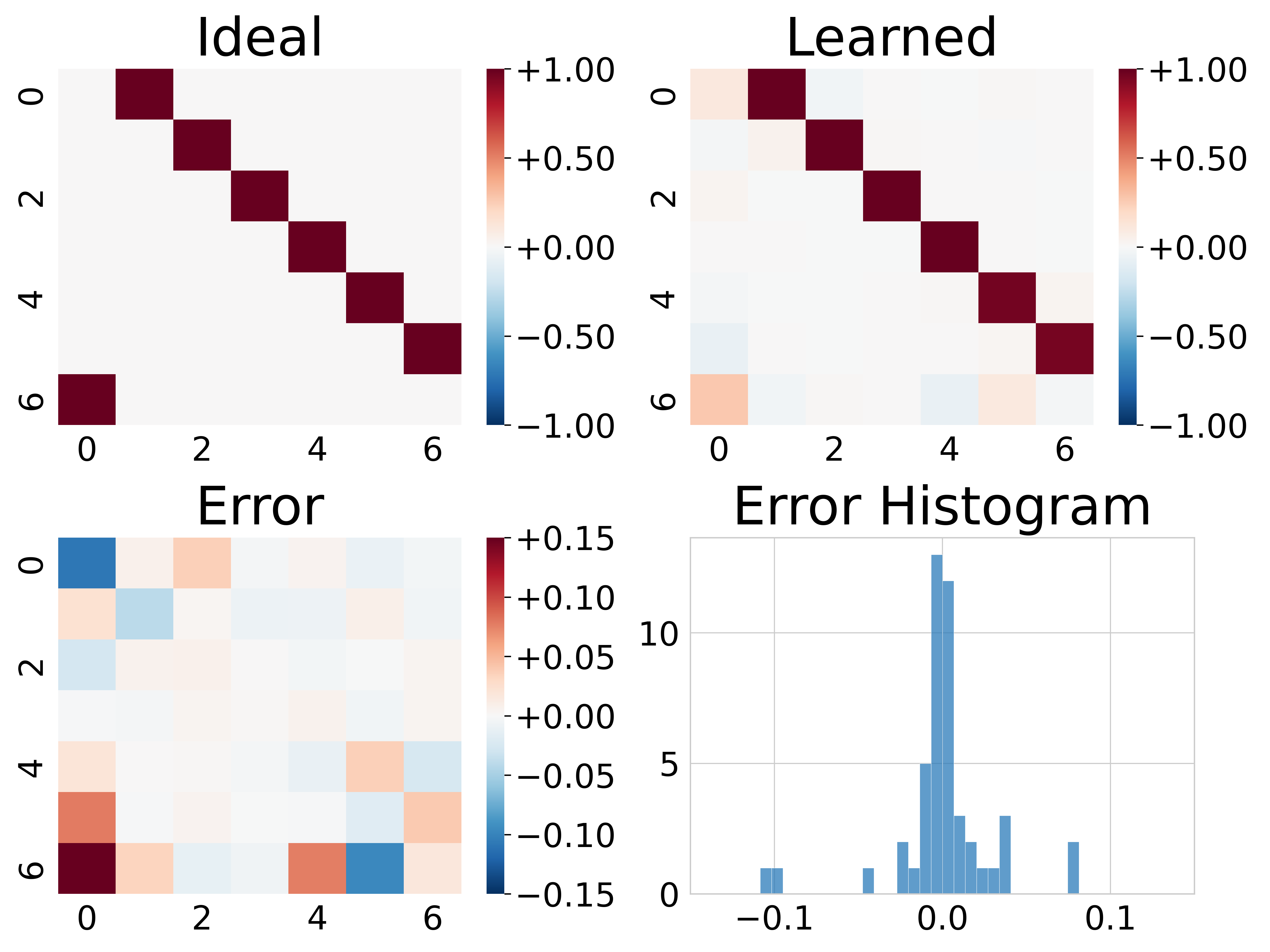}
    }
    \hfill
    \subfloat[\centering \label{fig:learned-group-convolution-matrix-7ts}]{
        \includegraphics[width=0.48\textwidth]{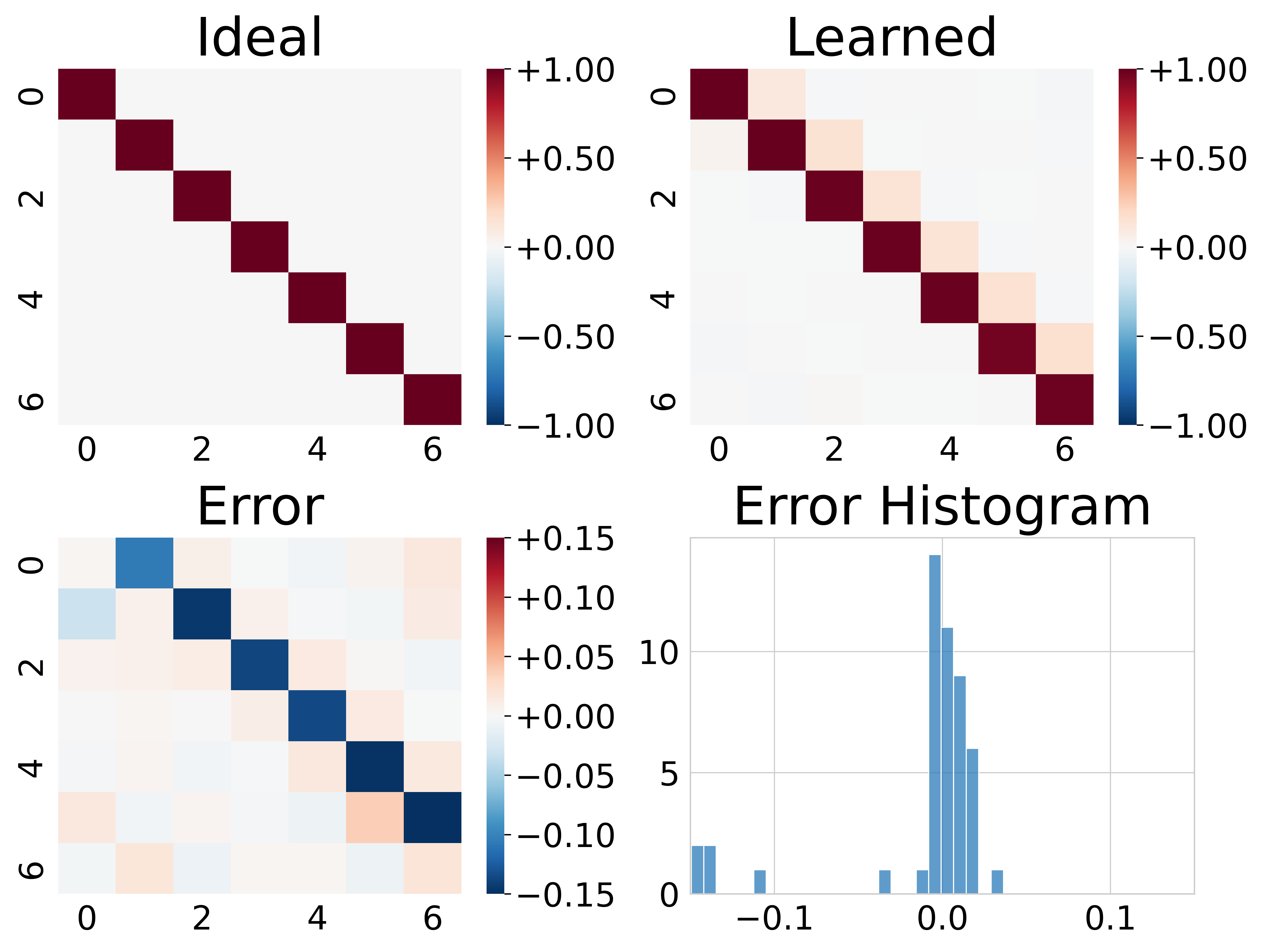}
    }
    \caption{The outcome of our method on the 7$-$dimensional dataset with circulant translation
    symmetry. We see that, compared to the GAN-based methods, our method learns the minimal symmetry generator and resulting group convolution matrix with higher accuracy. (\textbf{a}) Ideal and learned symmetry generators (\textbf{top}) and error distributions (\textbf{bottom}) for our method. (\textbf{b}) Ideal and learned group convolution matrices (\textbf{top}) and error distributions (\textbf{bottom}) for our method.}
    \label{fig:circulant-7ts-symmetrylens-results}
\end{figure}

In the end, the~best cosine similarity obtained by SymmetryGAN was \(0.527\)
and that obtained by LieGAN was \(0.425\), whereas
the cosine similarity for our model was \(0.958\). 
See Figures~\ref{fig:circulant-7ts-gan-results} and \ref{fig:circulant-7ts-symmetrylens-results}
for a visualization of the learned generators by each method, and~see Appendix \ref{app:gan-comparison} for the details of the~comparison.

\vspace{-15pt}

\section{Discussion}\label{section:discussion}

Our experiments show that the method described in this paper
can uncover the symmetries (generators of group representations) of quite
different sorts, starting with raw datasets that are only approximately 
local/symmetric under a given action of a \mbox{one-dimensional} Abelian Lie group. 
In addition to symmetry generators, a~symmetry-based 
representation \citep{higgins2022symmetry} is also learned.
The choice of the loss function and the approach used in optimization 
result in highly stable results. This is the outcome of extensive experimentation with
different approaches to the same intuitive ideas described in the 
Introduction. It is satisfying, and~perhaps not surprising, that the current
successful methodology is the simplest among the range of approaches we~tried.

In this paper, we focused on the action of one-dimensional Abelian Lie 
groups. More generally, the~symmetry group will be multidimensional.
The case of translation symmetry in the plane has
two generators that commute with each other, and~a generalization 
of the learning problem to such a case will involve adapting the
loss function to encode locality and symmetry under the 
joint actions of those two generators, taking into account
possible non-commutation of the latter. More generally, one 
should consider non-Abelian Lie groups whose generators do 
not commute with each~other.

The datasets used in our experiments were 33-dimensional
and thus are not
what one would call ``low-dimensional'' in the context of symmetry learning.
However, many real datasets have much higher dimensionalities. Our experiments indicate that 
our method can work on datasets with three times the dimensionality used in this paper without difficulty; however, for~dimensions that are orders of magnitude higher, one will likely 
require further computational, and~possibly methodological, improvements. Such 
improvements are necessary for, say, typical image~datasets.

In this paper, the~group actions learned by the model are all linear; i.e.,
the method learns a \emph{representation} of the underlying group of symmetries
by learning a matrix for each generator. However, more generally,
symmetry actions can be nonlinear. By~replacing the linear layer that
represents the symmetry action with a more general nonlinear architecture
it may be possible to apply the philosophy of this paper
to the nonlinear case as well, but, of course, the~practical problems with 
optimization and the architectural choices for representing a nonlinear map
will require work and~experimentation.

In transforming the data using the group convolution map $L$, 
we made the choice of having the number of 
output dimensions equal to the number of input dimensions. This meant 
setting the number of powers used for the symmetry generator
$\gen$ equal to the number of input dimensions. While this is a reasonable choice
and is suited to the datasets here, other choices could be appropriate for
other settings. We believe it would be worthwhile to experiment with other choices
here, possibly by using datasets with cyclic group symmetries whose
order is different from the number of dimensions of the datasets they act~on.

Finally, let us note that data often have locality directions that are not 
necessarily symmetry directions. For~instance, consider a zeroth-order
approximation to the distribution of (say daily) temperatures around  Earth.
This distribution will have locality under rotations around any axis passing 
through the center of  Earth
but will be approximately symmetric only under azimuthal rotations. In~other words,
temperature varies slowly as you move in any direction, but~only similar 
latitudes will have similar temperature distributions; the poles will not be as hot 
as the equator. Our model seeks group generators that have both locality and symmetry
properties for the dataset at hand, so it is not appropriate for such a dataset.
A generalization that would also work for cases like this would be an interesting
and worthwhile~project.

\vspace{6pt}

\authorcontributions{Conceptualization, O.E. and A.O.; methodology, O.E. and A.O.; software, O.E.; validation, O.E. and A.O.; formal analysis, O.E. and
A.O.; investigation, O.E.; resources, A.O.; data curation, O.E.; writing---original draft preparation, O.E. and A.O.; writing---review and editing, A.O.; visualization, O.E.; supervision, A.O.; project administration, A.O.; funding acquisition, A.O. All authors have read and agreed to the published version of the~manuscript.}

\funding{This research was funded by Turkish Scientific and Technical Research 
Council (TUBITAK) under the BIDEB-2232 program with grant number 118C203.}

\dataavailability{All data in this study are generated synthetically, and~the complete code for data generation, models, and~optimization code, as~well as the reproducible experiments, can be found in our GitHub repository (SymmetryLens v1.0) at 
 \url{https://github.com/onurefe/SymmetryLens.git} (accessed on 12 March 2025).}

\acknowledgments{We would like to thank the current and former members of the EarthML research group for fostering a collegial and dynamic research environment, Ayse 
 Ruya Efe for her assistance in figure preparation, and~Nurgul Ergin for administrative~support.}

\conflictsofinterest{The authors declare no conflicts of interest. The~funders had no role in the design of the study; in the collection, analyses, or~interpretation of data; in the writing of the manuscript; or in the decision to publish the~results.} 

\appendixtitles{yes} 
\appendixstart
\appendix
\section[\appendixname~\thesection. Estimation~Procedures]{Estimation~Procedures}\label{app:estimation-procedures}
\subsection[\appendixname~\thesubsection. Probability~Estimation]{Probability~Estimation}\label{app:probability_estimation}
For estimating probability densities, we use a modified version of a mixture of the Gaussian approach described in \citep{pichler2022differential} 
by minimizing the entropy of estimated probability densities. 
Since we need a large number $O(d^2)$ of joint distributions, a~
straightforward application of the approach in \citep{pichler2022differential} 
is computationally prohibitive, and~we developed an $O(d)$ approach as 
described in Appendix \ref{app:probability_estimation}.

In the stationarity/uniformity loss of Section~\ref{sec:stationarity-uniformity}, 
we need estimates of joint 
probability densities $p_{ij}$ for pairs $y_i$, $y_j$ of components of 
the transformed samples $\mathbf{y}$. We compute these using a 
decomposition in terms of the marginal densities $p_i$ and the conditional 
densities $p_{i|j}$:  $p_{ij}(y_i, y_j) = p_{i|j}(y_i|y_j)p_j(y_j)$. We describe 
below the estimation of these two~pieces.

\subsubsection[\appendixname~\thesubsubsection. Marginal Probability~Estimation]{Marginal Probability~Estimation}
We use a Gaussian mixture model to estimate the marginal densities. 
Denoting the probability density function of a normal distribution
with mean $\mu$ and standard deviation $\sigma$ by 
$\mathcal{K}(u; \mu, \sigma)$, we write the marginal distribution
for variable $y_i$ as a mixture of $M$ Gaussians~as
\begin{equation}\label{eq:marginal-mixture}
    \hat{p}_i(u) = \frac{1}{M} \sum_{m=1}^M w_{mi} \mathcal{K}(u; \mu_{mi}, \sigma_{mi})
\end{equation}
where $w_{mi}$ is the weight of Gaussian kernel number $m$ for variable $y_i$, 
normalized as $\sum_m w_{im} = 1$. We choose $M=4$ 
and, as in \citep{pichler2022differential}, we find the parameters $w_{mi}, \mu_{mi}, \sigma_{mi}$, using gradient descent on a loss function consisting of 
the estimated entropy, \linebreak  \(\hat{h_i} = -\mathbb{E}[\log(\hat{p}_i(y_{i}))]\),
where the expectation is computed as an average over each~batch.

\subsubsection[\appendixname~\thesubsubsection. Conditional Probability~Estimation]{Conditional Probability~Estimation}
\textbf{Overview}
To estimate conditional probabilities $p_{j|i}(u|y_j)$ 
\citep{pichler2022differential}, generalize the mixture model approach in 
\eqref{eq:marginal-mixture} to a parametrized form
\begin{align}
    \hat{p}_{j|i}(u|y_{i}) = \frac{1}{M} \sum_{m=1}^M w_{mji}(y_i) \mathcal{K}(u; \mu_{mji}(y_i), \sigma_{mji}(y_i))
\end{align}
where the parameters $w_{mji}(y_i)$, $\mu_{mji}(y_i)$ $\sigma_{mji}(y_i)$
are now functions of the conditioning parameter $y_i$ to~be learned. 
One could, in~principle, train neural networks for each one of these parameters,
including one entropy loss $-\mathbb{E}[\log(\hat{p}(y_{j}|y_{i}))]$
per pair; however, in~our case, we need $O(d^2)$ such conditional estimators,
and such a straightforward approach becomes computationally prohibitive. For~this 
reason, we propose a modification that involves a ``single-input multiple-output'' approach, as we describe~next.

Instead of including a separate set of networks for each choice of 
input coordinate $y_i$ for  parameters  $w_{mji}(y_i)$, $\mu_{mji}
(y_i)$ $\sigma_{mji}(y_i)$, we train a single network that takes in 
a vector $\mathbf{y}$ but~simply masks all components of this 
vector except the relevant one. In~other words, the~estimated 
conditional probability for a given conditioning variable $i$ is 
provided as
\begin{equation}\label{eq:conditional_probability_estimator}
    \hat{p}_{j|i}(u|\mathbf{y}) = \frac{1}{M} \sum_{m=1}^M w_{mj}(\mathbf{e}_i^T \cdot \mathbf{y}) \mathcal{K}(\mu_{mj}(u; \mathbf{e}_i^T \cdot \mathbf{y}), 
    \sigma_{j}(\mathbf{e}_i^T\cdot\mathbf{y}))
\end{equation}
where $\mathbf{e}_i$ denotes the standard column basis vector,
with a $1$ in the $i$th entry and~zeros everywhere else,
$\mathbf{e}_i^T = (0, 0, \ldots, 1, 0, \ldots, 0)$. Notice that 
the parameters $w_{mj}$, $\mu_{mj}$ $\sigma_{mj}$ no longer
have an $i$ index that labels the conditioning coordinate; 
the network is forced to ``notice'' the relevant coordinate
via the masking procedure and~training. Once again, 
the loss function consists of pieces 
$\hat{h}_{i|j} = -\mathbb{E}[\log(\hat{p}_{j|i}(y_{j}, y_{i}))]$ 
with the expectation 
approximated by an average over each batch
for each pair $i\ne j$. During~training, we let each
sample in a batch  contribute to the estimates for a single 
conditioning variable $i$, with~$i$ looping over all 
coordinates as the sample index $n$ increases, i.e.,~
 $i \equiv n \pmod{d}$. 

\textbf{Architecture}  We train 3 neural networks in total,
one for each of $w$, $\mu$, and $\sigma$. For~each sample, each
network outputs $M$ vectors (one vector for each Gaussian)
with one component for each coordinate $j$ in $f_{j|i}$ 
in the form of a flattened array with 
$d\times M$ components. The~three networks are identical except for the output activation~functions. 

Each network is composed of \(d\) input and \(d \times M\) output neurons, the~input representing the 
value of the conditioning
component via the masking procedure described and~the output corresponding to the flattened version of 
the tensor representing the parameters of the Gaussians. 
We use 
one
hidden fully connected layer with \(4 \times d \times M\) neurons~each.

We take the number of Gaussian kernels to be $M=4$. We use the LeakyReLU activation function
with $\alpha=0.1$ at the output of each layer except for the last 
one. The~final output activation functions for the three neural 
networks are given in Table~\ref{tab:activation_functions}. 
\begin{table}[H]
 \caption{Activation functions for the conditional probability~estimator.}
    \label{tab:activation_functions}
    \begin{tabular*}{\hsize}{c@{\extracolsep{\fill}}c}
    \toprule
    \textbf{Estimated Quantity} & \textbf{Output Activation Function} \\
    \midrule
    Kernel weights              & Softmax over the kernel axis \\
    \midrule
    Mean value of each kernel   & No (linear) activation \\
    \midrule
    Variances of each kernel    & Scaled tangent hyperbolic \\                                                & function followed by exponentiation \\
    \bottomrule
    \end{tabular*}
\end{table}
\unskip

\subsection[\appendixname~\thesubsection. Multidimensional Entropy~Estimation]{Multidimensional Entropy~Estimation}\label{app:low-rank-entropy-estimation}
To estimate the multidimensional differential entropy of the
transformed data $\mathbf{y}$, we use a multivariate 
Gaussian approximation. For~a multivariate Gaussian with covariance matrix 
$\mathbf{C}$ whose eigenvalues are $\lambda_i$, 
the total entropy $h(\mathbf{y})$ is given, up~to a constant shift,
by $h(\mathbf{y}) = \sum_i h_i$, where $h_i = \log(\lambda_{i})$.

For each batch, we compute the sample covariance
matrix $c_{ij} = \operatorname{cov}(y_i, y_j)$ and~
sort its eigenvalues in descending order. We define 
a rank-$k$ approximation to the entropy as a 
weighted average of the per-component contributions to entropy
via
\begin{align} \label{low-rank-entropy}
    \bar{h}_k \triangleq \frac{\sum_{i=1}^d w_{ki}\log(\lambda_i)}{\sum_{i=1}^d w_{ki}}
\end{align}
where the weights provide a soft thresholding at $i=k$ via
\begin{align}\label{entropy-rank-weights}
    w_{ki} = \frac{1}{e^{\alpha (i - k)} + 1} 
\end{align}
with $\alpha$ a hyperparameter determining the smoothness of the
transition of relative weights from 1 to 0 as $i$ crosses $k$. 
Our experiments have shown consistent results for various
values of $\alpha>1$ (we chose $\alpha=3.3$).

Due to the normalization, \eqref{low-rank-entropy} should be thought of as a 
per-rank version of the low-rank entropy. In~
particular, when combining $\bar{h}_k$ with the marginal entropies of $y_i$
during the computation of the total correlation loss of 
Section~\ref{sec:locality-loss}, it is more appropriate to combine the former with 
the average marginal entropy rather than the total marginal~entropy.

\subsection[\appendixname~\thesubsection. KL Divergence~Estimation]{KL Divergence~Estimation}\label{app:kl-divergence-estimation}
\subsubsection[\appendixname~\thesubsubsection. KL Divergence of Marginal~Probabilities]{KL Divergence of Marginal~Probabilities}
Having the marginal probability density estimates 
\(\hat{p}_{i}\) and \(\hat{p}_{k}\) for components $Y_i$ and $Y_k$, we approximate the KL divergence 
\(D_{KL}(p_{i}, p_{k})\) by
\begin{equation}
D_{KL}(p_{i} \parallel p_{k}) \approx \mathbb{E} \left[\hat{p}_{i}(y_{i}) \log \hat{p}_{k}(y_{i})\right]
\end{equation}
where the expectation $\mathbb{E}$ is computed as an average over
the samples in a batch during~optimization.

\subsubsection[\appendixname~\thesubsubsection. KL Divergence of Conditional~Probabilities]{KL Divergence of Conditional~Probabilities}
Having the conditional probability density estimates 
\(\hat{p}_{i|j}\) and \(\hat{p}_{k|l}\) for components $Y_i$ and $Y_k$ where $y_j$ and $y_l$ are given, we approximate the KL divergence 
\(D_{KL}(p_{i|j}, p_{k|l})\) by
\begin{equation}
D_{KL}(p_{i|j} \parallel p_{k|l}) \approx \mathbb{E} \left[\hat{p}_{i|j}(y_{i}, y_{j}) \log \hat{p}_{k|l}(y_{i}, y_{j})\right]
\end{equation}
where the expectation $\mathbb{E}$ is computed as an average over
the samples in a batch during optimization.

\section[\appendixname~\thesection. Equivariance of the Group Convolution~Layer]{Equivariance of the Group Convolution~Layer}\label{app:equivariance_proof}
The equivariance properties of group convolutions are well known
\citep{cohen2016group}. In~this paper, we consider group actions (specifically representations) on the 
sample space that do not necessarily form a regular representation. 
In other words, the~group action does not come from an action 
on the index set labeling the components. Here, we show that
the equivariance property holds in our setting as well.
The components of the symmetry-based representation \(\rvector{y}\) are given as
\begin{align}
    \rvcomp{y}{p} = \sum_{i, j} \psi_i {(\rho^T(p))}_{ij} x_j
\end{align}
where $\rvector{x} \in \mathbb{R}^d$ represents the input signal, and~
${(\rho(p))}_{ij}$ denotes entries of the group representation matrix
for group element index $p$ (an integer index
for the case of the discrete version
of a 1-dimensional Lie group action). 
The action of the group element $s$ on the input vectors transforms $\mathbf{x}$ as 
$x_j \rightarrow {(\rho(s))}_{jk} x_k$. The~resulting action 
$\rho_y(s)$ on the 
symmetry-based representation \(\rvector{y}\) is given as
\begin{align}
    {(\rho_y(s) \cdot \mathbf{y})}_p &= \sum_{i, j, k} \vcomp{\psi^0}{i}{(\rho^T(p))}_{ij} {(\rho(s))}_{jk} x_k
    = \sum_{i, k} \vcomp{\psi^0}{i} {(\rho^T(p-s))}_{ik} x_k = y_{p-s} \\
    &\implies \rho_y(s) \cdot \mathbf{y} = T^s_Y \cdot \mathbf{y}
\end{align}
while \(T_Y\) is a one-component translation operator acting on the symmetry-based representation. Thus, the~proposed group convolution layer results in a group-equivariant representation for the symmetry group action, with~the actions of the group being turned to simple component translations in the symmetry-based representation. This property makes the output of the model suitable for use with the machinery of regular (translation-based)~CNNs.


\section[\appendixname~\thesection. Datasets]{Datasets}\label{app:datasets}
We trained our model on a variety of synthetic and real datasets whose properties 
are given in Table~\ref{tab:datasets}. 

\vspace{-2pt}
\begin{table}[H]
\caption{Real and synthetic datasets prepared for experiments.
The parameters for each basis signal are sampled from a uniform 
distribution with the indicated ranges. }
\label{tab:datasets}
\begin{adjustwidth}{-\extralength}{0cm}\centering

\begin{tabular*}{\fulllength}{c@{\extracolsep{\fill}}ccccc}
  \toprule
  \textbf{ID} & \textbf{Signal Type} & \textbf{Invariance} & \textbf{Dimensionality} & \textbf{Parameter Ranges} & \textbf{Size (Samples)} \\
  \midrule
  1 & \makecell{Gaussian} & \makecell{Circulant\\translation} & \makecell{7} & \makecell{Scale: $[0.2, 1.0)$\\Amplitude: $[0.5, 1.5)$} & \makecell{63 K} \\
  \midrule
  2 & \makecell{Gaussian} & \makecell{Translation} & \makecell{7} & \makecell{Scale: $[0.2, 1.0)$\\Amplitude: $[0.5, 1.5)$} & \makecell{252 K} \\
  \midrule
  3 & \makecell{Mnist\\slices} & \makecell{Translation} & \makecell{27} & \makecell{-------} & \makecell{1.08 M} \\
  \midrule
  4 & \makecell{Gaussian} & \makecell{Translation} & \makecell{33} & \makecell{Scale: $[0.5, 5.0)$\\Amplitude: $[0.5, 1.5)$} & \makecell{1.65 M} \\
  \midrule
  5 & \makecell{Legendre\\($l=2\text{--}3$, $m=1$)} & \makecell{Translation} & \makecell{33} & \makecell{Scale: $[6.0, 15.0)$\\Amplitude: $[0.5, 1.5)$} & \makecell{1.65 M} \\
  \midrule
  6 & \makecell{Gaussian} & \makecell{Permuted\\translation} & \makecell{33} & \makecell{Scale: $[0.5, 5.0)$\\Amplitude: $[0.5, 1.5)$} & \makecell{1.65 M} \\
  \midrule
  7 & \makecell{Legendre\\($l=2\text{--}3$, $m=1$)} & \makecell{Permuted\\translation}& \makecell{33} & \makecell{Scale: $[6.0, 15.0)$\\Amplitude: $[0.5, 1.5)$} & \makecell{1.65 M} \\
  \midrule
  8 &\makecell{Gaussian} & \makecell{Frequency\\shift} & \makecell{33} & \makecell{Scale: $[0.5, 5.0)$\\Amplitude: $[0.5, 1.5)$} & \makecell{1.65 M} \\
  \midrule
  9 &\makecell{Legendre\\($l=2\text{--}3$, $m=1$)} & \makecell{Frequency\\shift} & \makecell{33} & \makecell{Scale: $[6.0, 15.0)$\\Amplitude: $[0.5, 1.5)$} & \makecell{1.65 M} \\
  \bottomrule
\end{tabular*}
\end{adjustwidth}
\end{table}

\subsection[\appendixname~\thesubsection. Details of Synthetic~Data]{Details of Synthetic~Data}\label{app:synthetic_data}
We start with creating a dataset that is local/symmetric under component translations. Then, we~apply an appropriate transformation to obtain datasets that are local and symmetric under a given unitary representation. Detailed procedure is as follows:

\begin{itemize}
    \item We select a parametrized family of local basis signals, such as the family of Gaussians parametrized by amplitude, center, and~width.
    \item Use a binomial distribution (probability $p=0.5$; $n=5$ trials) to~determine the number of local signals to include in each sample.
    \item Sample the parameters of the basis signal (e.g., center, width, and amplitude) from uniform distributions over finite ranges to obtain each local signal.
    \item Add up the local pieces to end up with a single sample that has information locality under component translations (we call this the \emph{raw sample
}).
    \item We apply an appropriate unitary transformation to the raw sample to obtain locality/symmetry under the desired group action. 
    \item Finally, we add Gaussian noise to the sample (with \(\sigma = 0.05\)).
\end{itemize}

This procedure can provide us any symmetry that is related to component translations via a similarity transformation. See Figure~\ref{fig:dataset-samples} involving samples from datasets with different kinds of~symmetries. 

\subsubsection[\appendixname~\thesubsubsection. Basis signal~types]{Basis signal~types}
\textbf{Gaussian signals 
}
Gaussian basis signals $f_{gaussian}(z;\mathcal{A}, 
\mu, \sigma)$ are parametrized by amplitude \(\mathcal{A}\), center \(\mu\), and~width \(\sigma\). 
The input $z$ is an integer ranging from 
$-\frac{d}{2}$ to $\frac{d}{2}$, labeling the components of the raw 
sample vectors.
We sample the center $\mu$ uniformly from the extended 
(tripled) range $-\frac{3d}{2}$ to $\frac{3d}{2}$ and~then crop the 
resulting  signals to the $z$ range of $-\frac{d}{2}$ to $\frac{d}{2}$ 
to allow for the possibility of signals that contain 
only a tail of a~Gaussian.

\textbf{Legendre signals}
These signals are given in terms of the associated Legendre polynomials
and provide localized waveforms that can change signs.
The relevant parameters are center \(c\), scale \(s\), amplitude \(\mathcal{A}\), and~the orders $l$, $m$:
$f^{(l,m)}_{legendre}(z;\mathcal{A}, c, s) =
\mathcal{A}P_l^m\left(\cos\left(\frac{z-c}{s}\right)\right)$.
We crop these signals to the range 
$|x-c|/s \le \pi$, i.e.,~set the values outside this
range to~zero. 

Once again, $z$ becomes the discrete dimension index, ranging from $-\frac{d}{2}$ to $\frac{d}{2}$. For~the $l,m$ parameters,
we use $l=2, m=1$ and $l=3, m=1$, with~equal probability for each sample. We sample
the centers as in the Gaussian~case.

\subsection[\appendixname~\thesubsection. Details of Real~Data]{Details of Real~Data}\label{app:real_data}
Symmetry--locality coupling is naturally found in image datasets as well as time-series 
datasets. To~demonstrate the generality of the proposed method, we use a cropped version of 
the MNIST dataset, which consists of $28\times 28$ grayscale images of handwritten digits. 
In order to enable uniform sampling, we zero-pad the MNIST dataset from 
the left and the right with 28 pixels, leading to images with dimensions of \(84 \times 
28\). Afterwards, we randomly sample \(27 \times 1\)-dimensional patches from the images, 
the coordinates of the patches being sampled uniformly from the image area. Finally, we 
drop the blank crops (crops with all pixel values equal to zero). This way, we obtain a 
dataset with 1.08 M 
 samples, which we use to train our model. 

\section[\appendixname~\thesection. Details of the Comparison with GAN-Based~Methods]{Details of the Comparison with GAN-Based~Methods}\label{app:gan-comparison}

While our method is set up to learn the minimal generator of 
the symmetry group, LieGAN \citep{yang2023generative}
aims to learn Lie algebra elements of 
the underlying Lie group, and~ SymmetryGAN \citep{desai2022symmetry}
aims to find \emph{some} 
group element. This difference makes it difficult to set up a 
numerical comparison of our method with the GAN-based methods.
We approach this  problem as follows. The~minimal group generator 
for an underlying 1-dimensional symmetry will be some power of the 
exponential of the corresponding Lie algebra element. If~LieGAN indeed
learns a correct Lie algebra element, then an appropriate power of its exponential 
should provide the minimal generator exactly. Thus, we find the multiple of the 
candidate Lie algebra generator by LieGAN whose exponential
 provides the best approximation (in the sense of 
cosine similarity) for the symmetry generator and~use its cosine similarity
as the score for LieGAN. In~the case of SymmetryGAN, 
we find the true group element that is closest to the candidate symmetry 
transformation learned by SymmetryGAN and~compare the cosine similarity 
between this true symmetry and the candidate, and~then perform a 
similar comparison to the corresponding symmetry element from our~model. 

The SymmetryGAN method \citep{desai2022symmetry} was based on a parametrization
of the group elements, but~the authors had not formulated a parametrization 
for datasets with dimensions higher than \(4\) considering the availability of
factorization-based parametrizations. Based on this, we used a QR-decomposition-based
parametrization to implement the symmetry generator in SymmetryGAN. We call
this extended method ``SymmetryGAN-QR'' considering that the choice of parametrization 
can effect the performance of the method. We experimented with LieGAN
\citep{yang2023generative} without any~modification. 

We conducted the experiments over a synthetic dataset containing \(2.56\) M
samples, which is the size of the largest dataset used in the experiments
reported in \citep{yang2023generative}. We used the same hyperparameter
settings used in \citep{yang2023generative}, \citep{desai2022symmetry}.
We trained both methods for 100 epochs and~took the best-scoring epoch
comparison with our method (thus, in~a sense, giving unfair advantage to 
both methods since the scores are computed by using the ground~truth).

We trained our method with a much  smaller sample from the same dataset, containing
 \(63\)K examples. Using 4000 epochs, we approximately equalized the~total number of examples for the two approaches.


We observe that the performance of the GAN-based methods was poorer than
the results reported in the corresponding papers. The~experiments reported 
in the respective papers were based on dataset dimensionalities up to $4$. The~number of 
parameters to learn increases quadratically with the dimension (\({d(d-1)}/{2}\) parameters 
are needed in \(d\) dimensions), so a \(7\)-dimensional symmetry-learning task is far 
more challenging than a \(4\)~dimensional one, which we suspect is the reason 
for the reduction in performance. This also puts into context 
the success of our method on $33$-dimensional datasets reported~above.


\section[\appendixname~\thesection. Hyperparameters and Loss~Terms]{Hyperparameters and Loss~Terms}\label{app:hyperparameters}
To choose our hyperparameter settings, we first ran a few experiments
to pick a parameter combination for the \(33\)-dimensional ``Legendre'' dataset with 
translation symmetry (see Appendix \ref{app:datasets} for dataset details) using a 
coarse tuning procedure. 
We then tried this same parameter combination with the
other datasets and learned that it provided similarly satisfactory results, making further dataset-specific
parameter tuning unnecessary. We then performed a sensitivity analysis (see Section 
 \ref{app:morris})
to confirm that the performance of the system is indeed relatively insensitive to hyperparameters 
around our settings. Finally, we performed an ablation study to demonstrate that
all the terms in the loss function indeed contribute to the success of the~method.

\subsection[\appendixname~\thesubsection. Initial choice of~hyperparameters]{Initial choice of~hyperparameters}\label{app:training-details}

Other than the learning rate, the~ADAM optimizer was used with the default
hyperparameters (\(\beta_1 = 0.9, \beta_2 = 0.999, \epsilon=10^{-7}\)) for both the model and the estimators. 
Our choice of the hyperparameters can be found in Table~\ref{tab:hyperparameters}.  

\begin{table}[H]
\caption{Hyperparameters 
 for our~method.}
\label{tab:hyperparameters}
\begin{adjustwidth}{-\extralength}{0cm}\centering
\begin{tabular*}{\fulllength}{c@{\extracolsep{\fill}}ccccccccc}
\toprule
\makecell{\textbf{Estimator}\\\textbf{Learning}\\\textbf{Rate}\\\((\times 10^{-3})\)} & \makecell{\textbf{Model}\\\textbf{Learning}\\\textbf{Rate}\\\((\times 10^{-3})\)} & \makecell{\textbf{Total}\\\textbf{Learning}\\\textbf{Rate}\\\textbf{Decay}} & \makecell{\textbf{Alignment}\\\textbf{Coefficient}} & \makecell{\textbf{Uniformity}\\\textbf{Coefficient}} & \makecell{\textbf{Resolution}\\\textbf{Coefficient}} & \makecell{\textbf{Information}\\\textbf{Preservation}\\\textbf{Coefficient}} \\
\midrule
2.5 & 0.1 & 0.1 & 1.0 & 2.0 & 1.0 & 2.0 \\
\bottomrule
\end{tabular*}
\end{adjustwidth}
\end{table}

With these settings, experiments were run on NVIDIA GTX4090 GPUs using the Tensorflow library for 
implementation. With~these settings, the~training took approximately 2.5 h for the 7-dimensional 
datasets, 16.5 h for the 27-dimensional sliced MNIST dataset, and~28 h for the \(33\)-dimensional~datasets.


\subsection[\appendixname~\thesubsection. Elementary Effect Sensitivity~Analysis]{Elementary Effect Sensitivity~Analysis}
\label{app:morris}
To investigate how various hyperparameters affect the performance of the method,
we performed a sensitivity analysis. We conducted 36 experiments whose parameters were sampled using 
SALib (version 1.47), Sensitivity Analysis Library in Python \citep{Herman2017,Iwanaga2022}. For~each parameter, we used
a sampling interval of $\pm 25\%$. The~Morris method 
\citep{morris1991factorial} was applied to evaluate the influence of various parameters
on the cosine similarity between the learned symmetry generator and the ideal minimal generator. 
The method is based on the notion of an \emph{effect} \(d_i(X_1, X_2, \cdots, X_n)\), 
which is defined as
\begin{align}
    d_i(X_1, X_2, \cdots, X_n) = \frac{f(X_1, X_2, \cdots, X_i + \Delta, \cdots, X_n) - f(X_1, X_2, \cdots, X_i, \cdots, X_n)}{\Delta} 
\end{align}
for a function \(f(X_1, X_2, \cdots, X_n)\), where \(\{X_j\}_{j=1}^n \in\mathbb{R}\) are the parameter values, and~\(\Delta\) is a step size. The~key idea is to measure the changes 
of the function $f$ along various trajectories, and~to compute 
quantities measuring the overall effect of the parameters on the function values. 
In our case, the~function $f$ is the aforementioned cosine similarity.
The important metrics~are
\begin{itemize}
    \item $\mu_i$: Mean effect \(\mathbb{E}[d_i]\);
    \item $\mu^*_i$: Mean absolute effect \(\mathbb{E}[|d_i|]\);
    \item $\sigma_i$: Standard deviation \(\sqrt{\mathbb{E}[(d_i-\mu_i)^2]}\);
\end{itemize}
where \(\mathbb{E}[.]\) denotes the average over~trajectories.

\begin{table}[H]
\caption{Elementary effect analysis results. Terms are sorted from top to bottom based on their~significance.}
\label{tab:morris-results}
\begin{tabular*}{\hsize}{l@{\extracolsep{\fill}}cccc}
\toprule
\textbf{Parameter} & \boldmath$\mu$ & \boldmath$\mu^*$ [\textbf{95\% CI}] & \boldmath$\sigma$ \\
\midrule

Estimator learning rate & 0.08981 & 0.09123 [0.08902] & 0.10771 \\
Model learning rate     & $-$0.03309 & 0.04060 [0.04458] & 0.06588 \\
Total learning rate decay     & $-$0.03529 & 0.03682 [0.05176] & 0.06890 \\
Alignment coefficient    & 0.02523 & 0.05384 [0.05455] & 0.08922 \\
Uniformity coefficient   & 0.05345 & 0.05619 [0.09583] & 0.10371 \\
Resolution coefficient   & $-$0.04455 & 0.04938 [0.06613] & 0.08628 \\
Information preservation coefficient & 0.07904 & 0.07909 [0.09634] & 0.11925 \\
Noise                   & $-$0.00298 & 0.01124 [0.00578] & 0.01424 \\
\bottomrule
\end{tabular*}
\end{table}

In Table~\ref{tab:morris-parameters}, we show the trajectories used in our experiments, and, in Table~\ref{tab:morris-results}, we show the effects of various hyperparamaters on the 
model performance. Looking at the cosine similarities and the
effect values, we see that, within the range of values used,
the results are quite stable. This explains the fact that 
we were able to obtain similar performance on all of our different
datasets despite using the same hyperparameter settings in all 
experiments without any per-dataset~tuning. 

\begin{table}[H]
\caption{Experiment configurations for elementary effect analysis. 
During sensitivity experiments, we used \(7\)-dimensional translation-invariant Gaussian datasets composed of \(252\) K samples.}
\label{tab:morris-parameters}
\begin{adjustwidth}{-\extralength}{0cm}\centering

\begin{tabular*}{\fulllength}{l@{\extracolsep{\fill}}ccccccccc}
\toprule
  & \makecell{\textbf{Estimator}\\\textbf{Learning}\\ \textbf{Rate}\\\((\times 10^{-3})\)} & \makecell{\textbf{Model}\\\textbf{Learning}\\\textbf{Rate}\\\((\times 10^{-3})\)} & \makecell{\textbf{Total}\\\textbf{Learning}\\\textbf{Rate}\\\textbf{Decay}} & \makecell{\textbf{Alignment}\\\textbf{Coefficient}} & \makecell{\textbf{Uniformity}\\\textbf{Coefficient}} & \makecell{\textbf{Resolution}\\\textbf{Coefficient}} & \makecell{\textbf{Information}\\\textbf{Preservation}\\\textbf{Coefficient}} & \makecell{\textbf{Noise}} & \makecell{\textbf{Cosine}\\\textbf{Similarity}}\\
\midrule
1&2.292&0.075&0.167&1.083&2.167&1.250&0.917&0.133&0.816 \\
2&3.125&0.075&0.167&1.083&2.167&1.250&0.917&0.133&0.955 \\
3&3.125&0.075&0.167&1.083&1.500&1.250&0.917&0.133&0.815 \\
4&3.125&0.075&0.167&1.083&1.500&1.250&0.917&0.000&0.827 \\
5&3.125&0.075&0.167&1.083&1.500&0.917&0.917&0.000&0.943 \\
6&3.125&0.108&0.167&1.083&1.500&0.917&0.917&0.000&0.856 \\
7&3.125&0.108&0.100&1.083&1.500&0.917&0.917&0.000&0.948 \\
8&3.125&0.108&0.100&0.750&1.500&0.917&0.917&0.000&0.843 \\
9&3.125&0.108&0.100&0.750&1.500&0.917&1.250&0.000&0.999 \\
10&1.875&0.092&0.167&1.083&2.500&1.083&1.083&0.067&0.885 \\
11&2.708&0.092&0.167&1.083&2.500&1.083&1.083&0.067&0.988 \\
12&2.708&0.092&0.167&1.083&2.500&0.750&1.083&0.067&0.982 \\
13&2.708&0.092&0.167&1.083&1.833&0.750&1.083&0.067&0.976 \\
14&2.708&0.092&0.167&0.750&1.833&0.750&1.083&0.067&0.997 \\
15&2.708&0.092&0.167&0.750&1.833&0.750&0.750&0.067&0.997 \\
16&2.708&0.125&0.167&0.750&1.833&0.750&0.750&0.067&0.986 \\
17&2.708&0.125&0.167&0.750&1.833&0.750&0.750&0.200&0.997 \\
18&2.708&0.125&0.100&0.750&1.833&0.750&0.750&0.200&0.997 \\
19&3.125&0.125&0.100&0.750&2.500&1.083&1.083&0.000&0.990 \\
20&3.125&0.125&0.100&0.750&2.500&0.750&1.083&0.000&0.999 \\
21&3.125&0.092&0.100&0.750&2.500&0.750&1.083&0.000&0.992 \\
22&3.125&0.092&0.100&1.083&2.500&0.750&1.083&0.000&0.988 \\
23&3.125&0.092&0.100&1.083&1.833&0.750&1.083&0.000&0.987 \\
24&2.292&0.092&0.100&1.083&1.833&0.750&1.083&0.000&0.987 \\
25&2.292&0.092&0.167&1.083&1.833&0.750&1.083&0.000&0.997 \\
26&2.292&0.092&0.167&1.083&1.833&0.750&1.083&0.133&0.994 \\
27&2.292&0.092&0.167&1.083&1.833&0.750&0.750&0.133&0.993 \\
28&3.125&0.108&0.133&0.917&2.167&1.083&1.083&0.133&0.997 \\
29&3.125&0.108&0.200&0.917&2.167&1.083&1.083&0.133&0.994 \\
30&3.125&0.108&0.200&0.917&2.167&0.750&1.083&0.133&0.993 \\
31&3.125&0.075&0.200&0.917&2.167&0.750&1.083&0.133&0.991 \\
32&3.125&0.075&0.200&0.917&1.500&0.750&1.083&0.133&0.994 \\
33&2.292&0.075&0.200&0.917&1.500&0.750&1.083&0.133&0.995 \\
34&2.292&0.075&0.200&1.250&1.500&0.750&1.083&0.133&0.982 \\
35&2.292&0.075&0.200&1.250&1.500&0.750&1.083&0.000&0.987 \\
36&2.292&0.075&0.200&1.250&1.500&0.750&0.750&0.000&0.818 \\
\bottomrule
\end{tabular*}
\end{adjustwidth}
\end{table}




\subsection[\appendixname~\thesubsection. Ablation~Study]{Ablation~Study}\label{app:ablation-study}
\begin{figure}[H]
    \subfloat[\centering \label{fig:no-alignment-translation-generator}]{
        \includegraphics[width=0.48\textwidth]{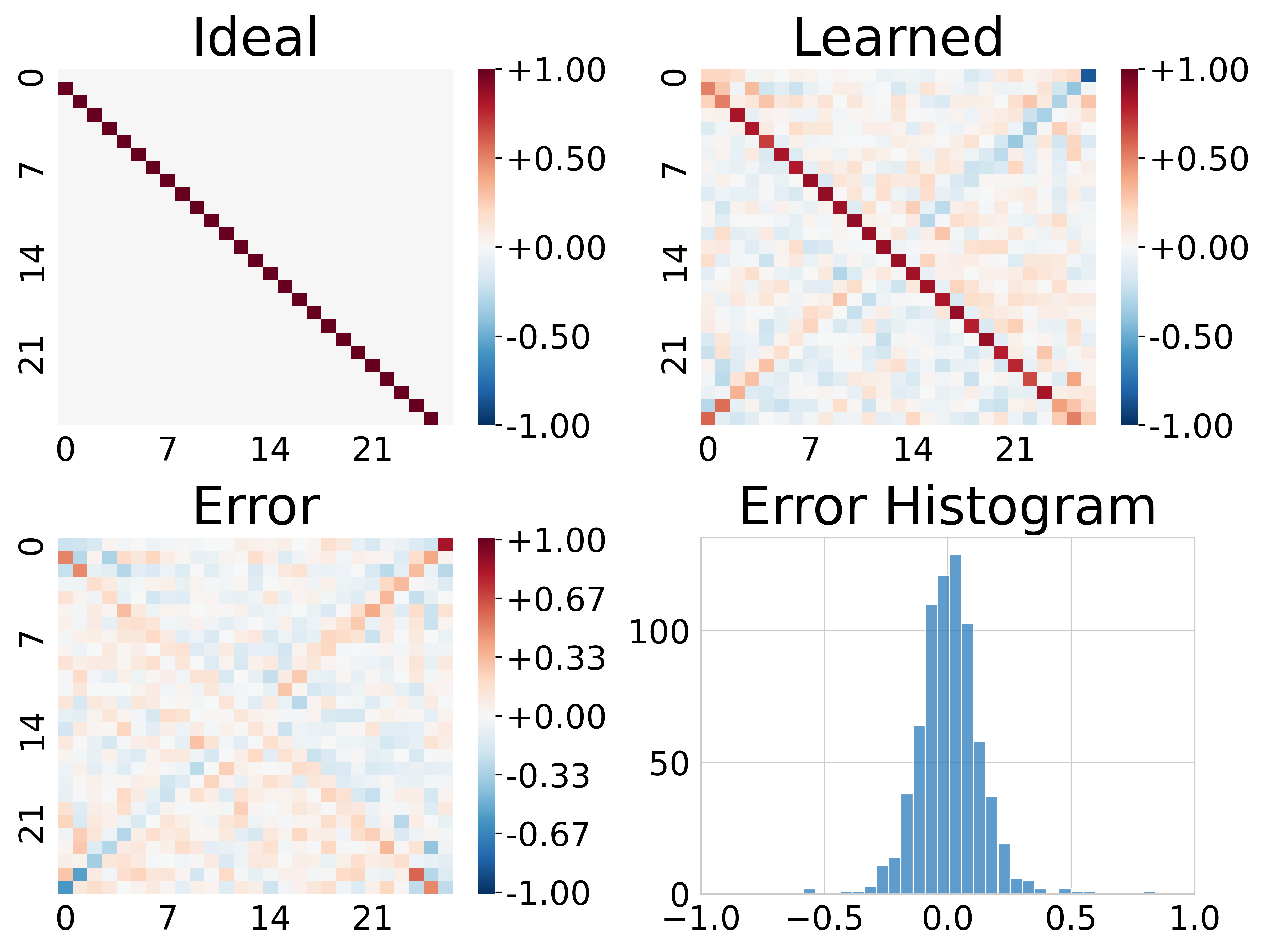}
    }
    \hfill
    \subfloat[\centering \label{fig:no-alignment-group-convolution-matrix}]{
        \includegraphics[width=0.48\textwidth]{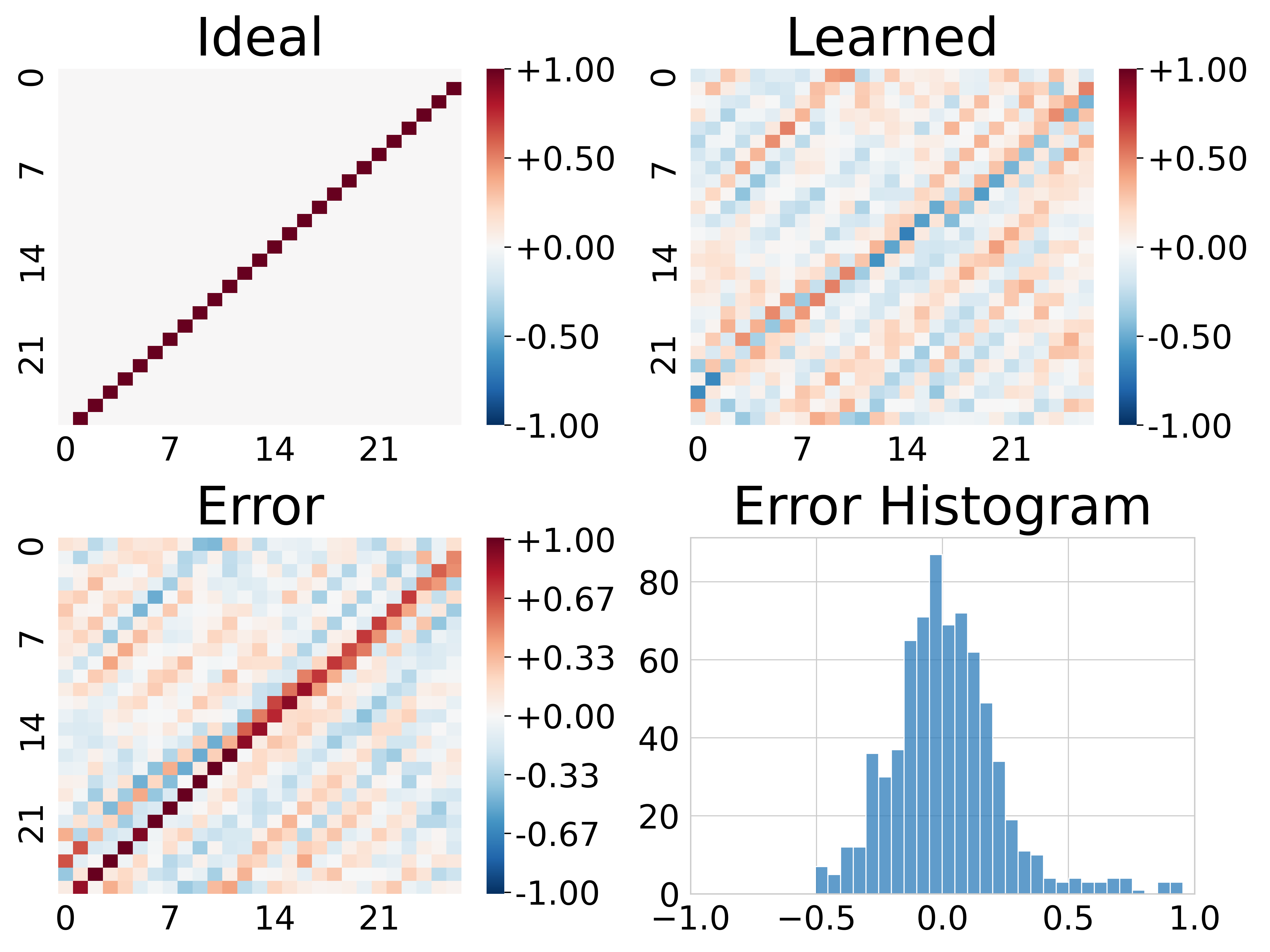}
    }
    \caption{Results 
 of the ablation experiment where the 
    alignment term is excluded from the loss function. The~system is still able to recover the symmetry generator to some extent, but~the performance is poor; the cosine similarity is reduced to 0.770 from  0.999, which was the result we had with the full loss function, reported  in Table~\ref{tab:cosine-similarity-generator}. The~lack of the alignment term results in a group convolution matrix that
    does not quite respect locality. (\textbf{a}) Ideal and learned symmetry generators (\textbf{top}) and error distributions (\textbf{bottom}). (\textbf{b}) Ideal and learned group convolution matrices (\textbf{top}) and error distributions (\textbf{bottom}).}\label{fig:ablation-study-no-alignment}
\end{figure}

\begin{figure}[H]
    \subfloat[\centering \label{fig:no-resolution-translation-generator}]{
        \includegraphics[width=0.48\textwidth]{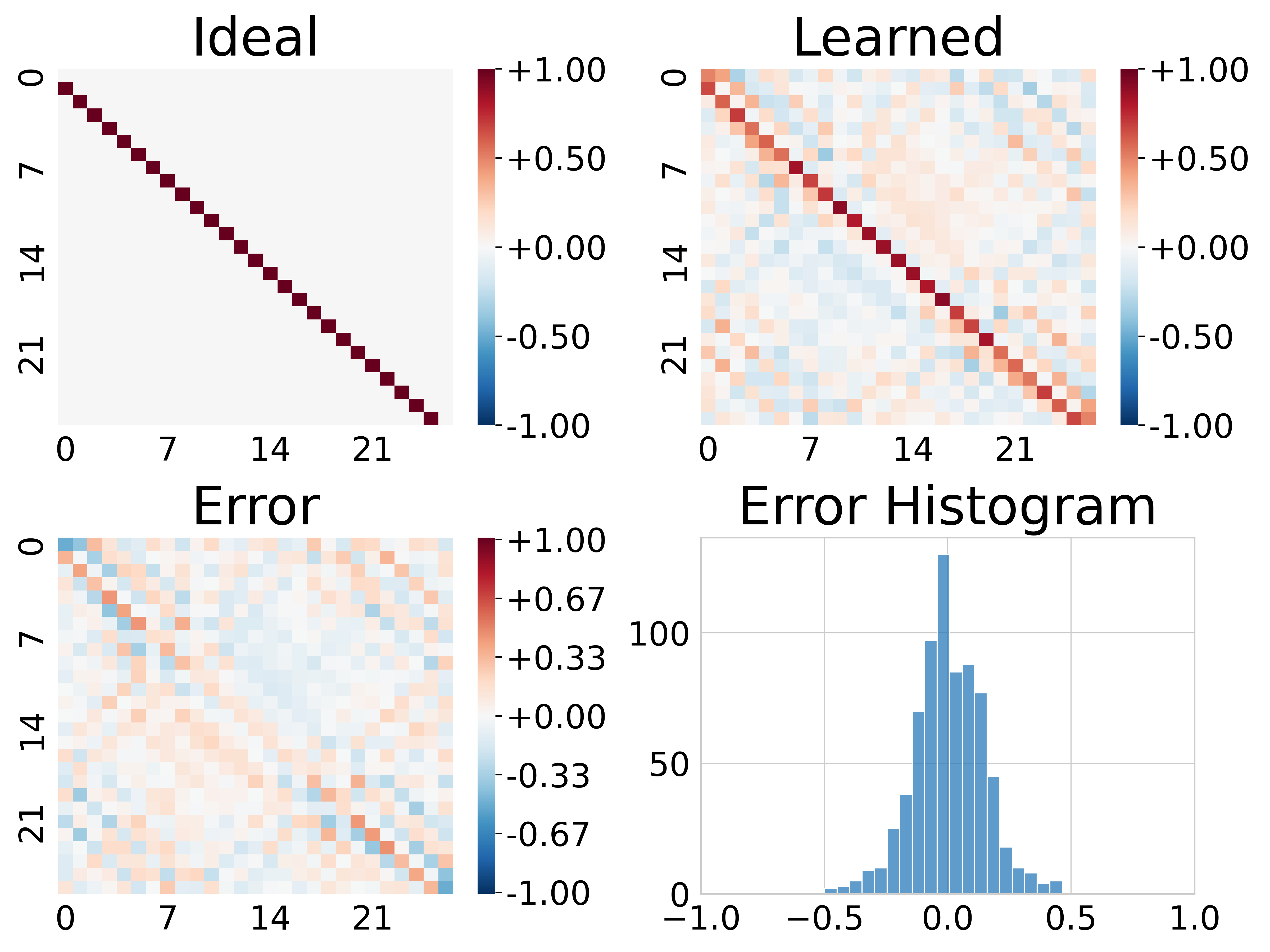}
    }
    \hfill
    \subfloat[\centering \label{fig:no-resolution-group-convolution-matrix}]{
        \includegraphics[width=0.48\textwidth]{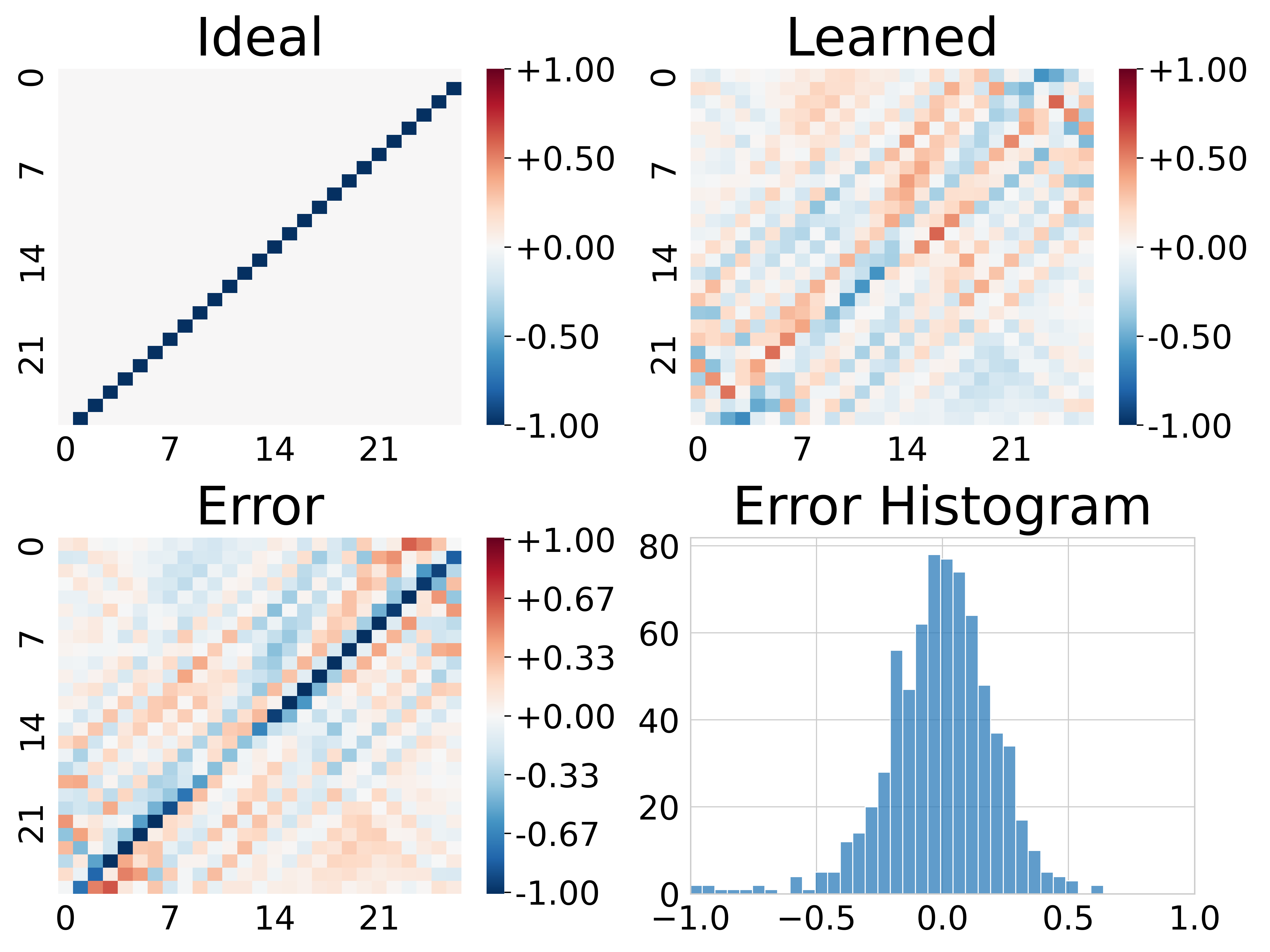}
    }
    \caption{Results 
 of the ablation experiment where the 
    resolution term is excluded from the loss function. As~in Figure~\ref{fig:ablation-study-no-alignment},
    the group convolution matrix mixes different components of input data, not quite respecting locality. This behavior is consistent with the assumption that the alignment and resolution terms together induce locality. The~cosine similarity score is 0.713, compared to the value of  0.999 reported in Table~\ref{tab:cosine-similarity-generator}. (\textbf{a}) Ideal and learned symmetry generators (\textbf{top}) and error distributions (\textbf{bottom}). (\textbf{b}) Ideal and learned group convolution matrices (\textbf{top}) and error distributions (\textbf{bottom}).}
    \label{fig:ablation-study-no-resolution}
\end{figure}

\begin{figure}[H]
    \subfloat[\centering \label{fig:no-infomax-translation-generator}]{
        \includegraphics[width=0.48\textwidth]{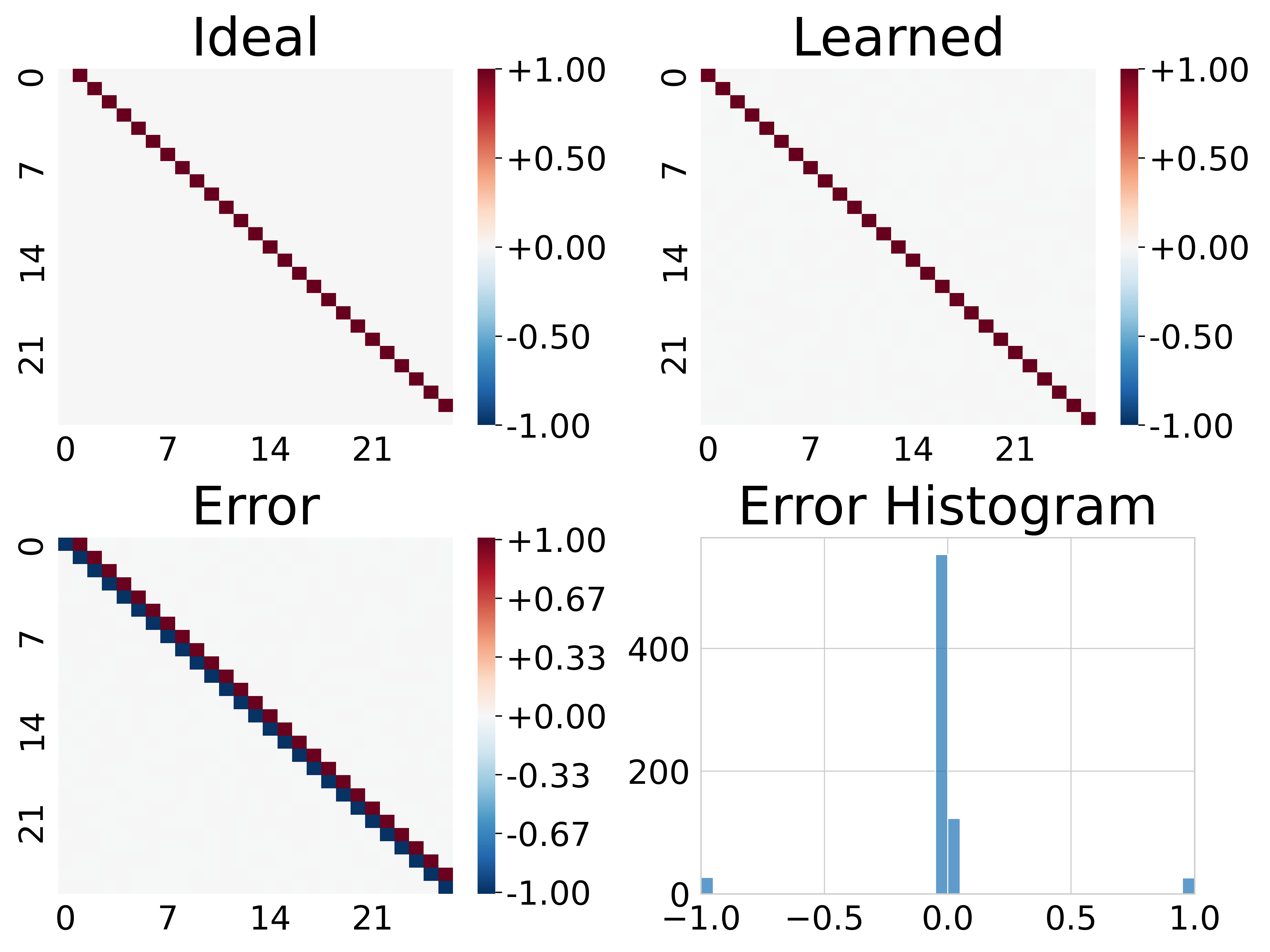}
    }
    \hfill
    \subfloat[\centering \label{fig:no-infomax-group-convolution-matrix}]{
        \includegraphics[width=0.48\textwidth]{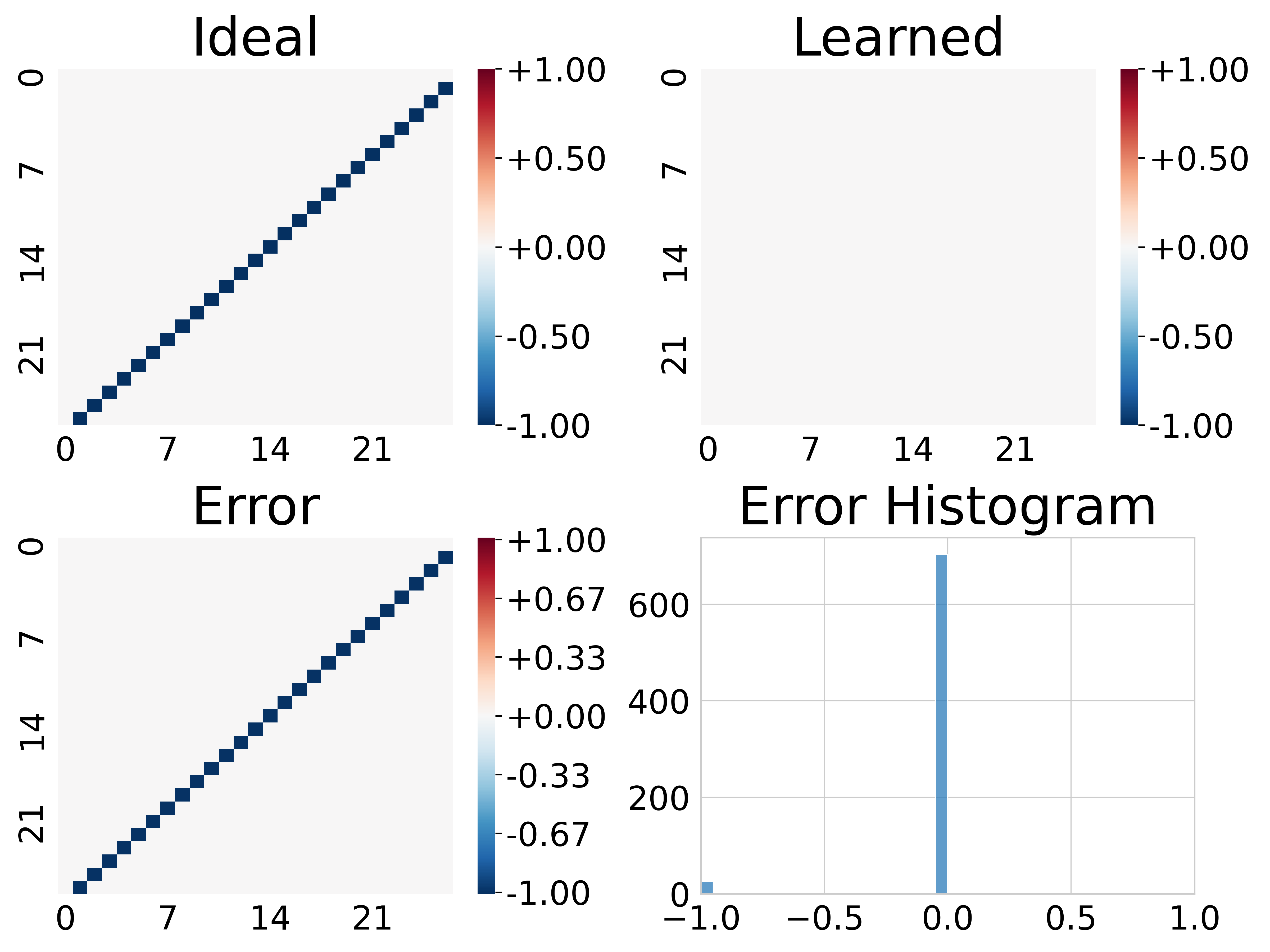}
    }
    \caption{Results 
 of the ablation experiment where the 
    information preservation term is excluded from the loss function. In~this case,  the~group convolution matrix becomes the zero matrix. This catastrophic solution saturates the uniformity loss maximally, as~well as the resolution term (when all marginal and joint entropies are zero, the~resolution term has its lowest value). Without~the term enforcing information preservation, 
    the system naturally collapses to this singular solution.
    In this case, symmetry generator does not have any effect over the output representation, and~the gradient of the loss with respect to it vanishes; therefore, the symmetry generator stays close to its initial value, which is close to the identity matrix. The~cosine similarity score in this case is 0.001, demonstrating inability to discover symmetry generator. (\textbf{a}) Ideal and learned symmetry generators (\textbf{top}) and error distributions (\textbf{bottom}). (\textbf{b}) Ideal and learned group convolution matrices (\textbf{top}) and error distributions (\textbf{bottom}).
    } 
    \label{fig:ablation-study-no-infomax}
\end{figure}

\begin{figure}[H]
    \subfloat[\centering \label{fig:no-uniformity-translation-generator}]{
        \includegraphics[width=0.48\textwidth]{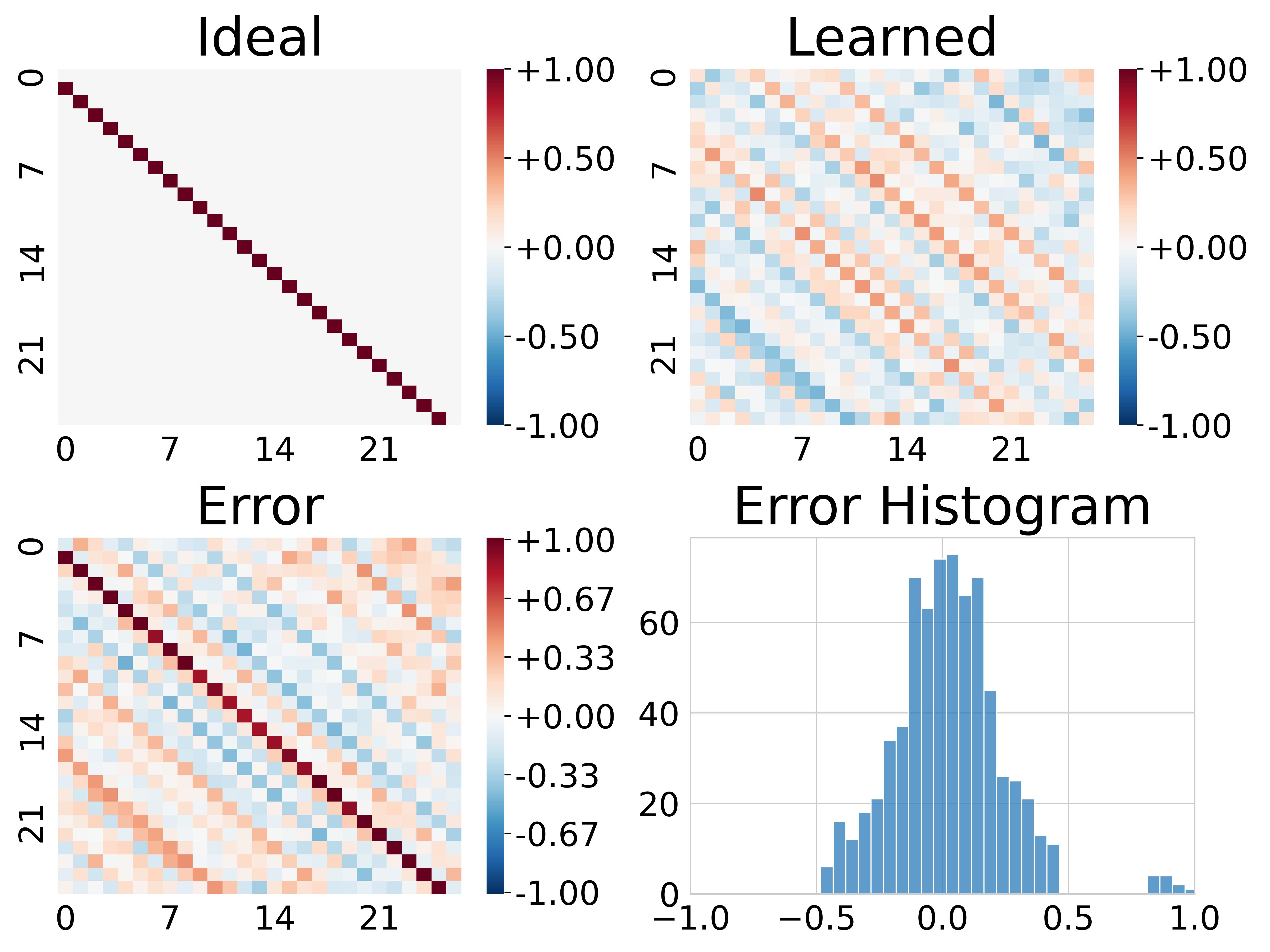}
    }
    \hfill
    \subfloat[\centering \label{fig:no-uniformity-group-convolution-matrix}]{
        \includegraphics[width=0.48\textwidth]{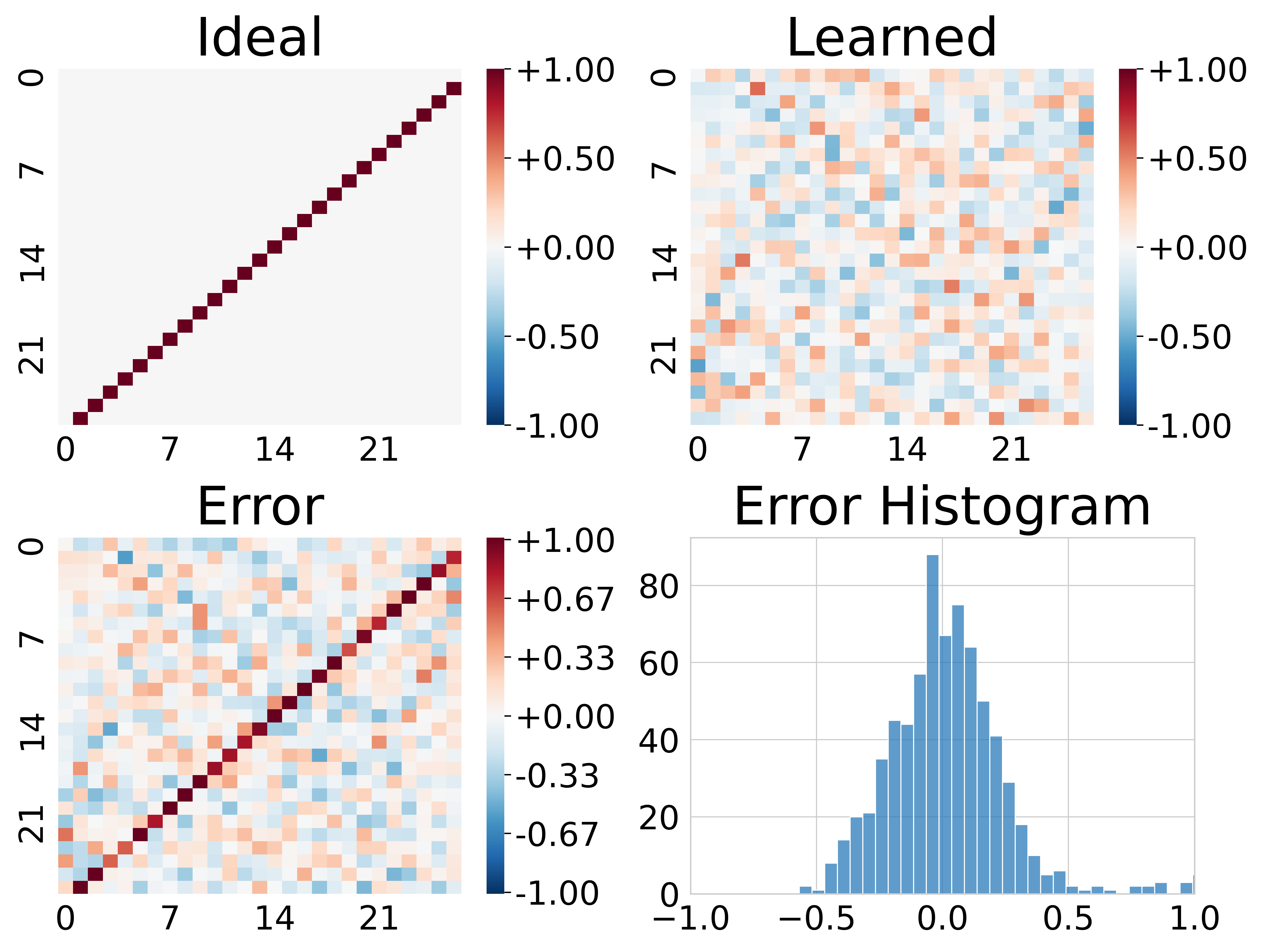}
    }
    \caption{Results 
 of the ablation experiment where the 
    uniformity term is excluded from the loss function. The~group convolution matrix has no regular structure and mixes the different components of input data. Moreover, the~model is unable to learn the true symmetry generator even partially, which could be expected since the uniformity term has a primary role in our formulation of symmetry. The~cosine similarity score in this case is $-0.037$, and the model is unable to discover the symmetry generator even in a partial manner. (\textbf{a}) Ideal and learned symmetry generators (\textbf{top}) and error distributions (\textbf{bottom}). (\textbf{b}) Ideal and learned group convolution matrices (\textbf{top}) and error distributions (\textbf{bottom}).}
    \label{fig:ablation-study-no-uniformity}
\end{figure}

To demonstrate that each piece of our loss function indeed
contributes to the success of the method, we conducted ablation experiments 
by dropping each piece and training. For~these experiments, we use 
the MNIST dataset (see Section \ref{app:real_data}). See Figures~\ref{fig:ablation-study-no-alignment}--\ref{fig:ablation-study-no-uniformity} 
for the results. It is clear that all 
components of the loss term indeed contribute to the performance,
working in a complementary~manner.

\vspace{-15pt}

\vspace{-15pt}

\vspace{-15pt}

\section[\appendixname~\thesection. Complexity~Analysis]{Complexity~Analysis}\label{app:complexity-analysis}
\subsection[\appendixname~\thesubsection. Time-Complexity]{Time Complexity}

See Table~\ref{tab:time-complexities} for a description of the computational complexity of our method.
The number \(d\) denotes the dimensionality, 
 \(k\) denotes the number of Gaussian kernels used for each
 pixel probability density estimator, and~\(h\) denotes the size of 
 the hidden layer used in the conditional probability estimator. In~
 our experiments, we set \(k=4\) and \(h = 4 \times k \times d\),
 and the dimensionalities of our datasets were
 \(7\), \(27\), and \(33\).
The most costly operations in our algorithm are eigendecompositions 
used in two steps, which lead to an overall time complexity of \(O(d^3)\). Although~insignificant in \(33\) dimensions, for~datasets 
with higher dimensionalities, this would be a significant burden. 
Conditional probability estimation is another expensive operation whose time complexity is  \(O(d \times h \times k)\),
which was  \(O(d^2)\) in our experiments since we took 
\(h\) proportional to \(d\). There could be room for improvement here
by  choosing smaller hidden layer~size.



\begin{table}[H]
\caption{Time complexities of the building blocks of our~method.}
\label{tab:time-complexities}
\begin{adjustwidth}{-\extralength}{0cm}\centering

\begin{tabular*}{\fulllength}{c@{\extracolsep{\fill}}cc}
\toprule
\textbf{Operation} & \textbf{Description} & \textbf{Time Complexity} \\
\midrule
\makecell{Probability estimation} & \makecell{Estimation of per-component probabilities\\via Gaussian kernels.} & O($d \times k$) \\
\midrule
\makecell{Conditional probability\\estimation} & \makecell{Estimation of conditional probabilities\\via Gaussian kernels parametrized with\\three neural networks.} & O($d \times h \times k$) \\
\midrule
\makecell{Forming the group convolution\\matrix} & \makecell{Formed by applying the generator\\to the resolving filter. We use an 
eigendecomposition-\\based algorithm for efficiency.} & O($d^3$) \\
\midrule
\makecell{Joint entropy computation\\(covariance step)} & \makecell{Computing the covariance matrix.} & O($d^2$)\\
\midrule
\makecell{Joint entropy computation\\(eigendecomposition step)} & \makecell{Eigendecomposition of the covariance matrix.} & O($d^3$)\\
\bottomrule
\end{tabular*}
\end{adjustwidth}
\end{table}
\unskip

\subsection[\appendixname~\thesubsection. Space~Complexity]{Space~Complexity}
The space complexity of our method is \(O(d^2)\), with~the
dominant components being the conditional probability estimator and 
the symmetry generator. See Table~\ref{tab:space-complexity-analysis} for~details. 

\begin{table}[H]
\caption{Space complexities for modules. We exclude temporary memory usage since it depends on the specifics of the library used
and the training setup (optimizer, etc.).}
\label{tab:space-complexity-analysis}
\begin{adjustwidth}{-\extralength}{0cm}\centering

\begin{tabular*}{\fulllength}{c@{\extracolsep{\fill}}cc}
\toprule
\textbf{Object} & \textbf{Description} & \textbf{Space Complexity} \\
\midrule
\makecell{Probability estimator} & \makecell{Estimation of probabilities\\via Gaussian kernels.} & O($d \times k$) \\
\midrule
\makecell{Conditional probability\\estimator} & \makecell{Estimation of conditional probabilities\\via Gaussian kernels parametrized by
\\three neural networks.} & O($d \times h \times k$) \\
\midrule
\makecell{Generator} & \makecell{Formed by applying the symmetry 
generator\\to the resolving filter. We use an
eigendecomposition-\\based algorithm for efficiency.} & O($d^2$) \\
\midrule
\makecell{Resolving filter} & \makecell{Computing the covariance matrix.} & O($d$)\\
\bottomrule
\end{tabular*}
\end{adjustwidth}
\end{table}

\begin{adjustwidth}{-\extralength}{0cm}

\reftitle{References}

\PublishersNote{}
\end{adjustwidth}
\end{document}